\documentclass{article}
\usepackage{geoword_preprint,times}

\usepackage{amsmath,amsfonts,bm}

\def\eqref#1{equation~\ref{#1}}

\def\1{\bm{1}}

\DeclareMathAlphabet{\mathsfit}{\encodingdefault}{\sfdefault}{m}{sl}
\SetMathAlphabet{\mathsfit}{bold}{\encodingdefault}{\sfdefault}{bx}{n}

\usepackage[colorinlistoftodos]{todonotes}
\usepackage{booktabs}
\usepackage{tabularx}
\usepackage{array}
\usepackage{enumitem}
\usepackage{wrapfig}
\usepackage{titletoc}
\usepackage{xcolor}
\usepackage{colortbl}
\usepackage{graphicx}
\usepackage{pifont}
\usepackage[most]{tcolorbox}
\usepackage{fontawesome5}
\usepackage{hyperref}
\usepackage{url}

\usepackage{subcaption}

\usepackage[T1]{fontenc}
\usepackage{helvet}
\usepackage{booktabs}
\usepackage{graphicx}
\usepackage{makecell}
\newcommand{\stderr}[1]{\,{\scriptsize $\pm$\,#1}}
\usepackage{ragged2e}

\usepackage{fancyhdr}

\usepackage{placeins}

\usepackage[most]{tcolorbox}
\usepackage{ragged2e}

\usepackage[
    a4paper,
    top=1.0in,
    bottom=1.0in,
    left=0.75in,
    right=0.75in,
    headheight=34pt,
    headsep=10pt,
    footskip=18pt
]{geometry}

\makeatletter
\renewcommand{\normalsize}{\@setfontsize\normalsize\@xpt{11.5pt}}
\makeatother
\normalsize

\renewcommand{\footnoterule}{\kern-3pt\hrule width\textwidth height 0.4pt\kern 2.6pt}

\makeatletter
\renewcommand\section{\@startsection{section}{1}{\z@}{-2.0ex plus -0.5ex minus -.2ex}{1.5ex plus 0.3ex minus 0.2ex}{\large\bfseries\raggedright}}
\renewcommand\subsection{\@startsection{subsection}{2}{\z@}{-1.8ex plus -0.5ex minus -.2ex}{0.8ex plus .2ex}{\normalsize\bfseries\raggedright}}
\renewcommand\subsubsection{\@startsection{subsubsection}{3}{\z@}{-1.5ex plus -0.5ex minus -.2ex}{0.5ex plus .2ex}{\normalsize\bfseries\raggedright}}
\makeatother

\definecolor{benchmarkBlue}{HTML}{285F8F}
\definecolor{benchmarkPale}{HTML}{F1F6FC}
\definecolor{benchmarkColumn}{HTML}{E7F0FA}
\definecolor{benchmarkIntersection}{HTML}{D5E5F6}
\definecolor{benchmarkMuted}{HTML}{98A6B5}

\definecolor{epigraphBackground}{HTML}{F2F4F7}
\definecolor{epigraphBorder}{HTML}{507BB6}

\newtcolorbox{geowordepigraph}{
    enhanced,
    colback=epigraphBackground,
    colframe=epigraphBorder,
    boxrule=0.45pt,
    arc=6pt,
    left=7pt,
    right=7pt,
    top=6pt,
    bottom=6pt,
    before skip=8pt,
    after skip=10pt
}

\newtcolorbox{taskexample}[1]{
    enhanced,
    colback=benchmarkPale,
    colframe=benchmarkBlue,
    colbacktitle=benchmarkBlue,
    coltitle=white,
    fonttitle=\bfseries,
    title={#1},
    boxrule=0.8pt,
    arc=5pt,
    sharp corners=north,
    left=9pt,
    right=9pt,
    top=8pt,
    bottom=8pt,
    before skip=8pt,
    after skip=10pt,
    before upper=\justifying,
    breakable
}

\definecolor{safetyYellow}{HTML}{C99A29}
\definecolor{safetyPale}{HTML}{FFF9E9}
\definecolor{safetyTitle}{HTML}{F5E3AA}

\newtcolorbox{safetyevidence}[1]{
    enhanced,
    colback=safetyPale,
    colframe=safetyYellow,
    colbacktitle=safetyTitle,
    coltitle=black,
    fonttitle=\bfseries,
    title={#1},
    boxrule=0.8pt,
    arc=5pt,
    sharp corners=north,
    left=9pt,
    right=9pt,
    top=8pt,
    bottom=8pt,
    before skip=4pt,
    after skip=10pt,
    before upper=\justifying
}

\definecolor{otherEventGray}{HTML}{98A6B5}
\definecolor{otherEventPale}{HTML}{F2F4F7}
\definecolor{otherEventTitle}{HTML}{E2E6EB}

\newtcolorbox{otherevent}[1]{
    enhanced,
    colback=otherEventPale,
    colframe=otherEventGray,
    colbacktitle=otherEventTitle,
    coltitle=black,
    fonttitle=\bfseries,
    title={#1},
    boxrule=0.8pt,
    arc=5pt,
    sharp corners=north,
    left=9pt,
    right=9pt,
    top=8pt,
    bottom=8pt,
    before skip=4pt,
    after skip=10pt,
    before upper=\justifying,
    breakable
}

\definecolor{promptPurple}{HTML}{7766A8}
\definecolor{promptPale}{HTML}{F7F5FC}
\definecolor{promptTitle}{HTML}{E9E4F5}

\newtcolorbox{promptbox}[1]{
    enhanced,
    breakable,
    colback=promptPale,
    colframe=promptPurple,
    colbacktitle=promptTitle,
    coltitle=black,
    fonttitle=\bfseries,
    title={#1},
    boxrule=0.8pt,
    arc=5pt,
    sharp corners=north,
    left=9pt,
    right=9pt,
    top=8pt,
    bottom=8pt,
    before skip=8pt,
    after skip=10pt,
    before upper=\justifying
}

\definecolor{findingBackground}{HTML}{FFFCF5}
\definecolor{findingBorder}{HTML}{EAD99F}
\definecolor{findingTitle}{HTML}{FFF6DE}
\definecolor{findingAccent}{HTML}{B88A00}
\definecolor{findingHeading}{HTML}{1D292E}

\newtcolorbox{findingbox}[2]{%
    enhanced,
    width=\linewidth,
    colback=findingBackground,
    frame hidden,
    boxrule=0pt,
    borderline north={0.3pt}{0pt}{findingBorder},
    borderline south={0.3pt}{0pt}{findingBorder!65!white},
    arc=2pt,
    left=10pt,
    right=10pt,
    top=13pt,
    bottom=8pt,
    before skip=12pt plus 2pt minus 2pt,
    after skip=10pt plus 2pt minus 2pt,
    before upper=\justifying,
    fontupper=\small,
    coltitle=findingHeading,
    fonttitle=\small\sffamily\bfseries,
    title={%
        {\color{findingAccent}\faIcon[regular]{lightbulb}%
        \hspace{5pt}FINDING~#1}%
        \hspace{8pt}#2%
    },
    attach boxed title to top left={
        xshift=8pt,
        yshift=-\tcboxedtitleheight/2
    },
    boxed title style={
        enhanced,
        colback=findingTitle,
        frame hidden,
        boxrule=0pt,
        arc=2pt,
        left=9pt,
        right=9pt,
        top=4pt,
        bottom=4pt
    }
}

\iclrfinalcopy

\hypersetup{
    colorlinks=true,
    linkcolor=benchmarkBlue,
    citecolor=benchmarkBlue,
    urlcolor=benchmarkBlue,
    pdftitle={LiveGeo: Benchmarking Geographic Agents in Continuously Evolving Worlds},
    pdfauthor={Pengcheng Zhou, Xuanyu Liu, Yanchen Yin, Tianjie Ju, Bobo Li, Shengqiong Wu, Hao Fei, Mong-Li Lee, Wynne Hsu}
}

\begin{document}


\pagestyle{fancy}
\fancyhf{}

\fancyhfoffset{0pt}
\setlength{\headwidth}{\textwidth}

\fancyfoot[C]{\makebox[\textwidth][c]{\thepage}}

\renewcommand{\headrulewidth}{0pt}
\renewcommand{\footrulewidth}{0pt}

\fancypagestyle{firstpagelogos}{
    \fancyhf{}
    \fancyfoot[C]{\thepage}
    \fancyhead[C]{%
        \includegraphics[height=0.42in]{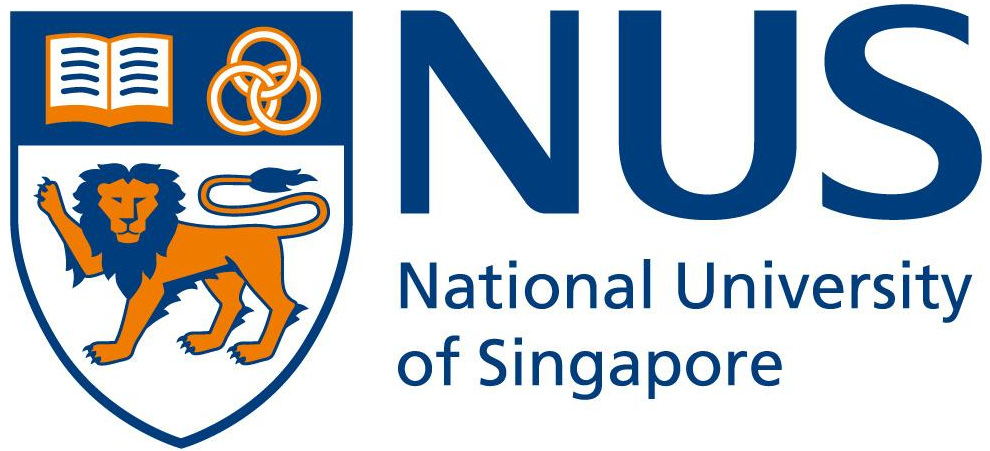}%
        \hfill
        \includegraphics[height=0.42in]{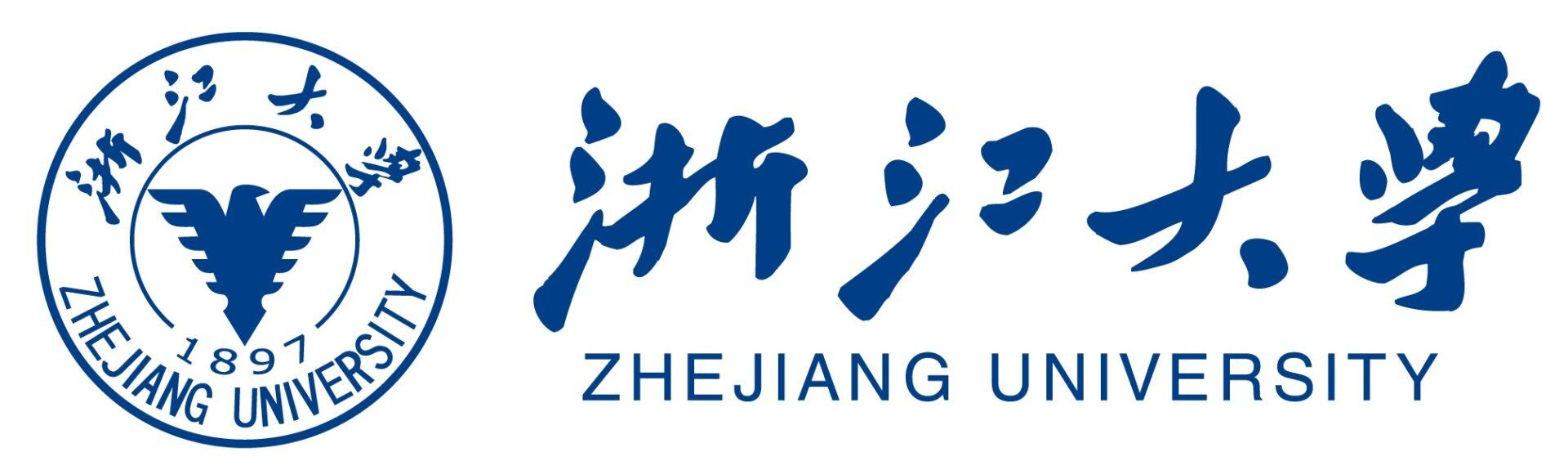}%
        \hfill
        \includegraphics[height=0.42in]{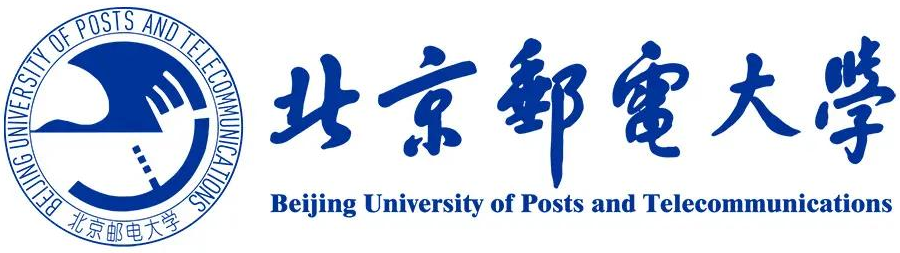}%
        \hfill
        \includegraphics[height=0.42in]{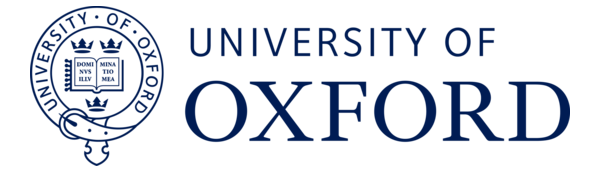}%
    }
    \renewcommand{\headrulewidth}{0.4pt}
}

\thispagestyle{firstpagelogos}

\begingroup
\setlength{\parindent}{0pt}
\vspace*{2pt}

{\centering
\rmfamily
\bfseries
\fontsize{16.5}{19.5}\selectfont
LiveGeo: Benchmarking Geographic Agents in Continuously Evolving Worlds
\par}


{\centering
\rmfamily
\bfseries
\fontsize{9.5}{12}\selectfont
Pengcheng Zhou\textsuperscript{1,$\dagger$},
Xuanyu Liu\textsuperscript{2,$\dagger$},
Yanchen Yin\textsuperscript{3},
Tianjie Ju\textsuperscript{1},
Bobo Li\textsuperscript{1},\\
Shengqiong Wu\textsuperscript{4,*},
Hao Fei\textsuperscript{4},
Mong-Li Lee\textsuperscript{1},
Wynne Hsu\textsuperscript{1}
\par}


{\centering
\rmfamily
\fontsize{9.2}{11}\selectfont
\textsuperscript{1}National University of Singapore,\quad
\textsuperscript{2}Zhejiang University,\\
\textsuperscript{3}Beijing University of Posts and Telecommunications,\quad
\textsuperscript{4}University of Oxford
\par}
{\renewcommand{\thefootnote}{\fnsymbol{footnote}}%
 \footnotetext[1]{Corresponding author.\quad
 \textsuperscript{\(\dagger\)}These authors contributed equally.}}

\vspace{5pt}

\newtcbox{\linkbadge}{on line, enhanced,
    colback=benchmarkPale, colframe=benchmarkBlue!35!white, boxrule=0.5pt, arc=4pt,
    left=6pt, right=6pt, top=2.5pt, bottom=2.5pt, boxsep=0pt}
{\centering\rmfamily\fontsize{9}{11}\selectfont
\hypersetup{hidelinks}
\href{https://livegeoworld.github.io/}{\linkbadge{\textcolor{benchmarkBlue}{\faGlobe}\hspace{0.5em}Project Page}}\hspace{1.2em}
\href{https://github.com/LiveGeoWorld/GeoWorld}{\linkbadge{\textcolor{benchmarkBlue}{\faGithub}\hspace{0.5em}GeoWorld Code}}\hspace{1.2em}
\href{https://github.com/LiveGeoWorld/LiveGeo-benchmark}{\linkbadge{\textcolor{benchmarkBlue}{\faDatabase}\hspace{0.5em}LiveGeo Benchmark}}
\par}

\vspace{5pt}

{\rmfamily\normalsize
We introduce \textbf{GeoWorld}, a controllable and continuously evolving geographic environment platform for LLM agents, and \textbf{LiveGeo}, a benchmark for evaluating reliable decision-making in real-world geographic environments using GeoWorld. 
Unlike existing evaluations that largely reason and plan over fixed geographic states, GeoWorld is grounded in real-world geographic structure and allows external world states to evolve along their own temporal processes while preserving verifiable state histories. 
LiveGeo places geographic tasks within this evolving environment, requiring agents to actively acquire task-relevant information, verify whether the evidence supporting prior decisions remains valid, and revise their plans when necessary. 
Our results reveal that even frontier LLMs struggle with static tasks, with no model consistently excelling across tasks, while strong static performance does not ensure dynamic reliability. 
Further analysis identifies a key bottleneck: agents struggle to acquire and translate relevant evidence into valid decisions and maintain their correctness as the world evolves. 
By releasing LiveGeo together with its GeoWorld platform, we aim to provide the community with a flexible and extensible infrastructure for studying the next generation of agents capable of reliable action in evolving geographic worlds.
\par}

\endgroup
\vspace{12pt}

\begin{figure}[!h]
\centering
\includegraphics[width=0.97\textwidth]{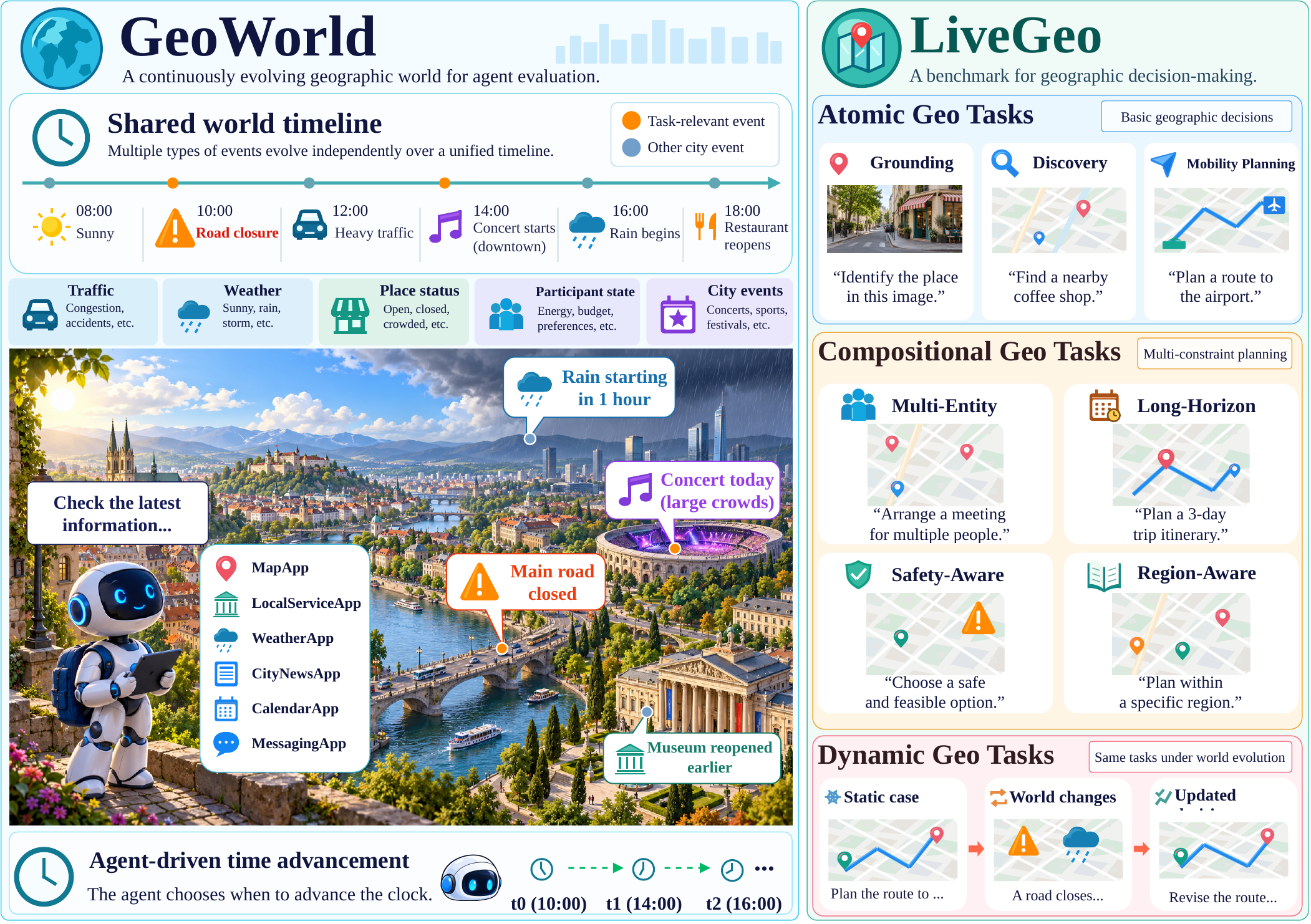}
\caption{\textbf{Overview of GeoWorld and LiveGeo.} GeoWorld is a geographic environment whose world states evolve independently of the agent, with time advanced by agent actions. LiveGeo builds on it to evaluate agents on atomic, compositional, and dynamic geographic decision-making tasks.}
\label{fig:overview}
\end{figure}
\vfill
\clearpage

\section{Introduction}
Geographic intelligence, the ability to understand, reason about, and act upon real-world geographic information, has advanced substantially with the emergence of large language and vision-language models~\citep{roberts2023gpt4geo,manvi2024geollm,dihan2025mapeval} and is increasingly being deployed in diverse practical applications~\citep{ju2024globe,feng2025agentmove,yuan2025llmap,jiang2024urbanllm}.
In parallel, its evaluation has evolved considerably.
Early benchmarks primarily assessed isolated capabilities, such as geographic knowledge and spatial reasoning~\citep{roberts2023gpt4geo, manvi2024geollm}, or perception-oriented abilities including image geolocation and map reasoning~\citep{dihan2025mapeval}. 
These tasks typically operate over fixed inputs and do not require interaction with an external environment.
With the emergence of agentic systems, evaluation has shifted toward end-to-end task execution in geographic environments.
Recent benchmarks place agents in simulated or real-world urban settings~\citep{feng2025citybench, lai2026ustbench}, or formulate realistic planning scenarios in which agents must actively acquire information, interact with external tools, and construct feasible plans~\citep{pmlr-v235-xie24j, chaudhuri2025tripcraft, ni2025tp, cheng2026beyond, cheng2026grouptravelbench, song2026mobilitybench}. 
Collectively, these efforts have broadened both the geographic information agents must process and the complexity of the tasks they must accomplish.

However, existing benchmarks largely evaluate agents in environments where the task-relevant geographic state remains fixed during execution, typically using a static snapshot of the world~\citep{cheng2026beyond, song2026mobilitybench}. 
Even benchmarks that incorporate environmental dynamics generally focus on specific urban processes or predefined disruptions that are explicitly exposed to the agent~\citep{feng2025citybench, lai2026ustbench, karmakar2026triptide}.
Real-world geographic environments, by contrast, evolve continuously and independently of the agent, such as road conditions, place availability, and regional events, while only some of these changes matter to the task.
Reliable decision-making in such circumstances therefore requires agents not only to act on available and dynamic information, but also to acquire task-relevant evidence, verify whether prior evidence and assumptions remain valid, and revise decisions when they do not.
A separate limitation concerns what existing benchmarks evaluate (Table~\ref{tab:benchmark_comparison}).
Because they typically focus on specific applications or workflows, they primarily measure whether an end task can be completed rather than which geographic constraints and decision dependencies the agent understands and relies on. 
Consequently, failures are difficult to attribute to deficiencies in geographic understanding, feasibility assessment, or the ability to maintain valid decisions over time. 
This raises a central question: \emph{can agents sustain reliable decision-making in continuously evolving geographic environments while keeping their decisions grounded in evidence that remains valid as the world changes?}

\begin{table}[!ht]
\centering
\small
\setlength{\tabcolsep}{1.8pt}
\renewcommand{\arraystretch}{1.05}
\hypersetup{hidelinks}
\newcommand{\benchyes}{\textcolor{benchmarkBlue}{\ding{51}}}
\newcommand{\benchno}{\textcolor{benchmarkMuted}{\textendash}}
\newcommand{\benchours}{\textcolor{benchmarkBlue}{\ding{52}}}
\newcommand{\benchhead}[2]{%
    {\fontsize{8.5}{10}\selectfont\shortstack{\textbf{#1}\\#2}}}

\caption{Comparison of representative geographic-agent benchmarks.}
\label{tab:benchmark_comparison}
\vspace{-2mm}
\begin{tabularx}{\textwidth}{@{}>{\raggedright\arraybackslash}X
    *{7}{c} >{\columncolor{benchmarkColumn}}c@{}}
\toprule
\textbf{Evaluation dimension}
& \benchhead{MapEval}{\citeyearpar{dihan2025mapeval}}
& \benchhead{CityBench}{\citeyearpar{feng2025citybench}}
& \benchhead{USTBench}{\citeyearpar{lai2026ustbench}}
& \benchhead{\shortstack{Travel\\Planner}}{\citeyearpar{pmlr-v235-xie24j}}
& \benchhead{\shortstack{Travel\\Bench}}{\citeyearpar{cheng2026grouptravelbench}}
& \benchhead{\shortstack{Mobility\\Bench}}{\citeyearpar{song2026mobilitybench}}
& \benchhead{TripTide}{\citeyearpar{karmakar2026triptide}}
& \textcolor{benchmarkBlue}{\benchhead{LiveGeo}{(ours)}} \\
\midrule
\multicolumn{8}{@{}l}{\textit{Geographic decision capabilities}}
& \\
Geo understanding
& \benchyes & \benchyes & \benchyes & \benchno
& \benchno & \benchno & \benchno & \benchours \\
Geo feasibility
& \benchno & \benchyes & \benchyes & \benchyes
& \benchyes & \benchyes & \benchyes & \benchours \\
Decision grounding
& \benchno & \benchno & \benchno & \benchno
& \benchno & \benchyes & \benchno & \benchours \\
\midrule
\multicolumn{8}{@{}l}{{\textit{Reliability as the world evolves}}}
& \\
Disruption adaptation
& \benchno & \benchno & \benchno & \benchno
& \benchno & \benchno & \benchyes & \benchours \\
\rowcolor{benchmarkPale}
\textcolor{benchmarkBlue}{\textbf{Evolving world}}
& \benchno & \benchno & \benchno & \benchno
& \benchno & \benchno & \benchno
& \cellcolor{benchmarkIntersection}\benchours \\
\rowcolor{benchmarkPale}
\textcolor{benchmarkBlue}{\textbf{Change relevance}}
& \benchno & \benchno & \benchno & \benchno
& \benchno & \benchno & \benchno
& \cellcolor{benchmarkIntersection}\benchours \\
\rowcolor{benchmarkPale}
\textcolor{benchmarkBlue}{\textbf{Decision validity over time}}
& \benchno & \benchno & \benchno & \benchno
& \benchno & \benchno & \benchno
& \cellcolor{benchmarkIntersection}\benchours \\
\bottomrule
\end{tabularx}
\end{table}

To address this question, we first develop \textbf{GeoWorld}, a controllable and continuously evolving interactive environment platform grounded in real-world urban geography.
GeoWorld jointly maintains multiple world states, including road networks, place states, and regional events, and allows external events to unfold according to their own temporal processes, independent of agent actions. 
The platform enables different tasks to be executed within a unified and continuously evolving geographic world while preserving verifiable world states throughout. 
Building on GeoWorld, we further introduce \textbf{LiveGeo}, a benchmark for evaluating the reliable geographic decision-making of general-purpose LLM agents in continuously evolving environments. 
LiveGeo contains 7 task categories and 1,237 verifiable task instances, covering 6 real-world cities, 1,376 POIs, and 31 tool types, with matched static and dynamic task settings. Rather than making decisions based only on the environmental state at a single point in time, agents in LiveGeo must actively acquire relevant information during task execution and adjust their plans when changes in the world state invalidate the basis of previous decisions.

We systematically evaluate 9 representative proprietary and open-source models under a unified ReAct-based agent scaffold and tool-access protocol. Our results show that current agents remain far from solving geographic decision-making even on atomic tasks: on extensively studied tasks such as place grounding and place discovery, the strongest agent achieves only 38.0\% pass@1. Performance drops further on compositional tasks, indicating that the primary difficulty lies not merely in invoking individual tools, but in jointly satisfying multiple interdependent geographic constraints. World dynamics introduce an additional and consistent performance cost: under matched dynamic settings, performance decreases by 70.3\% under Task-Only Dynamics and 79.3\% under Shared Dynamics relative to the corresponding static settings across models and task categories. Further analysis identifies \textit{Evidence Hallucination} in static settings and \textit{Change Monitoring Failure} in dynamic settings as major bottlenecks to reliable decision-making. Agents often make unsupported geographic claims and fail to acquire updated information after necessary world changes. These results show that current agents have some ability to make one-shot decisions, but remain limited in maintaining decision validity as the world continues to evolve.

Our main contributions are threefold: 
\begin{itemize}[leftmargin=2.0em]
    \item We introduce \textbf{GeoWorld}, a controllable and continuously evolving geographic platform grounded in real-world urban structure, in which world states evolve independently of agent actions while remaining fully verifiable over time.
    \item We present \textbf{LiveGeo}, a capability-structured benchmark with matched static and dynamic settings for evaluating general-purpose LLM agents on continuous information acquisition, evidence validation, and adaptive replanning in evolving geographic environments. 
    \item We conduct a systematic evaluation of representative LLM agents on LiveGeo, revealing substantial performance degradation under world dynamics and characterizing key failure modes in maintaining evidence-grounded decisions and adapting plans as the environment changes.
\end{itemize}

\section{Related Work}

\noindent \textbf{Geographic Intelligence Benchmarks}
Existing geographic intelligence benchmarks cover a broad range of capabilities, from geographic knowledge and spatial reasoning to visual geographic understanding and real-world decision-making. 
One line of work examines the geographic knowledge encoded in language models and their ability to reason about spatial relations~\citep{roberts2023gpt4geo,manvi2024geollm}; MapEval further evaluates reasoning over maps and other structured geographic representations~\citep{dihan2025mapeval}. 
Another line of work focuses on visual geographic understanding. 
Visual geolocation benchmarks evaluate whether models can infer real-world locations from geographic cues in street-level and cross-view imagery~\citep{astruc2024openstreetview,kristoffersen2026mmlandmarks,talreja2026georc,zhou2026hologeo}; remote-sensing and geospatial vision-language benchmarks further cover visual question answering, scene and object understanding, visual grounding, and spatial and relational reasoning~\citep{kuckreja2024geochat,li2024vrsbench,danish2025geobench,Xiao_2026_CVPR}. 
A separate line of work extends geographic intelligence evaluation to interactive decision-making. 
CityBench~\citep{feng2025citybench} and USTBench~\citep{lai2026ustbench} place agents in urban environments and evaluate their ability to interpret and use heterogeneous urban information; benchmarks for travel, mobility, and local-life planning further require agents to actively acquire information, invoke external tools, and construct feasible plans under real-world constraints~\citep{pmlr-v235-xie24j,chaudhuri2025tripcraft,ni2025tp,cheng2026beyond,cheng2026grouptravelbench,song2026mobilitybench}. 
These studies cover diverse input modalities and application settings, but their task structures typically focus on specific problems or application workflows. 
LiveGeo instead adopts a capability-structured evaluation perspective, progressing from basic geographic operations to compositional decisions whose feasibility is jointly determined by multiple interdependent constraints.

\noindent \textbf{Interactive and Dynamic Geographic Environments}
Interactive urban environments have been used to evaluate agents' reasoning and decision-making over complex spatiotemporal states. Existing urban-agent benchmarks simulate urban processes or provide interactive environments in which agents can plan and adapt as spatiotemporal conditions evolve~\citep{feng2025citybench,lai2026ustbench}; more recent work has also begun to evaluate adaptation to real-world disruptions, such as revising existing travel plans when transportation or place availability changes~\citep{karmakar2026triptide}. 
However, existing dynamic evaluations are typically organized around specific urban processes or application tasks, and less often systematically assess whether the evidence supporting a decision remains valid as the environment changes, and whether the decision should be updated accordingly.
This aspect is also difficult to capture in some real-world planning benchmarks, where external tool outputs are stabilized through caching or deterministic replay, leaving the information available during task execution unchanged~\citep{cheng2026beyond,song2026mobilitybench}.
LiveGeo places this problem within a unified geographic world, GeoWorld, where changes in external states can alter the evidence on which an ongoing geographic task depends; whether a change affects the current decision is determined jointly by the task and the world state. Accordingly, LiveGeo focuses not on whether an agent can respond to a single predefined disruption, but on whether it can maintain consistency between evidence and decisions in a continuously evolving geographic environment.

\section{GeoWorld}
\begin{figure*}[t]
    \centering
    \includegraphics[width=0.99\textwidth]{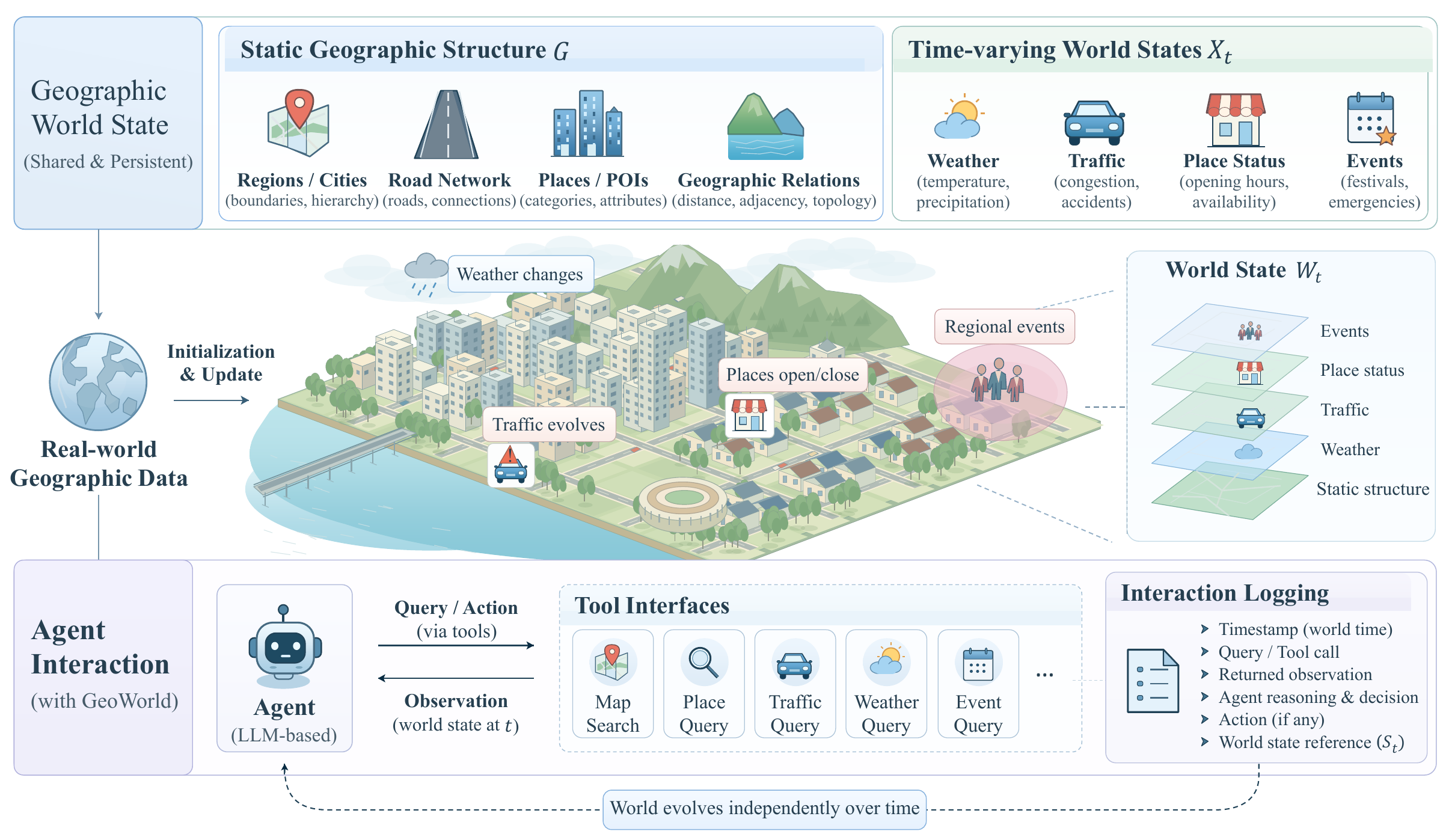}
    \vspace{-2mm}
    \caption{\textbf{GeoWorld architecture.}
    A persistent geographic structure $G$ is paired with time-varying world states $X_t$ grounded in real-world geographic data. The world evolves independently of agent actions; agents act through unified tools, and each interaction is recorded against the world state at which it occurred, so that decisions can be verified afterwards.
    }
    \label{fig:GeoWorld_architecture}
    \vspace{-2mm}
\end{figure*}

GeoWorld provides a persistent geographic environment in which heterogeneous world states are grounded in a shared spatial structure and evolve along a common timeline. 
Its architecture comprises three interconnected components: world state representation, temporal world dynamics, and an agent-world interface, as illustrated in Figure~\ref{fig:GeoWorld_architecture}.

\paragraph{World State Representation}
GeoWorld represents the geographic world at time $t$ as $W_t=(G,X_t)$, where $G$ captures the relatively stable geographic structure, including places, roads, regions, and their spatial relations, while $X_t$ describes dynamic conditions such as traffic, weather, place status, and regional events.
Rather than maintaining these states as independent data sources, GeoWorld associates them with a common set of geographic entities and spatial references, providing a consistent spatial foundation for world-state updates and agent observations.
We construct the geographic environment from real-world data, primarily using Google Places, geocoding, and routing services~\citep{googlemaps2026}.
The current implementation spans \textbf{six cities}: \textit{Shanghai}, \textit{Tokyo}, \textit{Singapore}, \textit{London}, \textit{New York}, and \textit{Cape Town}, comprising \textbf{88 regions}, \textbf{1,376 POIs}, and \textbf{107,372 cached route records}.
Details of the geographic representation, data preprocessing, and spatial coverage are provided in Appendix~\ref{app:world_representation}.

\paragraph{Temporal World Dynamics.}
Building on this shared spatial representation, GeoWorld models how geographic conditions change over time.
A global virtual clock provides a common temporal reference for heterogeneous state updates, each defined by an effective time and, where applicable, a duration.
Each episode begins at a specified time and advances within a bounded temporal window.
As the clock advances, GeoWorld resolves scheduled changes to determine the world state $W_t$ at the current time.
These changes are exogenous to agent actions, ensuring that the underlying world evolves according to a predefined trajectory rather than being altered by agent decisions.
This separation enables reproducible and replayable world dynamics while allowing agents to encounter changing conditions during task execution.
Further details of clock management and temporal state updates are provided in Appendix~\ref{app:temporal_dynamics}.

\paragraph{Agent-World Interface.}
To expose the evolving world to agents, GeoWorld provides a unified tool interface for place search, routing and traffic, weather, place status, and regional events.
All tool responses are resolved against the world state at the current virtual time, ensuring that agents observe temporally and spatially consistent information across different services.
GeoWorld records agent queries, tool observations, and actions together with their corresponding timestamps and world-state references.
These aligned interaction traces capture both the information available to agents and their subsequent decisions, providing a basis for reconstructing task execution and evaluating agent behavior under evolving world conditions.
The complete tool set and interaction protocol are described in Appendix~\ref{app:tool_interfaces}.

\section{LiveGeo}
Building on GeoWorld, LiveGeo provides a benchmark for evaluating geographic decision-making under both static and evolving world conditions.
It organizes tasks by the geographic capabilities they require and uses matched static-dynamic settings to examine whether agents can not only make valid initial decisions but also reassess and revise them as the underlying world changes.

\begin{table}[t]
\centering
\small
\setlength{\tabcolsep}{4.5pt}
\renewcommand{\arraystretch}{1.12}
\newcommand{\taxcategory}[1]{\textcolor{benchmarkBlue}{\textbf{#1}}}

\caption{LiveGeo task taxonomy along geographic capability and temporal setting.}
\label{tab:task_taxonomy}
\begin{tabularx}{\textwidth}{@{}>{\raggedright\arraybackslash}p{0.18\textwidth}
    >{\raggedright\arraybackslash}p{0.40\textwidth}
    >{\raggedright\arraybackslash}X@{}}
\toprule
\textbf{Task category} & \textbf{Decision dependency} & \textbf{Illustrative request} \\
\midrule
\rowcolor{benchmarkColumn} \multicolumn{3}{l}{\textit{Atomic Geo Tasks}\enspace\textcolor{benchmarkBlue}{--- Establish local validity}} \\
\taxcategory{Grounding}
& \textbf{Evidence to entity.} Resolve visual or textual clues to the exact requested place.
& Identify the South Bank cinema from street-view clues near Hayward Gallery. \\[3pt]
\taxcategory{Discovery}
& \textbf{Intent to destination.} Select places satisfying joint spatial, service, and preference constraints.
& Find a quiet vegetarian caf\'e within a 10-minute walk of Oxford Street that accepts reservations. \\[3pt]
\taxcategory{Mobility Planning}
& \textbf{Journey feasibility.} Satisfy waypoints, dwell times, travel modes, and deadlines.
& Stop at The Vineyard for 20 minutes and reach the final destination within 65 minutes total. \\[3pt]
\rowcolor{benchmarkColumn} \multicolumn{3}{l}{\textit{Compositional Geo Tasks}\enspace\textcolor{benchmarkBlue}{--- Compose jointly valid decisions}} \\
\taxcategory{Multi-Entity}
& \textbf{Across participants.} Find a common solution under individual location and availability constraints.
& Arrange dinner for three people with different origins and departure times, sharing at least 60 minutes before 20:30. \\[3pt]
\taxcategory{Long-Horizon}
& \textbf{Across decisions.} Preserve downstream feasibility across linked activities and time windows.
& Plan a three-day Shanghai trip around fixed arrival, hotel check-in/out, public transit, and daily budgets. \\[3pt]
\taxcategory{Safety-Aware}
& \textbf{Environmental conditions.} Ground destination and access-route choices in relevant safety evidence.
& Find late-night lodging within a 12-minute drive of UCT, supported by accessibility and safety evidence. \\[3pt]
\taxcategory{Region-Aware}
& \textbf{Regional context.} Make local practices, schedules, and geographic conditions constrain valid choices.
& Find a reservable bak kut teh venue in Raffles Place, reachable from Dhoby Ghaut. \\[3pt]
\bottomrule
\end{tabularx}

\end{table}

\subsection{Task Taxonomy}
As summarized in Table~\ref{tab:task_taxonomy}, LiveGeo organizes tasks along two dimensions: geographic capability and temporal setting.

\paragraph{Geographic Capabilities.}
We consider seven task categories, comprising three atomic and four compositional tasks.
\emph{Atomic Geo Tasks} target fundamental geographic operations:
\emph{Grounding} identifies and localizes geographic entities;
\emph{Discovery} selects places satisfying spatial, semantic, or situational constraints;
and \emph{Mobility Planning} constructs feasible routes or travel plans under transportation constraints.

\emph{Compositional Geo Tasks} integrate these operations into more complex decision scenarios, emphasizing different forms of interdependence.
\emph{Multi-Entity} tasks coordinate constraints across multiple participants, places, or activities;
\emph{Long-Horizon} tasks involve sequences of interdependent decisions;
\emph{Safety-Aware} tasks require decisions to account for environmental or situational risks;
and \emph{Region-Aware} tasks incorporate region-specific geographic structures and constraints.
These tasks extend beyond individual geographic operations to evaluate whether multiple decisions and their associated constraints can be satisfied jointly.

\paragraph{Temporal Settings.}
To examine decision-making under changing conditions, LiveGeo instantiates selected tasks in matched static and dynamic settings.
In the static setting, the relevant world state remains fixed throughout task execution.
In the dynamic setting, task-relevant conditions, including traffic, weather, place status, and availability, may change over time.
Each matched pair shares the same initial world state, task constraints, and user request, differing in the introduction of temporal state changes.
This design allows us to examine how agents respond when previously valid decisions must be reassessed or revised, while controlling for the initial task requirements.
Complete task definitions and examples are provided in Appendix~\ref{app:task_taxonomy}.

\subsection{Benchmark Construction}

\begin{figure*}[t]
    \centering
    \includegraphics[width=0.99\textwidth]{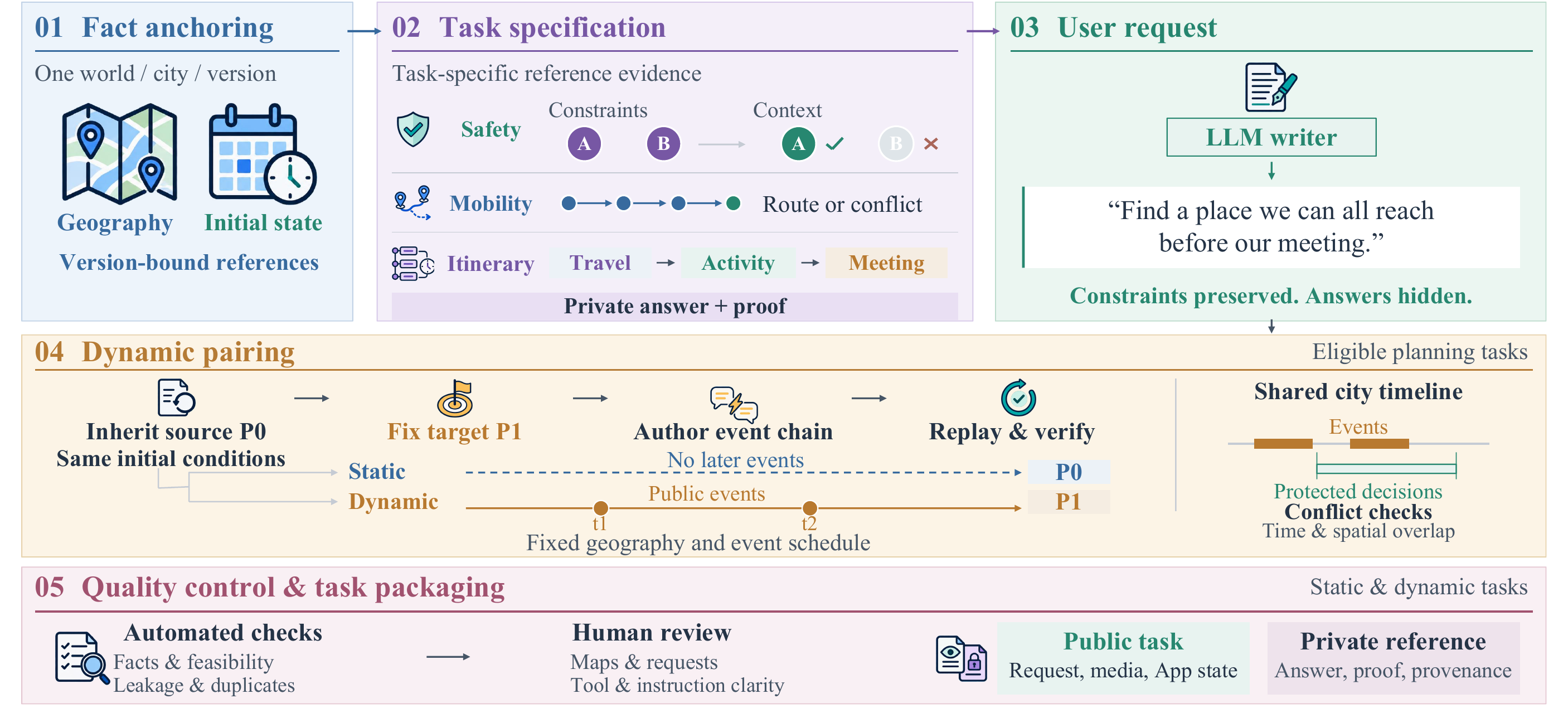}
    \caption{\textbf{LiveGeo construction pipeline.}
        (a) Static tasks: a target solution is selected from GeoWorld, verifiable constraints are specified, and a natural-language request is generated by an LLM, then validated programmatically and reviewed by humans.
        (b) Dynamic tasks keep the same initial conditions and request but inject validated temporal events that change the reference solution, yielding matched static and dynamic task pairs.
    }
    \label{fig:LiveGeo_Overview}
\end{figure*}

\paragraph{Construction Pipeline.}
Following the task taxonomy, LiveGeo first constructs verified static tasks and then introduces controlled world changes to create matched dynamic counterparts. 
This process preserves the initial task requirements while enabling evaluation under evolving conditions, as illustrated in Figure~\ref{fig:LiveGeo_Overview}.

\textit{(a) Static task construction.}
Given an initial world state $W_0=(G,X_0)$, we first identify a feasible reference solution $GT_0$ and derive a set of constraints $C_0$ from the corresponding geographic facts.
For example, in the group-dinner scenario in Figure~\ref{fig:LiveGeo_Overview}, Restaurant A is selected as a reference solution, with constraints requiring vegetarian options, availability at the requested dinner time, and travel times within 20 minutes for all participants.
We then apply programmatic checks and constraint solvers to verify that the reference solution satisfies all constraints, that the constraints are grounded in the world state, and that the information required to solve the task is accessible through the agent's tools.
Once verified, an LLM converts the structured task into a natural-language user request while preserving its requirements and withholding the reference solution.
The request is subsequently reviewed by human annotators, and only approved tasks are retained.

\textit{(b) Dynamic task construction.}
For each selected static task, we construct a dynamic counterpart preserving the initial world state $W_0$, constraints $C_0$, reference solution $GT_0$, and user request.
We then specify an updated reference solution $GT_1$ and use an LLM to propose a sequence of timestamped events that affect the world state.
In the group-dinner example, Restaurant A becomes unavailable while Restaurant B extends its opening hours, requiring the original choice to be reconsidered.
Each event specifies an observable description and a corresponding world-state update.
We verify the temporal consistency of the event sequence, the validity of the reference solutions under their corresponding world states, and the observability of task-relevant changes through the agent's tools.
The events are reviewed by human annotators before inclusion in the benchmark.
Consequently, matched tasks begin with the same decision problem but differ in the evolution of the underlying world, allowing us to evaluate whether agents can maintain valid decisions as conditions change.

\paragraph{Benchmark Statistics.}
LiveGeo contains \textbf{1,237 task instances} across seven geographic task categories, including \textbf{1,088 static} and \textbf{149 matched dynamic} instances.
Spanning six cities, the benchmark covers diverse geographic scenarios, including place discovery across different POI categories, mobility planning with multiple transportation modes, and compositional decisions involving multiple participants or multi-day itineraries.
Detailed task distributions and geographic coverage are reported in Appendix~\ref{app:benchmark_statistics}, while reference-solution selection, constraint generation, and verification procedures are described in Appendix~\ref{app:construction_pipeline}.

\subsection{Evaluation Protocol}

LiveGeo evaluates agents through the unified GeoWorld interface, using the same tool access and interaction protocol across tasks.
Static tasks are executed against a fixed world state, whereas dynamic tasks unfold within a bounded virtual-time window following a predefined event sequence.
Agents decide when to query the environment, advance virtual time, and revisit earlier decisions.
Their interactions are recorded alongside the corresponding world states, enabling both outcome-based verification and trajectory-level analysis.

We determine task success using task-specific verification criteria. These criteria cover four main aspects of geographic decision-making, with the applicable aspects varying across task categories.
\emph{Decision Correctness} checks whether the final decision meets the task-specific correctness criteria, including place selection and feasibility judgments.
\emph{Plan Executability} further checks whether the generated decisions satisfy the relevant spatial, temporal, transportation mode, and other operational constraints.
Beyond outcome validity, \emph{Evidence Support} assesses whether the agent's interaction history contains observations that substantiate its decisions.
For dynamic tasks, \emph{Dynamic Adaptability} further examines whether agents respond to task-relevant world changes by revalidating and, when necessary, revising their decisions.

Evaluation relies on predefined output contracts and task-specific constraint checks.
State-aligned interaction traces further support evidence verification and analysis of agent behavior under changing conditions.
Detailed metrics, verification rules, and trajectory analysis procedures are provided in Appendix~\ref{app:bench_evaluation}.

\section{Experiment}

Our analysis proceeds in three steps, from forming decisions, to maintaining them as the world evolves, to locating where reliability breaks down:
\begin{itemize}[leftmargin=2.0em, nosep]
    \item \textbf{Q1:} How well can current agents make valid geographic decisions? (Section~\ref{sec:core_results})
    \item \textbf{Q2:} How well can current agents maintain valid geographic decisions as the world evolves? (Sections~\ref{sec:world_dynamics} and~\ref{sec:decision_maintenance})
    \item \textbf{Q3:} Where does reliable decision-making break down? (Section~\ref{sec:failure_modes})
\end{itemize}
Additional task-specific analyses are provided in Appendix~\ref{app:failure_analysis}.

\begin{table*}[t]
\centering
\small
\setlength{\tabcolsep}{5pt}
\renewcommand{\arraystretch}{1.15}
\caption{\textbf{Success rates (\%) on LiveGeo.}
Dynamic Evaluation reports the matched static baseline (Static) and the two dynamic settings on the same tasks.
Per column, \textbf{bold} and \underline{underlined} mark the best and second-best distinct scores (ties share a mark). Models are grouped by family. Values are reported as mean $\pm$ standard error.
}
\label{tab:main_results}
\resizebox{\textwidth}{!}{%
\begin{tabular}{@{}l*{10}{c}@{}}
\toprule
& \multicolumn{3}{c}{\textbf{Atomic Geo Tasks}}
& \multicolumn{4}{c}{\textbf{Compositional Geo Tasks}}
& \multicolumn{3}{c}{\textbf{Dynamic Evaluation}} \\
\cmidrule(lr){2-4}
\cmidrule(lr){5-8}
\cmidrule(l){9-11}
\textbf{Model}
& \textbf{Grounding}
& \textbf{Discovery}
& \textbf{Mobility}
& \textbf{Multi-Entity}
& \textbf{Long-Horizon}
& \textbf{Safety}
& \textbf{Region}
& \textbf{Static}
& \textbf{Shared}
& \textbf{Task-Only} \\
\midrule

GPT-5.6 Sol
& \underline{31.1}\stderr{4.0}
& 26.9\stderr{3.4}
& \textbf{31.4}\stderr{3.5}
& 35.2\stderr{3.6}
& 2.0\stderr{1.1}
& 7.4\stderr{2.4}
& 12.3\stderr{2.6}
& 18.1\stderr{3.2}
& 1.3\stderr{0.9}
& 2.7\stderr{1.3} \\

Claude Opus 5
& 26.5\stderr{3.8}
& 26.3\stderr{3.4}
& 26.3\stderr{3.3}
& 36.9\stderr{3.6}
& 1.3\stderr{0.9}
& \underline{17.4}\stderr{3.4}
& 24.5\stderr{3.4}
& 21.5\stderr{3.4}
& 1.3\stderr{0.9}
& 4.7\stderr{1.7} \\

Claude Sonnet 5
& 26.5\stderr{3.8}
& 25.7\stderr{3.3}
& 25.1\stderr{3.3}
& 36.4\stderr{3.6}
& 0.0\stderr{0.0}
& \textbf{19.8}\stderr{3.6}
& 23.9\stderr{3.3}
& \underline{22.8}\stderr{3.4}
& 2.0\stderr{1.2}
& 2.7\stderr{1.3} \\

Gemini 3.1 Pro Preview
& 22.0\stderr{3.6}
& 30.4\stderr{3.5}
& 21.1\stderr{3.1}
& 33.5\stderr{3.6}
& \textbf{8.7}\stderr{2.3}
& 15.7\stderr{3.3}
& 27.0\stderr{3.5}
& 21.5\stderr{3.4}
& 3.4\stderr{1.5}
& 6.0\stderr{2.0} \\

Gemini 3.7 Flash
& \textbf{37.1}\stderr{4.2}
& \textbf{38.0}\stderr{3.7}
& 19.4\stderr{3.0}
& \underline{42.6}\stderr{3.7}
& \textbf{8.7}\stderr{2.3}
& \underline{17.4}\stderr{3.4}
& \underline{27.6}\stderr{3.5}
& \textbf{26.2}\stderr{3.6}
& \textbf{13.4}\stderr{2.8}
& \textbf{14.1}\stderr{2.9} \\

Grok 4.5
& 6.1\stderr{2.1}
& 17.5\stderr{2.9}
& 17.1\stderr{2.8}
& 17.6\stderr{2.9}
& \underline{6.0}\stderr{1.9}
& 11.6\stderr{2.9}
& 20.2\stderr{3.1}
& 14.8\stderr{2.9}
& \underline{6.7}\stderr{2.0}
& \underline{10.1}\stderr{2.5} \\

Qwen3.8-Max
& 28.0\stderr{3.9}
& \underline{32.7}\stderr{3.6}
& \underline{30.3}\stderr{3.5}
& \textbf{43.8}\stderr{3.7}
& 1.3\stderr{0.9}
& 10.7\stderr{2.8}
& \textbf{28.2}\stderr{3.5}
& \underline{22.8}\stderr{3.4}
& 4.0\stderr{1.6}
& 4.7\stderr{1.7} \\

Kimi K3
& 13.6\stderr{3.0}
& 6.4\stderr{1.9}
& 6.3\stderr{1.8}
& 11.4\stderr{2.4}
& 0.7\stderr{0.7}
& 2.5\stderr{1.4}
& 7.4\stderr{2.0}
& 6.0\stderr{2.0}
& 2.0\stderr{1.2}
& 2.7\stderr{1.3} \\

DeepSeek-V4-Flash-Vision-Exp
& 20.5\stderr{3.5}
& 12.9\stderr{2.6}
& 21.1\stderr{3.1}
& 19.9\stderr{3.0}
& 2.0\stderr{1.1}
& 13.2\stderr{3.1}
& 22.1\stderr{3.2}
& 18.1\stderr{3.2}
& 1.3\stderr{0.9}
& 3.4\stderr{1.5} \\

\bottomrule
\end{tabular}%
}
\end{table*}

\begin{figure*}[!t]
    \centering
    \includegraphics[width=0.99\textwidth]{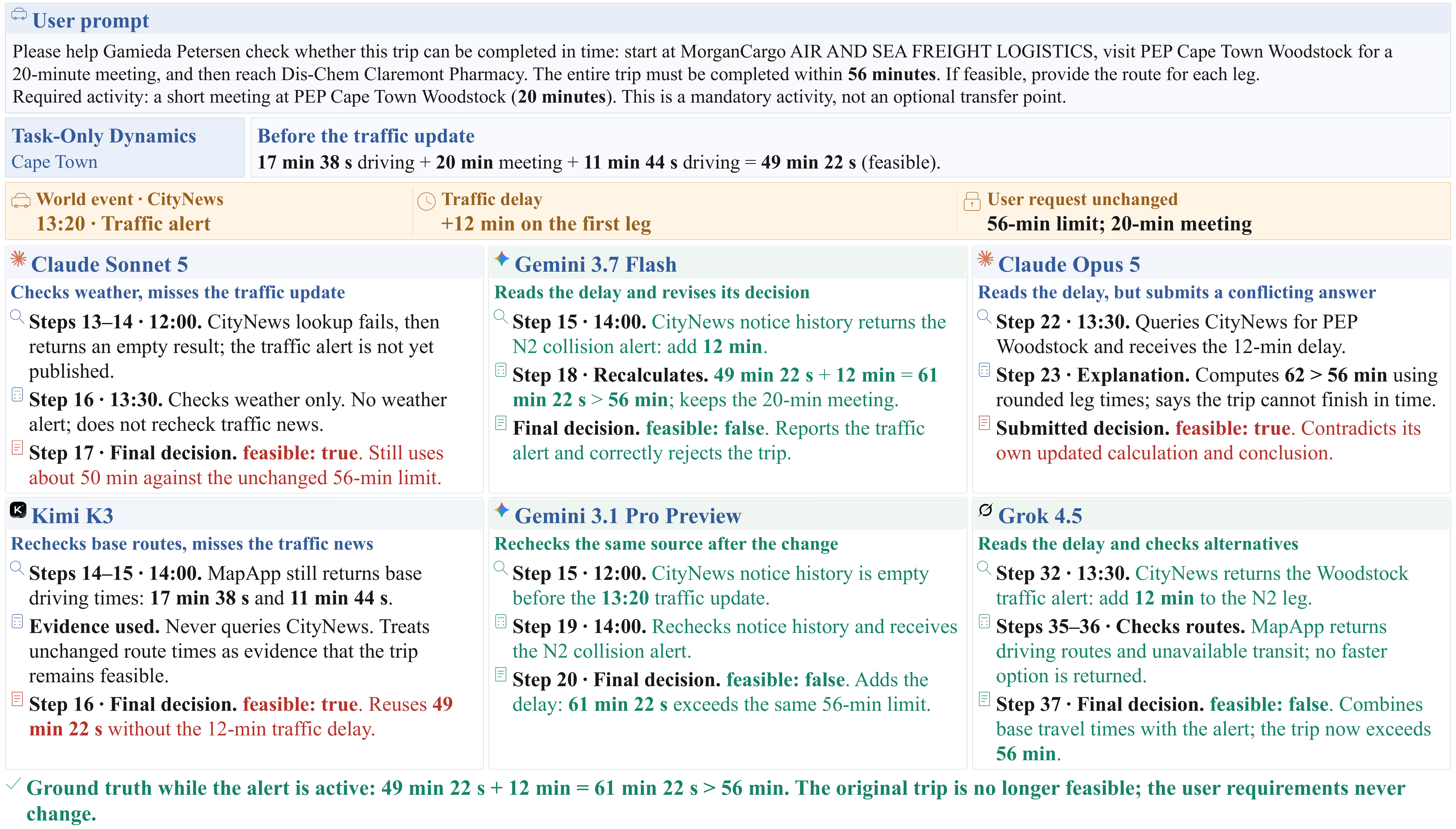}
    \caption{
        \textbf{Six models facing the same world change.}
        A Task-Only Dynamics case in which a traffic disruption invalidates an initially feasible trip. The traces differ in whether a model acquires the update, interprets its impact, and revises its final decision; full traces are given in Appendix~\ref{app:case_study}.
    }
    \label{fig:dynamic_case}
\end{figure*}

\subsection{Experimental Setup}
We evaluate representative proprietary and open-source models~\citep{openai2026gpt56,anthropic2026claude5,googledeepmind2026modelcards,spacexai2026grok45,qwen38,team2026kimi,xu2026deepseek} under a unified ReAct-based agent scaffold. 
All models receive the same tool-access protocol and interaction budget, with a maximum of 200 interaction steps per task. 
Task-specific output contracts are used to standardize the required response format for different task categories, while remaining identical across models evaluated on the same task. 
We retain the full interaction context throughout execution, with all evaluated models supporting context windows of at least 128K tokens. The temperature is set to 0.1, with a maximum generation length of 10K tokens per turn. 
For dynamic tasks, all models are evaluated against the same replayed world timeline and event schedule, ensuring that each model encounters identical environmental changes. 
Each task instance is evaluated over three independent runs. 
Detailed model versions, prompts, output contracts, and runtime configurations are provided in Appendix~\ref{app:experimental_setup}.

\subsection{Geographic Decision-Making Performance}
\label{sec:core_results}

Table~\ref{tab:main_results} presents the performance of nine models across seven static geographic decision-making task categories. Overall, current agents remain limited in their geographic decision-making capabilities, with even the best-performing model, Gemini 3.7 Flash, achieving an overall success rate of only 27.8\%. Meanwhile, models exhibit distinct capability profiles across tasks. For example, Gemini 3.7 Flash achieves 37.1\% and 38.0\% on Grounding and Discovery, respectively, but only 19.4\% on Mobility Planning, compared with 31.4\% for GPT-5.6 Sol. Qwen3.8-Max performs best on Multi-Entity at 43.8\%, while Claude Sonnet 5 leads Safety-Aware at 19.8\%.

Task-level results further reveal substantial differences in the difficulty of geographic decisions. Grounding, Discovery, and Mobility Planning achieve comparable average success rates of 23.5\%, 24.1\%, and 22.0\%, respectively. 
Notably, Multi-Entity achieves the highest average success rate (30.8\%), despite requiring decisions that jointly satisfy constraints across multiple participants. 
In contrast, Safety-Aware, another compositional task category, achieves only 12.9\%. 
This disparity suggests that the difficulty of compositional geographic decisions depends not merely on combining multiple constraints, but also on the nature of the information and conditions required to establish decision validity. 
The most pronounced limitation emerges in Long-Horizon tasks, where the average success rate drops to 3.4\%, and even the best-performing model reaches only 8.7\%. 
These tasks require maintaining feasibility across interdependent decisions, where earlier choices constrain subsequent options. 
The sharp performance decline highlights the difficulty of producing end-to-end feasible plans, even when the underlying geographic state remains unchanged.

\begin{findingbox}{1}{Geographic decision-making limitations extend beyond satisfying explicit constraints.}
Agents perform relatively better on Multi-Entity tasks involving explicit joint constraints, but struggle when decisions must additionally account for safety-related conditions or remain feasible across interdependent steps. This contrast reveals an important limitation in extending constraint-based geographic decisions to more demanding validity requirements.
\end{findingbox}

\begin{figure*}[t]
    \centering
    \includegraphics[width=0.99\textwidth]{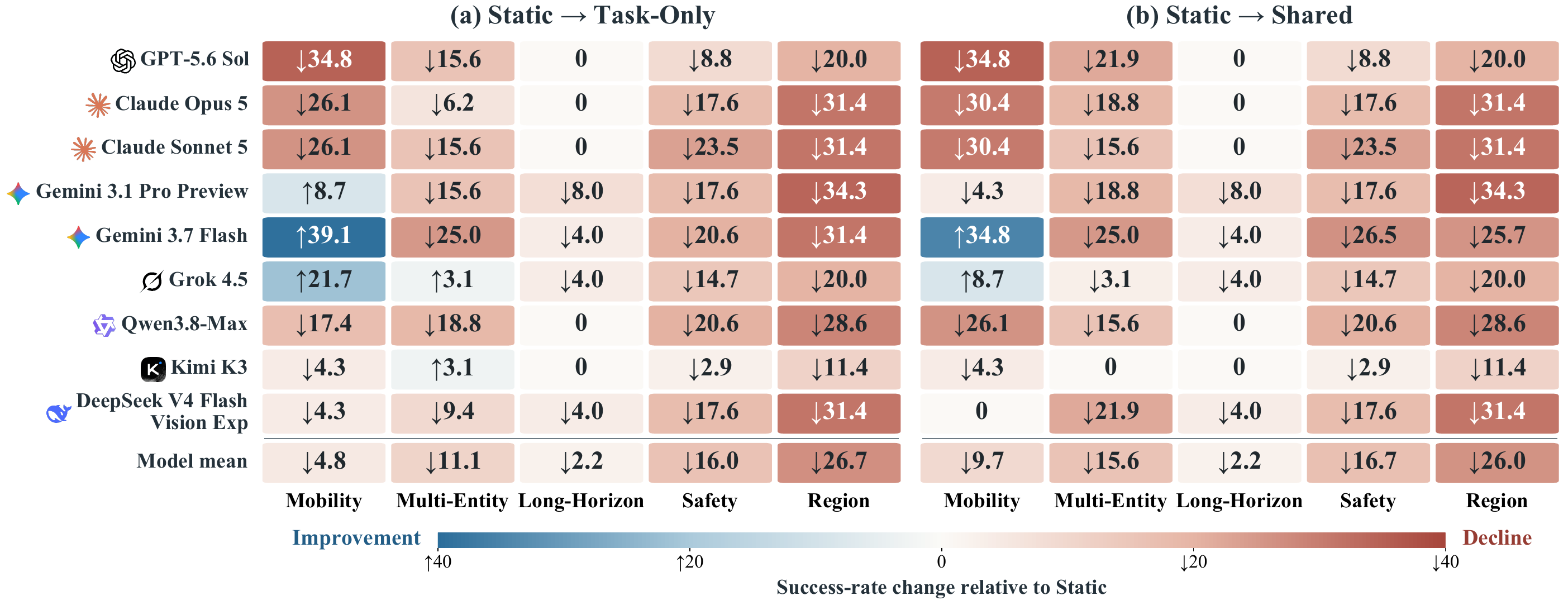}
    \caption{
        \textbf{Per-category change in success under world dynamics.}
        Change in success rate (percentage points) from Static to (a) Task-Only and (b) Shared Dynamics for each model and task category.
    }
    \label{fig:dynamic_model_bars}
\end{figure*}

\subsection{Impact of World Dynamics}
\label{sec:world_dynamics}

To examine how world dynamics affect geographic decision reliability, we evaluate nine models under Static, Task-Only Dynamics, and Shared Dynamics using 149 matched tasks. 
As shown in Table~\ref{tab:main_results}, mean success drops from 19.1\% under Static to 5.7\% and 4.0\% under Task-Only and Shared Dynamics, respectively, corresponding to relative declines of 70.3\% and 79.3\%. 
Notably, substantial degradation already occurs under Task-Only Dynamics, with a further decline under Shared Dynamics. 
However, the impact varies considerably across models. 
While Claude Sonnet 5 drops from 22.8\% under Static to 2.7\% and 2.0\% under the two dynamic settings, Gemini 3.7 Flash retains much higher success rates, declining from 26.2\% to 14.1\% and 13.4\%. 
These contrasting outcomes show that models with comparable static performance can differ substantially in their robustness to world changes. 
Figure~\ref{fig:dynamic_case} further illustrates that dynamic failures arise not only when agents miss updated geographic information, but also when they fail to incorporate correctly acquired updates into their final decisions. We systematically examine these behaviors in Section~\ref{sec:decision_maintenance}.

The impact of world dynamics also varies across geographic decision requirements. 
As shown in Figure~\ref{fig:dynamic_model_bars}, Region-Aware, Safety-Aware, and Multi-Entity experience average declines of 26.7, 16.0, and 11.1 percentage points under Task-Only Dynamics, respectively, compared with 26.0, 16.7, and 15.6 points under Shared Dynamics. 
In contrast, Mobility Planning shows smaller declines of 4.8 and 9.7 points, with three models even improving under Task-Only Dynamics. 
One possible explanation is that traffic changes may be captured through routine routing queries, whereas changes affecting regional restrictions, safety conditions, or participant constraints may require agents to identify relevant information needs across multiple constraints and reassess candidate validity.
The additional degradation of Multi-Entity under Shared Dynamics further suggests that different task requirements may be affected differently by shared-world changes. 
Long-Horizon remains near the performance floor across all settings, limiting meaningful comparisons of additional degradation. 
We further examine the types of world changes affecting different task categories in Appendix~\ref{sec:dynamic}.

\begin{findingbox}{2}{World evolution slashes success; static strength does not predict dynamic robustness.}
World changes cause substantial performance degradation, with markedly different levels of performance retention across models and task categories. This exposes vulnerabilities that are not captured by evaluating decisions against a fixed geographic state.
\end{findingbox}

\begin{figure*}[t]
    \centering

    \begin{subfigure}[t]{0.40\textwidth}
        \centering
        \includegraphics[width=\linewidth]{
            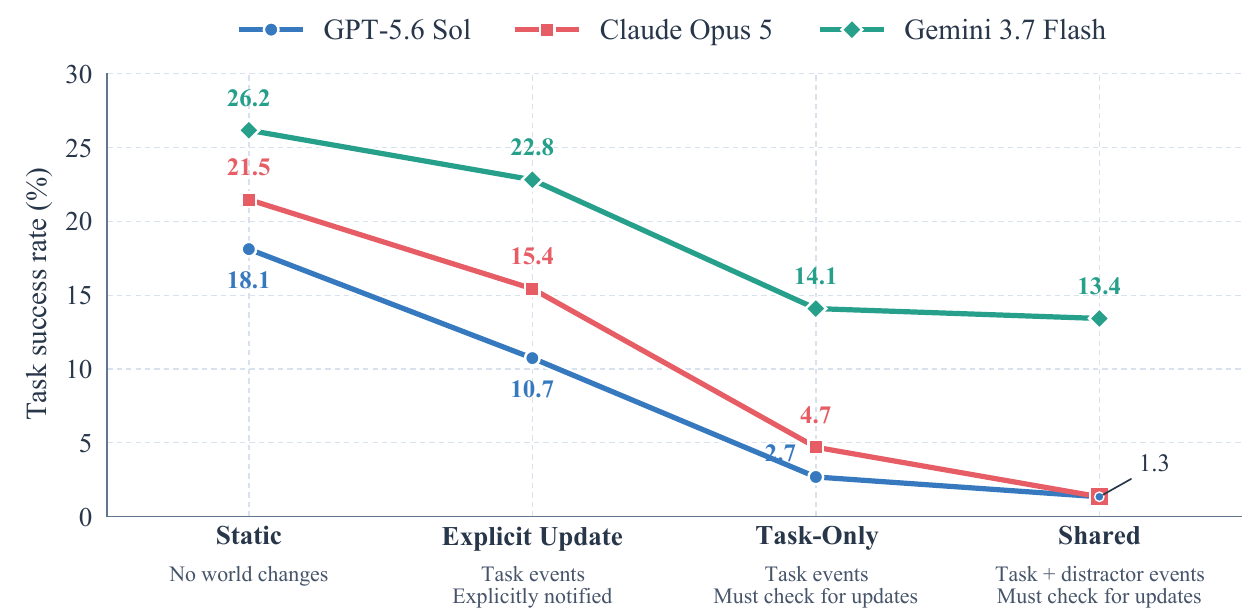
        }
        \caption{Success across dynamic settings}
        \label{fig:oracle_event_information}
    \end{subfigure}%
    \hfill
    \begin{subfigure}[t]{0.285\textwidth}
        \centering
        \includegraphics[width=\linewidth]{
            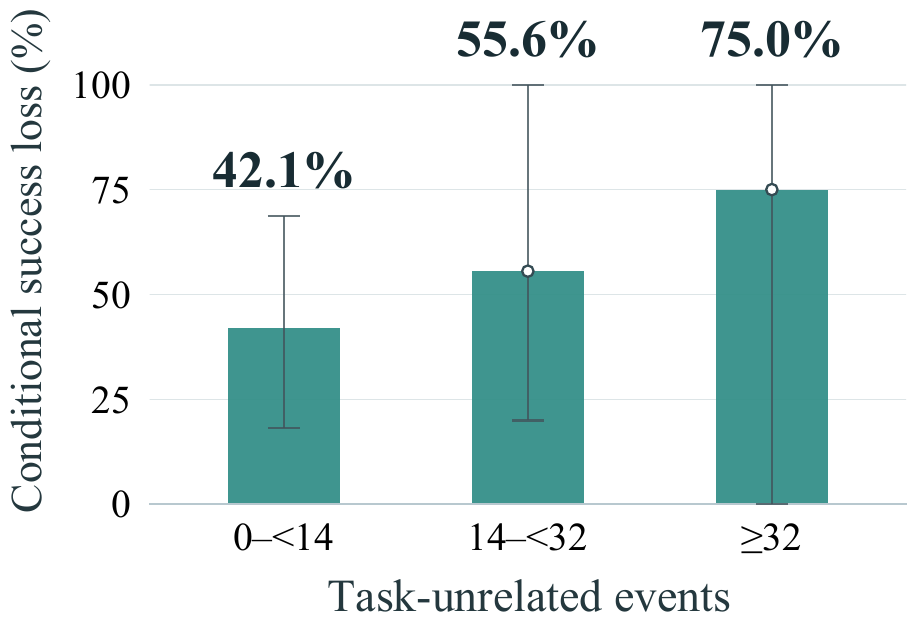
        }
        \caption{Impact of unrelated events}
        \label{fig:irrelevant_event_density}
    \end{subfigure}%
    \hfill
    \begin{subfigure}[t]{0.285\textwidth}
        \centering
        \includegraphics[width=\linewidth]{
            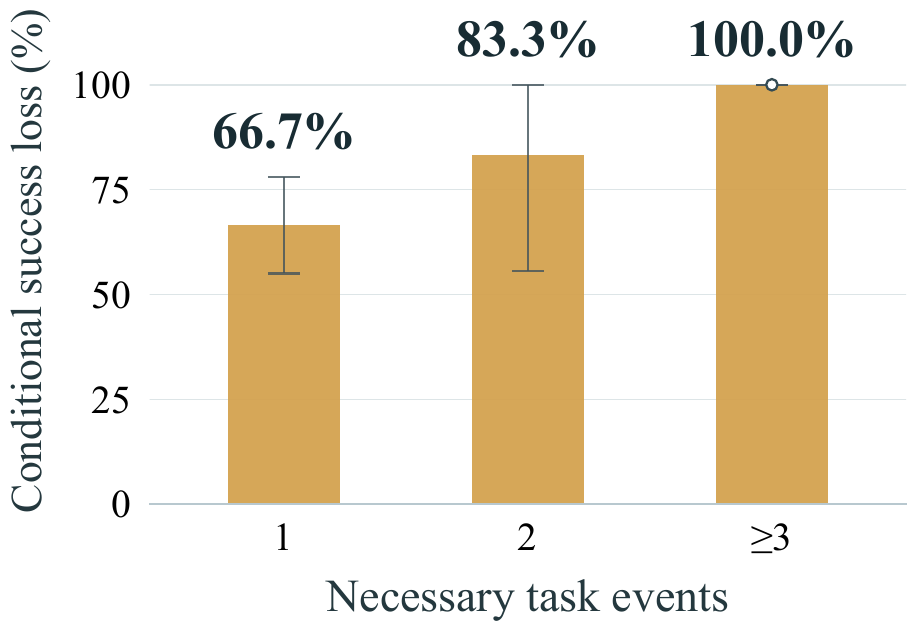
        }
        \caption{Impact of necessary events}
        \label{fig:necessary_event_density}
    \end{subfigure}

    \caption{
        \textbf{Isolating the sources of dynamic difficulty.}
        (a) Success of three models under Static, Explicit Update, Task-Only, and Shared Dynamics. Explicit Update hands the agent all necessary events; the two dynamic settings require it to discover them.
        (b) Shared failure rates among Task-Only-successful model--task pairs, grouped by task-unrelated event count.
        (c) Explicit Update failure rates among Static-successful pairs, grouped by necessary-event count.
    }
    \label{fig:additional_dynamics}
\end{figure*}

To further investigate how information acquisition affects dynamic decision-making, we evaluate three models under an additional \emph{Explicit Update} setting, where all necessary task-related events are directly provided. 
As shown in Figure~\ref{fig:additional_dynamics}(a), providing these updates does not restore static-level performance: the three models' mean success rate drops from 21.9\% under Static to 16.3\% under Explicit Update. 
This indicates that access to updated information alone is insufficient to ensure valid decisions under changed geographic conditions. 
When agents must discover the relevant updates themselves under Task-Only Dynamics, mean success drops further to 7.2\%. 
Notably, the relative decline from Explicit Update to Task-Only Dynamics is particularly pronounced for GPT-5.6 Sol (75.0\%) and Claude Opus 5 (69.6\%), compared with Gemini 3.7 Flash (38.2\%), suggesting substantial differences in sensitivity to information acquisition requirements.
Performance declines further to 5.4\% under Shared Dynamics, where task-relevant events occur alongside unrelated world changes.

To investigate the relationship between the performance declines observed above and world events, we examine how the number of necessary and unrelated events affect conditional success loss. 
As shown in Figure~\ref{fig:additional_dynamics}(c), among model–task pairs that succeed under Static, conditional success loss rises from 66.7\% to 100.0\% as the number of necessary events increases. 
This suggests that processing multiple updates makes it increasingly challenging for agents to maintain valid decisions. 
Similarly, Figure~\ref{fig:additional_dynamics}(b) shows that conditional success loss increases from 42.1\% to 75.0\% with the number of unrelated events, indicating that background world changes may further complicate decision maintenance.
Overall, the increase in necessary events is associated with more pronounced success loss than the increase in unrelated events, as agents must not only acquire all relevant updates but also understand their implications for existing decisions and correctly incorporate multiple updates into their decision-making. 
We further investigate these challenges in Section~\ref{sec:decision_maintenance}.

\begin{findingbox}{3}{Understanding, discovering, and isolating changes each cost success.}
Even when the agent receives every necessary update, success drops below Static; it drops again when updates must be discovered, and again when they must be separated from unrelated events. Moreover, success loss increases with the number of events, with a more pronounced increase associated with necessary events.
\end{findingbox}

\subsection{Decision Maintenance under World Changes}
\label{sec:decision_maintenance}

To investigate why agents struggle to maintain valid decisions in evolving geographic environments, we analyze state-aligned interaction trajectories from 9 models on 149 matched tasks, with three independent runs per model-task pair, yielding 4,023 executions under each dynamic setting. 
We examine two aspects of decision maintenance: whether agents acquire the information needed to track relevant world changes, and whether they correctly incorporate the observed changes into their final decisions.

\begin{figure*}[t]
    \centering
    \begin{minipage}[c]{0.34\textwidth}
        \centering
        \includegraphics[
            width=\linewidth,
            trim=0cm 0.1cm 0cm 0.1cm,
            clip
        ]{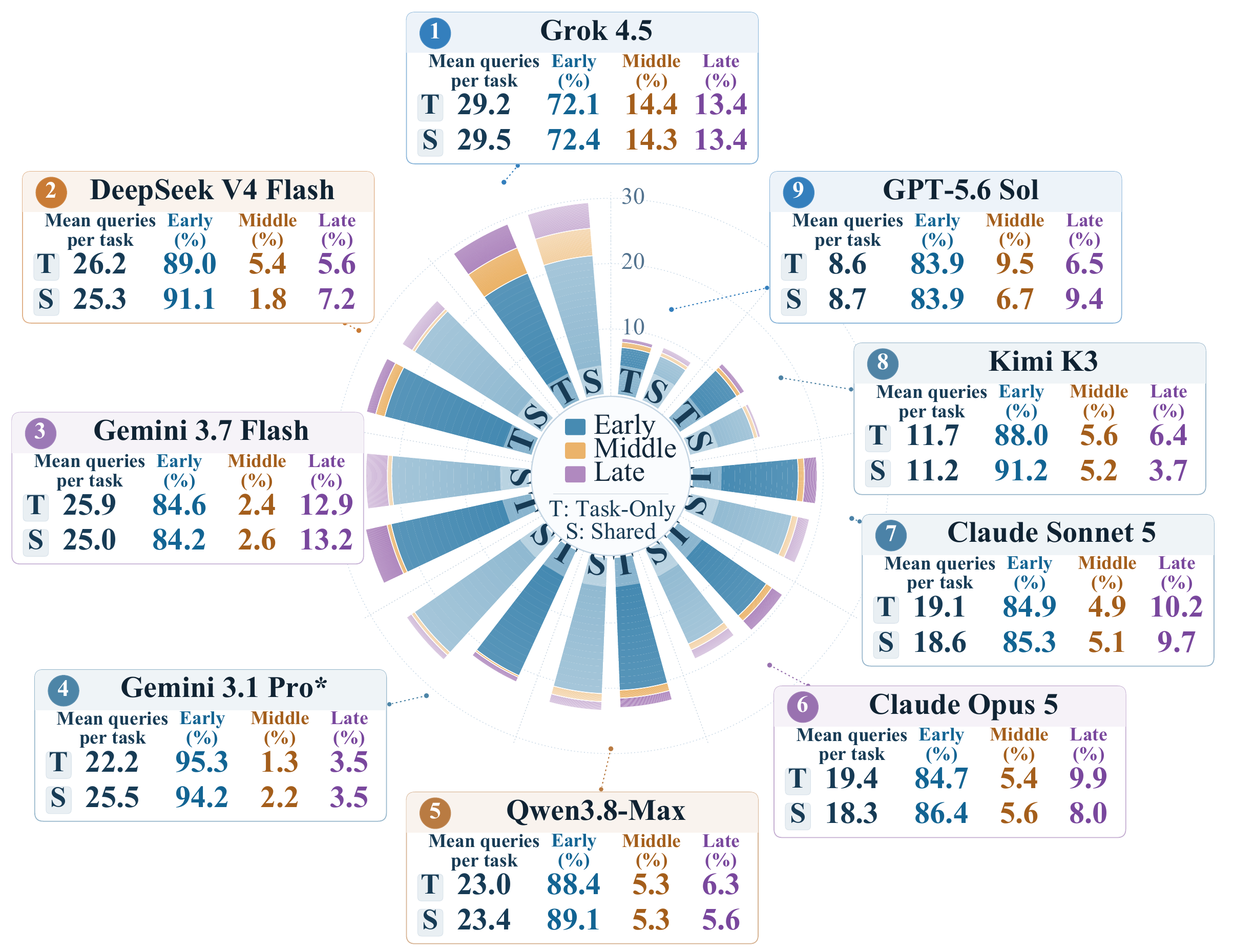}
        \par\vspace{2pt}
        {\small (a) Query timing distribution}
    \end{minipage}
    \hfill
    \begin{minipage}[c]{0.64\textwidth}
        \centering
        \includegraphics[
            width=\linewidth
        ]{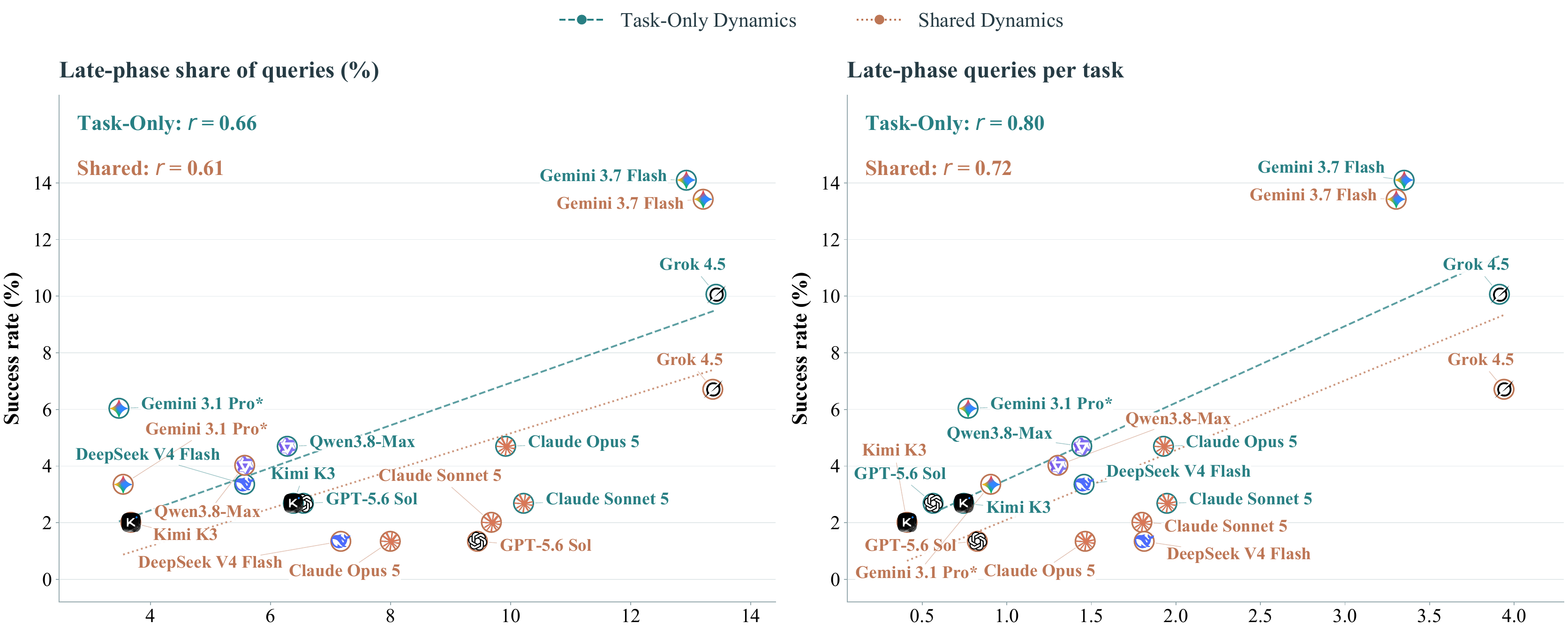}
        \par\vspace{2pt}
        {\small (b) Late-phase querying and task success}
    \end{minipage}

    \caption{
        \textbf{When agents query under world dynamics.}
        (a) Mean queries per task for nine models under Task-Only (T) and Shared (S) Dynamics, split into Early, Middle, and Late phases (equal thirds of the task's virtual-time window); the tables list each model's mean query count and phase shares.
        (b) Late-phase query share (left) and count (right) against dynamic success rate; each point is one model, with Pearson $r$ per panel.
    }
    \label{fig:dynamic_query}
\end{figure*}

\paragraph{When do agents query?}
As shown in Figure~\ref{fig:dynamic_query}(a), agents concentrate most information queries in the early stage of a task, with approximately 72–95\% of queries occurring in the first third of the virtual-time window. This pattern persists across both dynamic settings, despite substantial differences in query volume across models. 
For instance, Gemini 3.7 Flash issues nearly three times as many queries as GPT-5.6 Sol under Task-Only Dynamics (25.9 versus 8.6 per task), yet both concentrate over 80\% of their queries in the early phase. 
Shared Dynamics also induces no systematic shift toward later querying, suggesting that agents do not substantially adapt their information-seeking behavior to the more complex event environment. 
Nevertheless, Figure~\ref{fig:dynamic_query}(b) shows positive model-level correlations between late-phase querying and task success across both settings ($r=0.61$-$0.80$). 
Although these correlations do not establish causality, they motivate a closer examination of whether agents revisit information affected by world changes.

\begin{figure*}[t]
    \centering
    \includegraphics[width=\textwidth]{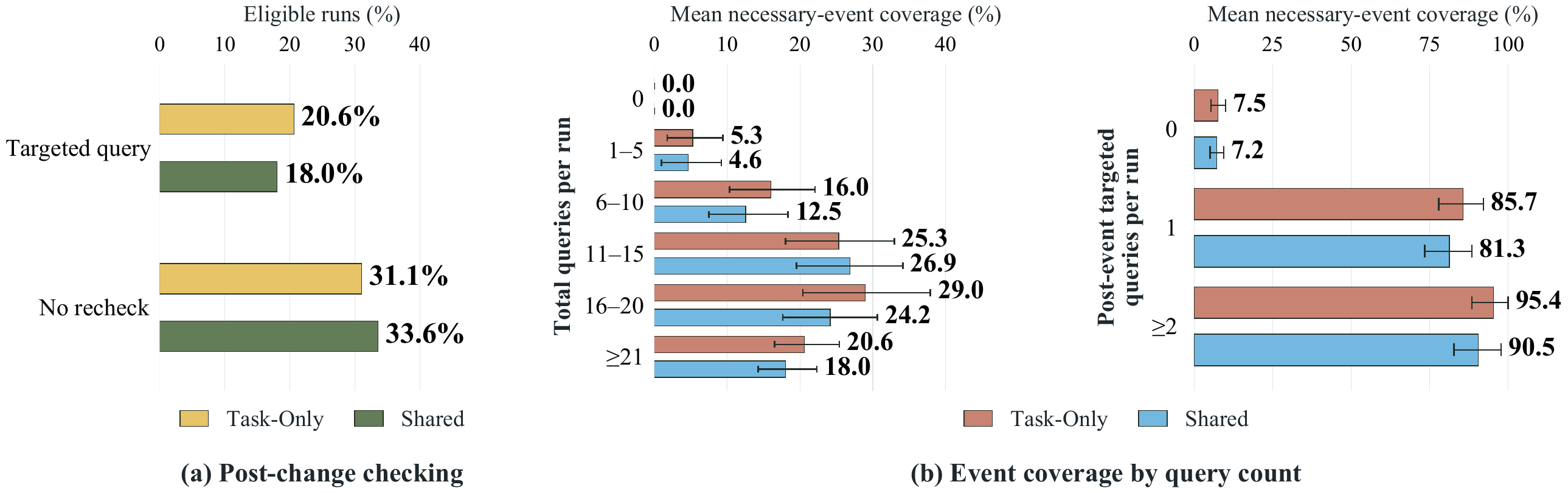}
    \caption{
        \textbf{Whether agents query what a change affects.}
        (a) Among runs that reach a necessary event, the share issuing a post-event targeted query (about an affected entity or updated constraint), and, among those that had consulted relevant information before the change, the share that never recheck it.
        (b) Mean percentage of necessary events observed per run, grouped by total queries (left) and by post-event targeted queries (right).
    }
    \label{fig:dynamic_bc}
\end{figure*}

\paragraph{Do agents track relevant changes?}
Querying throughout a task does not necessarily ensure that agents observe the changes relevant to their decisions. 
We therefore align queries with the effective times and affected entities of necessary events. 
As shown in Figure~\ref{fig:dynamic_bc}(a), only 20.6\% and 18.0\% of eligible runs issue targeted queries after necessary events under Task-Only and Shared Dynamics, respectively. 
Moreover, among runs that had previously acquired the affected information, 31.1\% and 33.6\% never recheck it after the change. Figure~\ref{fig:dynamic_bc}(b) further shows that issuing more queries does not consistently improve necessary-event coverage: coverage remains below 30\% across all total-query-count groups. 
In contrast, runs with at least one post-event targeted query exhibit substantially higher event coverage than those without such queries. 
While this association partly follows from how targeted queries and event observation are defined, it highlights the importance of reaching the information actually affected by a change, rather than simply increasing query volume.

\begin{findingbox}{4}{Agents fail to consistently track task-relevant world changes.}
Information queries are heavily concentrated at the beginning of tasks, with limited targeted checking after necessary events. Higher query volume alone does not ensure broader event coverage, exposing a gap between information-seeking activity and maintaining awareness of an evolving world.
\end{findingbox}

\begin{figure*}[t]
    \centering
    \begin{subfigure}[c]{0.73\textwidth}
        \centering
        \includegraphics[width=\linewidth]{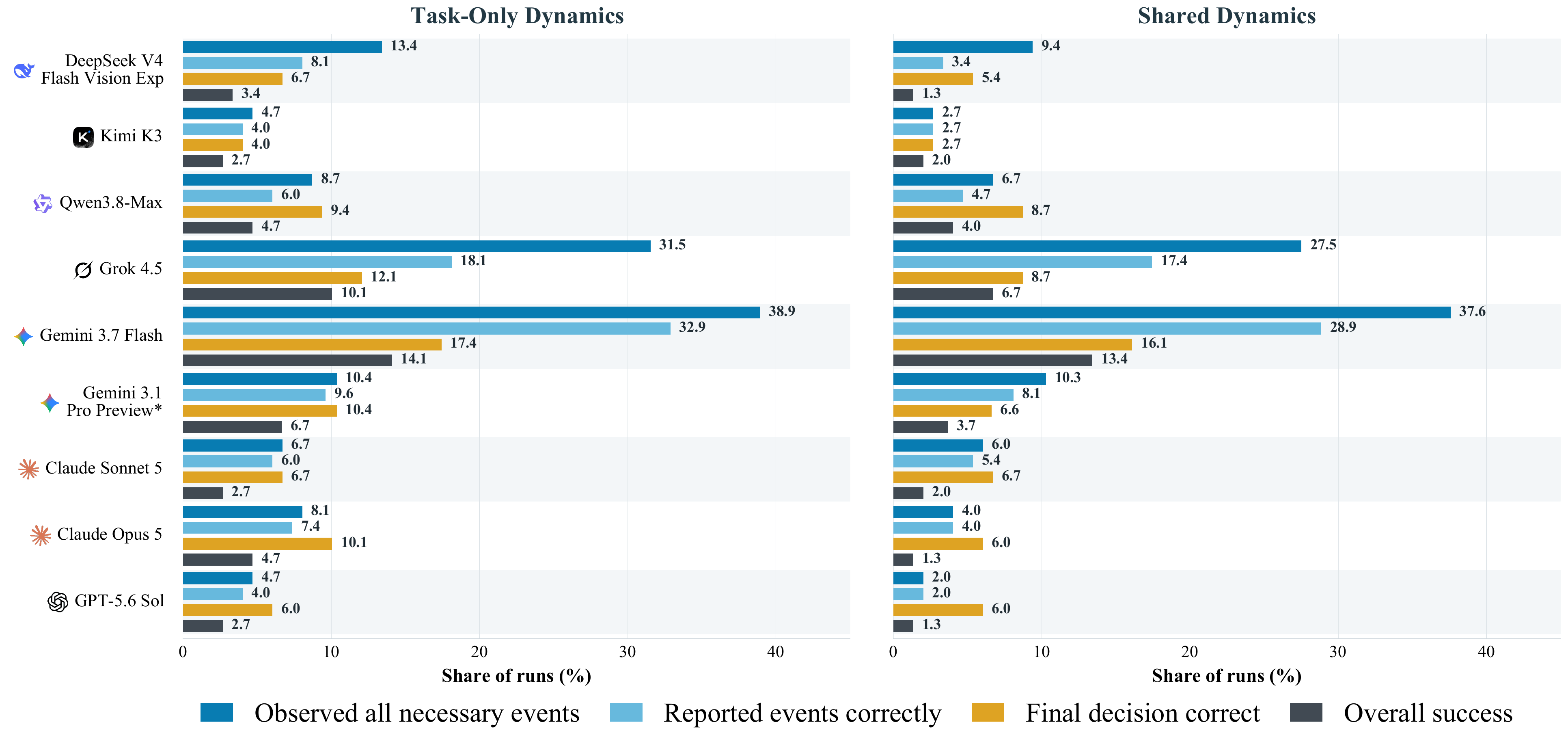}
        \caption{Observation, reporting, and decision outcomes}
        \label{fig:dynamic_observation}
    \end{subfigure}
    \hfill
    \begin{subfigure}[c]{0.225\textwidth}
        \centering
        \includegraphics[width=\linewidth]{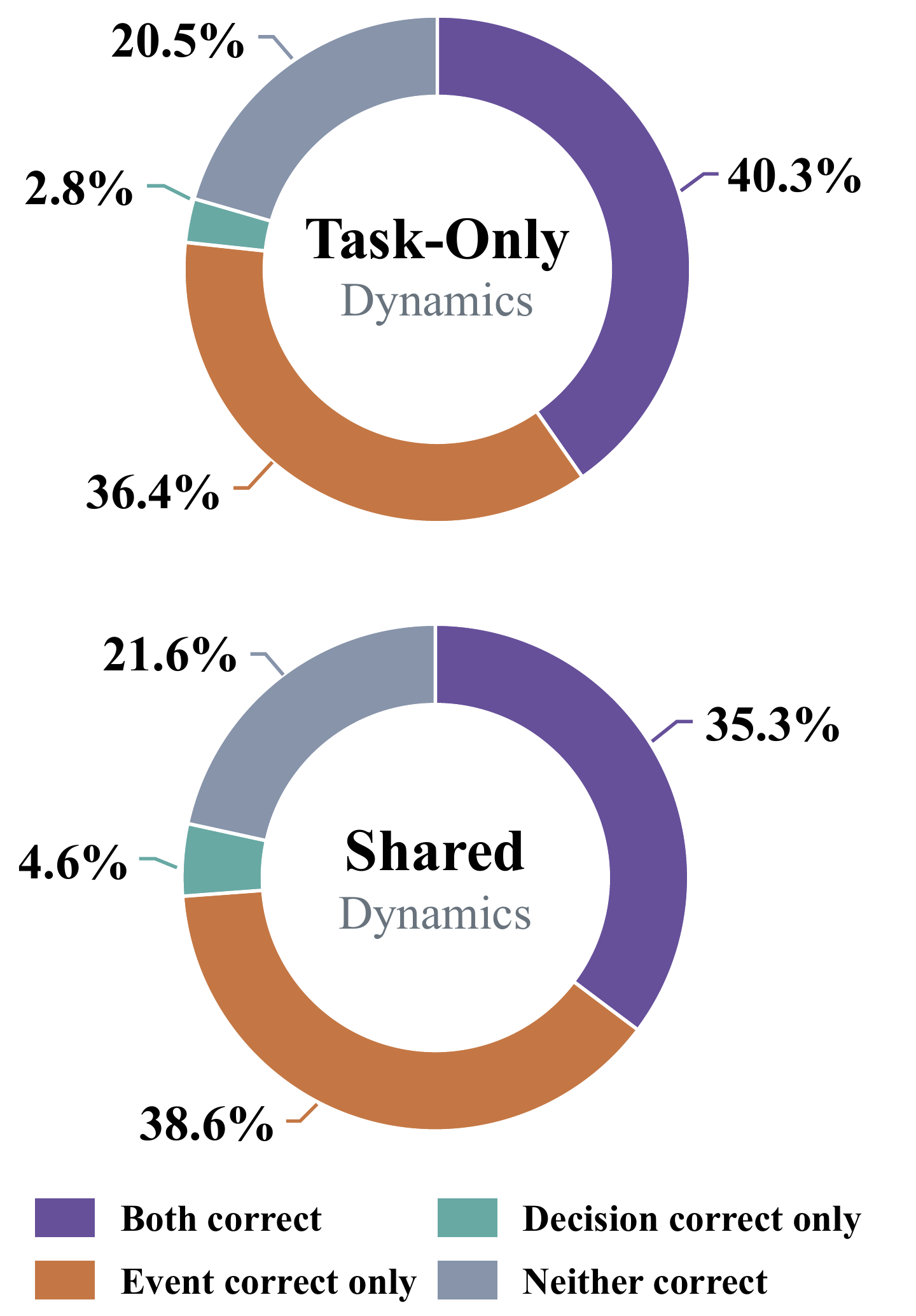}
        \caption{Decision grounding}
        \label{fig:decision_grounding}
    \end{subfigure}

    \caption{\textbf{From observing changes to correct decisions.}
        (a) Per model, the share of runs that observe all necessary events, report them correctly, reach a correct final decision, and succeed overall, under Task-Only and Shared Dynamics.
        (b) Among runs that observe all necessary events, the joint distribution of event-reporting and final-decision correctness.
    }
    \label{fig:dynamic_decision_maintenance}
\end{figure*}

\paragraph{Does observing changes lead to valid decisions?}
Even when agents acquire necessary updates, they may fail to incorporate them into their final decisions. 
Figure~\ref{fig:dynamic_decision_maintenance}(a) shows that only 14.1\% and 11.8\% of runs observe all necessary events under Task-Only and Shared Dynamics, respectively, with substantial variation across models. However, incomplete observation is only part of the problem. 
Among runs that observe every necessary event, Figure~\ref{fig:dynamic_decision_maintenance}(b) shows that 76.7\% and 73.9\% correctly report the events, whereas only 43.1\% and 39.9\% produce correct final decisions. 
Notably, 36.4\% and 38.6\% correctly report the relevant changes but still reach incorrect decisions. 
This reveals a distinct gap between acquiring accurate information about the current geographic state and using that information to satisfy the task constraints. 
Thus, failures in dynamic decision maintenance arise not only from incomplete awareness of world changes, but also from difficulties in translating observed changes into valid decisions.

\begin{findingbox}{5}{Awareness of world changes does not guarantee valid decisions.}
Even among runs that observe every necessary event, fewer than half produce correct final decisions, and over one-third correctly report the changes yet still decide incorrectly. Reliable decision maintenance therefore requires more than acquiring and recognizing updated geographic information.
\end{findingbox}

\begin{figure*}[t]
    \centering
    \includegraphics[width=\textwidth]{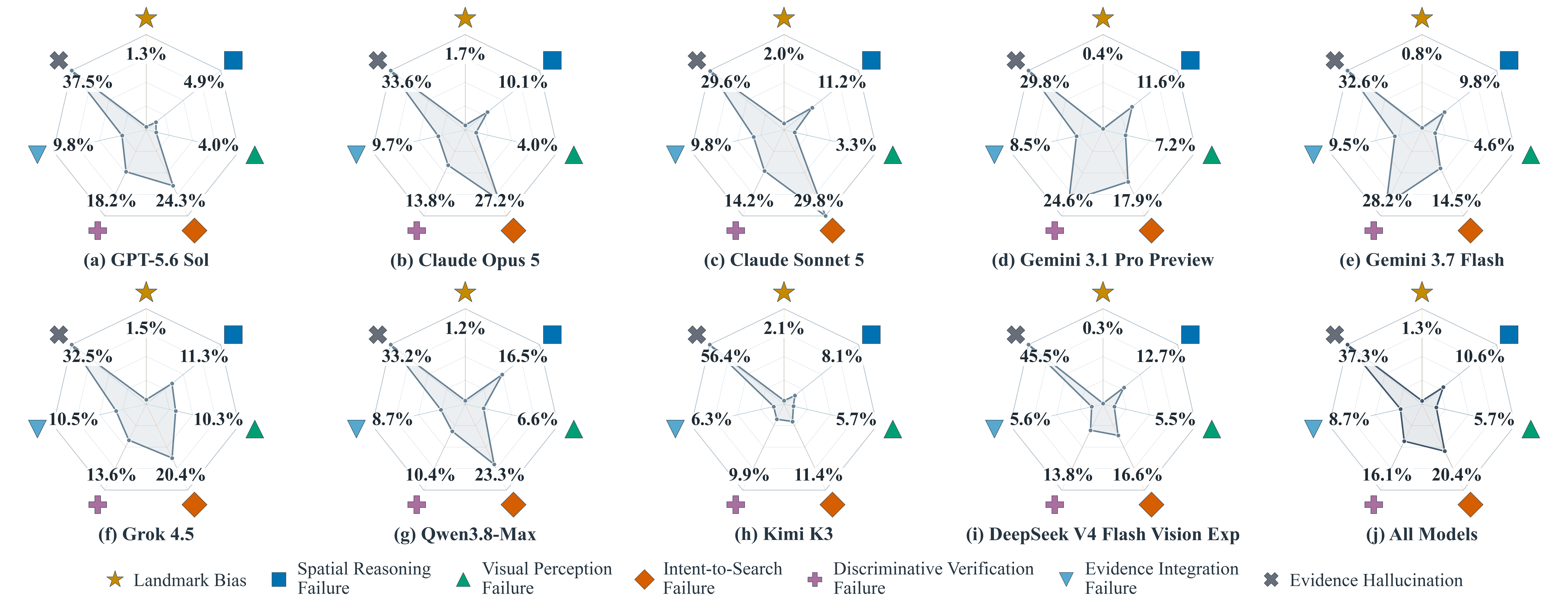}
    \caption{
        \textbf{Primary failure modes under the Static setting.}
        Share of seven failure modes among attributed failures, per model in (a)--(i) and aggregated over all models in (j).
    }
    \label{fig:failure_profiles}
\end{figure*}

\begin{figure*}[t]
    \centering
    \includegraphics[
        draft=false,
        width=0.99\textwidth,
        trim=0 1cm 0 0.1cm,
        clip
    ]{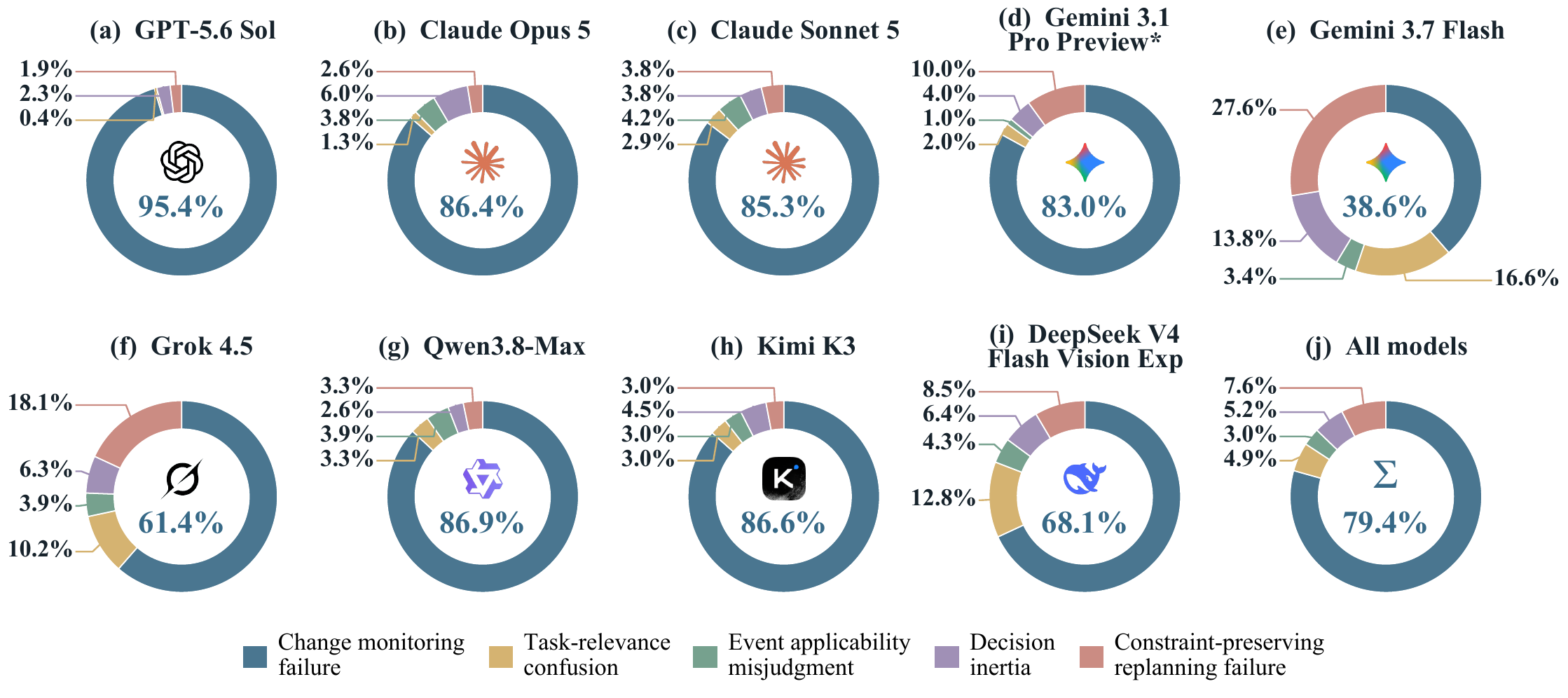}
    \caption{
        \textbf{Primary failure modes under world dynamics.}
        Share of five decision-maintenance failure modes among attributed failures, per model in (a)--(i) and aggregated over all models in the bottom-right panel.
    }
    \label{fig:dynamic_failure_modes}
\end{figure*}

\subsection{Failure Modes in Geographic Decision-Making}
\label{sec:failure_modes}

To identify where geographic decision-making breaks down, we analyze state-aligned interaction trajectories and attribute failed executions to their primary failure modes under static and dynamic settings. 
We employ a two-stage LLM-based attribution protocol, in which candidate causes are proposed and independently reviewed using trajectory and world-state evidence. 
Failure definitions, attribution procedures, and representative cases are provided in Appendices~\ref{app:failure_attribution} and~\ref{app:case_study}, with task-level analyses in Appendix~\ref{app:failure_analysis}.

\paragraph{Failure Modes under Static Conditions.}
As shown in Figure~\ref{fig:failure_profiles}, static geographic decision-making failures are predominantly associated with how agents acquire, verify, and integrate geographic evidence. 
\emph{Evidence Hallucination} is the most prevalent failure mode, accounting for 37.3\% of attributed failures, followed by \emph{Intent-to-Search Failure} (20.4\%) and \emph{Discriminative Verification Failure} (16.1\%). Together with \emph{Evidence Integration Failure} (8.7\%), these evidence-related categories account for 82.5\% of attributed failures, substantially exceeding those associated with spatial reasoning or visual perception. 
This suggests that producing geographically valid decisions is limited not only by spatial understanding, but also by the reliability of the evidence used to support candidate choices. 
Although evidence hallucination is the dominant failure mode for most models, their failure profiles differ. 
For instance, Claude Sonnet 5 exhibits comparable proportions of search failures (29.8\%) and evidence hallucination (29.6\%), whereas Gemini 3.7 Flash shows a substantial proportion of discriminative verification failures (28.2\%). 
These variations indicate that failures can arise at different stages of evidence-based decision-making, rather than reflecting a single shared weakness across models.

\paragraph{Failure Modes under World Dynamics.}
Under evolving geographic conditions, the failure distribution becomes strongly concentrated in \emph{Change Monitoring Failure}, which accounts for 79.4\% of primary attributions and is the largest failure category for every model (Figure~\ref{fig:dynamic_failure_modes}). 
This contrasts with the more distributed failure profiles under Static, indicating that maintaining awareness of relevant world changes is a dominant challenge in dynamic decision-making. 
This finding is consistent with the behavioral evidence in Section~\ref{sec:decision_maintenance}, where agents concentrate queries early and infrequently revisit affected entities after changes occur. 
However, the dominance of monitoring failures does not imply that subsequent decision revision is reliable. Because each failed execution receives a single primary attribution, earlier monitoring failures may obscure downstream difficulties. 
Indeed, the decision-maintenance analysis shows that agents frequently produce incorrect decisions even after observing and correctly reporting necessary changes. 
Together, these results identify two distinct limitations: insufficient monitoring of an evolving geographic world and unreliable incorporation of observed changes into subsequent decisions.

\begin{findingbox}{6}{Geographic decision failures arise from distinct evidence and state-maintenance bottlenecks.}
Under static conditions, failures are primarily associated with acquiring, verifying, and integrating geographic evidence. Under world dynamics, inadequate change monitoring dominates primary failure attributions, while decision revision remains unreliable even when relevant changes are observed.
\end{findingbox}

\section{Conclusion \& Discussion}

We introduced \textbf{GeoWorld}, a persistent geographic environment in which world states evolve through independently occurring events, and \textbf{LiveGeo}, a benchmark for evaluating agents' ability to make and maintain valid geographic decisions under static and dynamic conditions. 
By aligning agent interactions with verifiable world states, our framework enables not only systematic evaluation of decision outcomes but also trajectory-level analysis of how agents acquire information, respond to changes, and revise decisions. 
Experiments across nine models reveal substantial limitations in geographic decision-making, with performance varying across decision requirements and deteriorating markedly under world dynamics. 
Further behavioral and failure analyses identify distinct bottlenecks: under static conditions, agents struggle to acquire, verify, and integrate geographic evidence; under dynamic conditions, they frequently fail to monitor relevant changes and may still produce invalid decisions even after observing the necessary updates.

These findings highlight a fundamental distinction between \emph{establishing decision validity} and \emph{maintaining it over time}. Reliable geographic agents must go beyond making decisions from a snapshot of the world: they need to track the conditions on which their decisions depend, detect when those conditions change, and reassess affected choices while preserving other applicable constraints. 
GeoWorld and LiveGeo provide a verifiable testbed for studying these capabilities, supporting future research toward agents that can sustain valid decisions in continuously evolving geographic environments.

\bibliography{iclr2027_conference}
\bibliographystyle{iclr2027_conference}

\appendix
\newpage
\startcontents[appendix]

\section*{Appendix}
\printcontents[appendix]{}{1}{\setcounter{tocdepth}{2}}

\section{GeoWorld: Design and Implementation Details}
\label{app:geoworld}

\subsection{Geographic World Instantiation}
\label{app:world_representation}

Table~\ref{tab:geoworld_coverage} summarizes the geographic assets in the current GeoWorld evaluation version across six cities, including Regions, POIs, and cached route records. In GeoWorld, a Region is a functional urban area constructed around real-world locations and their associated roles. Each Region is described by its name, center coordinates, functional labels, supporting evidence, and provenance information, and is represented spatially by its center point. Candidate Regions are generated during city discovery and reviewed with reference to urban geographic structure, functional roles, and supporting map and web evidence. POIs are associated with Regions based on spatial proximity, using a maximum assignment distance of 3,500~m. This representation provides a common spatial reference for downstream POI retrieval, task constraints, and region-level events. POIs are primarily collected through the Google Places API~\citep{googlemaps2026}. GeoWorld performs Nearby Search and Text Search around Region centers, transportation corridor nodes, and corridor service areas, and uses Place Details to supplement place attributes. The search covers a predefined set of categories, such as dining, retail, accommodation, healthcare, and education. Candidate POIs are then subject to coordinate checks, city-boundary filtering, provider-identity deduplication, and quality filtering. Deduplication prioritizes provider identities such as Google \texttt{place\_id}, thereby preventing the same location from being included multiple times across different searches. For service attributes required by downstream tasks, GeoWorld further combines Place Details with publicly available web evidence to supplement opening information, service descriptions, menus, and other verifiable merchant information. Route assets are progressively constructed around urban transportation corridors and the origin--destination pairs required by downstream tasks. Routes are obtained through the Google Routes API~\citep{googlemaps2026}, using three travel modes: \texttt{DRIVE}, \texttt{WALK}, and \texttt{TRANSIT}. Each cached route record contains a normalized origin and destination, travel mode, distance, estimated duration, route geometry.

GeoWorld retains provenance and version information for its geographic assets to support factual traceability and reproducible evaluation. City and POI records include data sources, build time, generation method, provider identifiers, and their corresponding world and city versions. GeoWorld adopts a two-level versioning scheme consisting of a global world version and city-level versions, with each city's assets bound to a specific version. Route caches are treated as derived assets that can be extended as needed during construction and are fixed to a versioned snapshot before evaluation, ensuring that different tasks and repeated runs are based on a consistent geographic substrate.

\subsection{Temporal Dynamics and Runtime}
\label{app:temporal_dynamics}

\begin{figure*}[t]
    \centering
    \includegraphics[width=\textwidth]{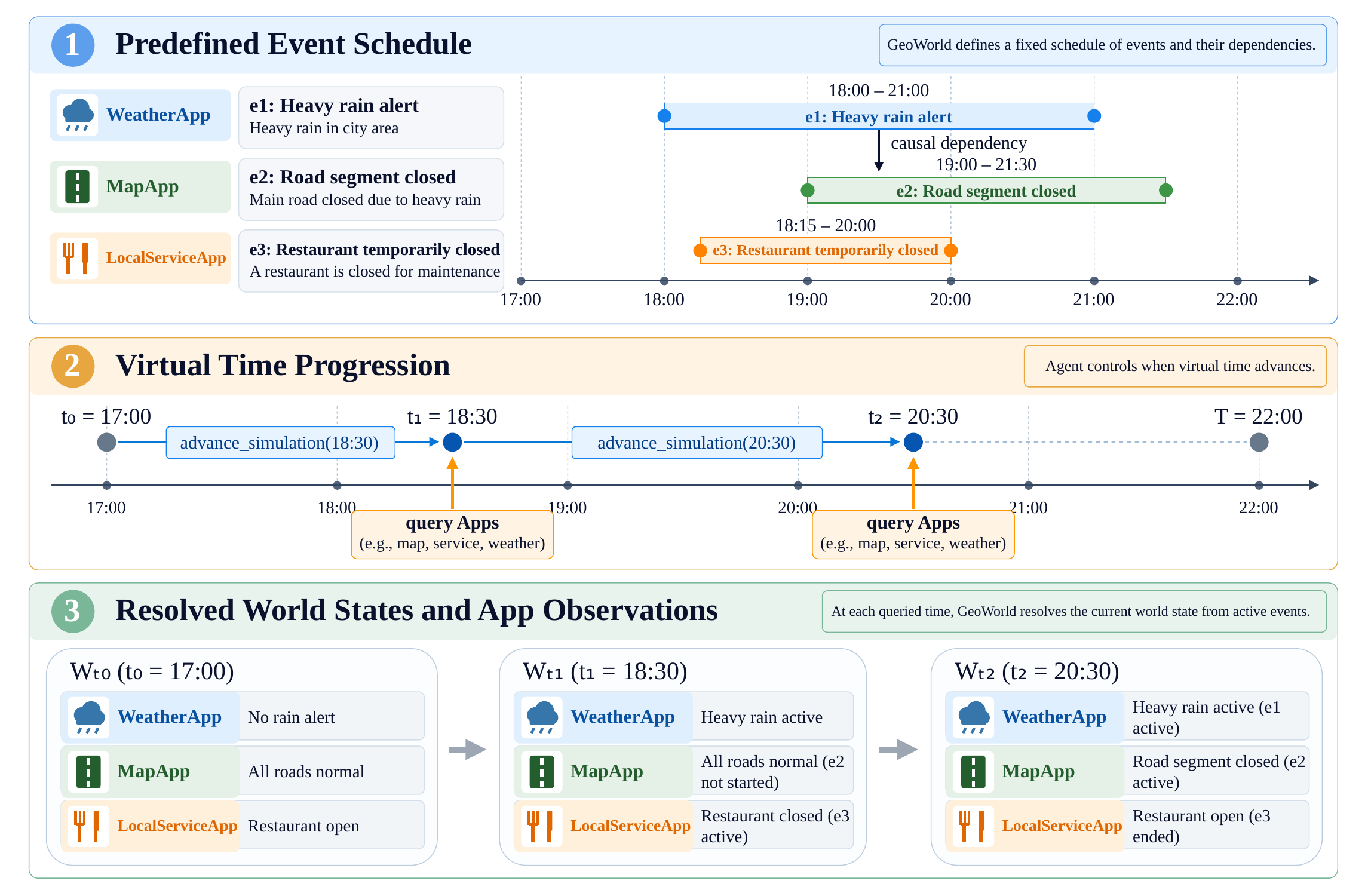}
    \caption{Example of temporal dynamics in GeoWorld. A predefined event schedule determines which state changes become active over time, while the Agent explicitly controls when virtual time advances and observes the resulting world state through Apps.}
    \label{fig:temporal_dynamics}
\end{figure*}

GeoWorld represents world changes as events occurring along a unified virtual timeline. Each event describes an external state change, including its occurrence time, the affected place or region, and the corresponding state update. Changes that persist for a period of time additionally record their effective interval, while changes with sequential relationships are linked through event dependencies. These events operate on the fixed geographic substrate, allowing the same place, route, or region to correspond to different service, traffic, weather, or regional states at different times. Complete event definitions are maintained internally by the environment, while the information observable to the Agent consists only of the public states and queryable information produced by events at the current time. At any virtual time, GeoWorld resolves the world state from events that have already occurred and remain effective. For events with causal relationships, a subsequent change enters the world state only after its prerequisite change has occurred. For example, an event defined as the consequence of another change does not appear simply because its scheduled time has been reached. Time therefore determines when a change becomes eligible to occur, while event dependencies ensure that related changes unfold according to their predefined relationships. For changes with a finite duration, their effects are removed from the current state after the effective interval ends, and the resulting state is determined by the underlying state together with other updates that remain effective. A dynamic episode begins from a predefined world state and virtual time. When the Agent wishes to continue the task, it can explicitly advance the virtual clock to a later absolute time. GeoWorld then processes the events that have taken effect during this interval and updates the environment to the world state to the target time. Subsequent queries are executed against that state, as illustrated in Figure~\ref{fig:temporal_dynamics}.

This separates temporal control from environment evolution. The Agent decides when to advance and re-observe the world, whereas the external changes that occur during the interval are determined by GeoWorld's predefined event schedule. For these exogenous changes, the Agent's queries and decisions do not alter their order or content. Different Agents can therefore face the same initial state and event schedule while independently deciding when to acquire updated world information. The fixed event schedule also allows the same dynamic process to be replayed across runs, providing controlled conditions for comparing Agent behavior under consistent environmental changes.

\begin{table}[t]
\centering
\caption{Geographic coverage of the GeoWorld instantiation.}
\label{tab:geoworld_coverage}
\begin{tabular}{lrrr}
\toprule
\textbf{City} & \textbf{Regions} & \textbf{POIs} & \textbf{Cached routes} \\
\midrule
Shanghai   & 8  & 290 & 21,357 \\
Tokyo      & 28 & 238 & 7,404 \\
Singapore  & 12 & 200 & 11,209 \\
London     & 16 & 212 & 23,267 \\
New York   & 12 & 216 & 7,295 \\
Cape Town  & 12 & 220 & 36,840 \\
\midrule
\textbf{Total} & \textbf{88} & \textbf{1,376} & \textbf{107,372} \\
\bottomrule
\end{tabular}
\end{table}
                                                                                        
\subsection{Tool Interfaces and Interaction Protocol}
\label{app:tool_interfaces}
GeoWorld provides agents with a unified interface through \textbf{8 public App namespaces and 31 tools}. Apps organize different types of world information and operations, while Tools are the concrete interfaces that agents can invoke within the corresponding Apps. The currently available Apps include \texttt{SystemApp}, \texttt{MapApp}, \texttt{LocalServiceApp}, \texttt{WeatherApp}, \texttt{CityNewsApp}, \texttt{CalendarApp}, \texttt{MessagingApp}, and \texttt{UserInterfaceApp}. At the beginning of a run, only the system entry point for opening Apps is typically available. After an agent opens an App, it obtains the tools and corresponding schemas available from that App for the current task. This organization enables tool capabilities to be progressively exposed according to task requirements without modifying the underlying world state. These Apps provide different views of the same runtime environment. Map and routing interfaces expose places, spatial relations, and mobility information; local-service, weather, and city-information interfaces provide current service and environmental conditions; calendar and messaging interfaces provide participant-related information and communication capabilities; and \texttt{UserInterfaceApp} handles plan submission, virtual-time advancement, and task termination. All queries are executed against the world state at the current virtual time. Consequently, invoking the same interface at different times may return different results as the world evolves.

GeoWorld further distinguishes agent operations from exogenous environmental changes. Some tools can update interaction states within the current run, such as sending messages, recording location selections, or submitting plans. \texttt{advance\_simulation} advances the virtual clock, after which GeoWorld applies the pre-scheduled events whose temporal conditions have been satisfied. The agent has read-only access to environmental states and cannot directly modify weather, traffic, predefined service states, or the event schedule through the available tools. The agent can therefore decide what information to query, which permitted operations to perform, and when to advance virtual time, while changes in the external environment remain governed by GeoWorld's event process. GeoWorld records the tool interaction trajectory, including tool names, invocation arguments, returned results, real execution timestamps, and tool execution latency. Model outputs and available token-usage information are also stored during each run. Dynamic episodes additionally maintain virtual-time and event logs, which are used to determine which environmental changes had taken effect when each query was issued. Tool trajectories and temporal event logs are aligned according to execution order. This makes it possible to reconstruct what information the agent queried, what observations it actually received, and which environmental state those observations corresponded to, providing auditable interaction records for subsequent task verification and behavioral analysis.

\subsection{World Inspection and Visualization Interface}
\label{app:world_inspection}

\begin{figure*}[t]
    \centering
    \includegraphics[width=\textwidth]{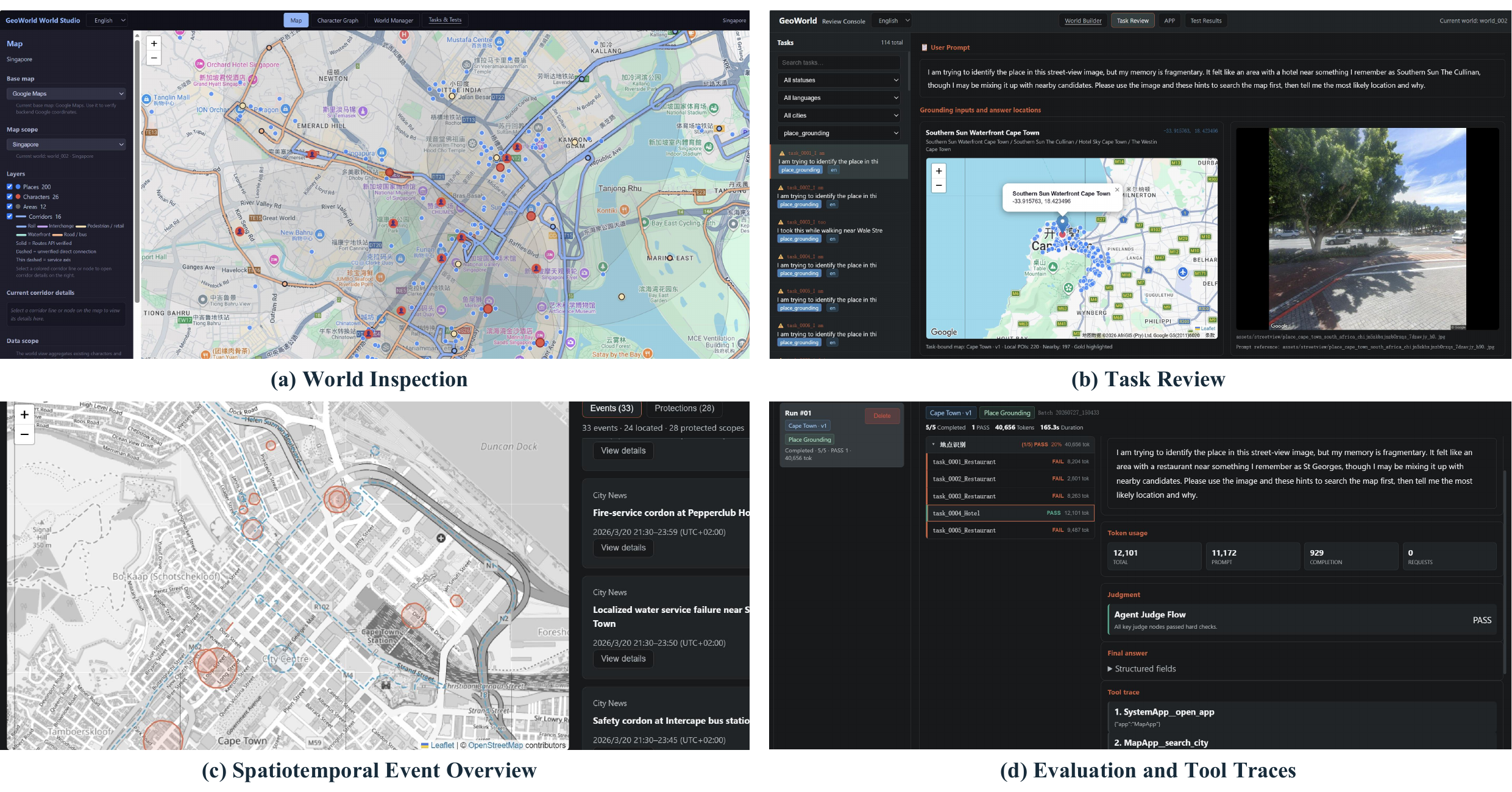}
    \caption{Example views of the GeoWorld inspection and review interface.}
    \label{fig:world_inspection}
\end{figure*}

GeoWorld provides a unified review interface for jointly inspecting tasks, world information, and model execution results, as shown in Figure~\ref{fig:world_inspection}. The review workspace integrates task review, App content inspection, and test results, while the world view provides the spatial and temporal information. The map displays POIs, transportation corridors and routes, task-relevant locations, participant positions, event-affected areas with spatial extents, and task protection areas, allowing reviewers to directly inspect the locations, mobility relations, and spatial constraints involved in a task.

For dynamic tasks, the interface further provides temporal and state information about events. Reviewers can inspect the event type, associated App, affected object or spatial scope, effective time, state update, and causal parents, and can filter events by time range. The event overview also presents their start and end times together with their spatial distribution. This information is used to verify whether the environmental changes relevant to a task are consistent with the corresponding locations, times, and constraints. The interface also supports inspection of saved agent trajectories, including tool calls, arguments, returned results, model inputs and outputs, usage information, and execution latency, allowing task content, world changes, and the information actually obtained by the agent to be examined within the same review workflow.

The interface primarily supports benchmark construction, quality review, and post-hoc analysis. During task construction, reviewers can inspect the relevant locations, routes, events, and constraints, and verify whether the reference answer is consistent with the underlying world information. After experiments, task information and saved tool trajectories can be examined together to analyze the agent's information acquisition and decision-making process. The review interface is primarily inspection-oriented, while allowing task prompts, reference answers, and constraints to be edited within the permitted scope and review comments to be recorded.

\section{LiveGeo: Benchmark Details}

\subsection{Task Taxonomy}
\label{app:task_taxonomy}

\paragraph{A dependency view of geographic decision-making.}
LiveGeo is organized around a simple premise: a geographic decision is valid only insofar as the geographic evidence and constraints on which it depends remain satisfied. We therefore structure the benchmark not by application scenario, but by the \emph{dependency structure} underlying a decision. The taxonomy progresses from decisions whose validity can be established locally, to decisions whose feasibility depends on several conditions holding jointly, and finally to settings in which those supporting conditions may themselves change over time. In this sense, the three levels of LiveGeo correspond to \emph{establishing}, \emph{composing}, and \emph{maintaining} geographic decision validity.

The static benchmark contains \textbf{1,088 instances}, including \textbf{478 Atomic Geo Tasks} and \textbf{610 Compositional Geo Tasks}. Their distribution is summarized in Table~\ref{tab:task_taxonomy_counts}. Dynamic Tasks are constructed as matched extensions of applicable static instances and are therefore counted separately.

All candidate instances are independently inspected by three human reviewers before inclusion in the benchmark. Reviewers check the instantiated entities and world state, task constraints, reference solution, supporting evidence, and, for dynamic tasks, the consistency of the event timeline. Inclusion requires unanimous reviewer approval. Table~\ref{tab:task_taxonomy_counts} reports the fraction of generated candidates that pass this review process. The consistently high acceptance rates indicate that the construction pipeline produces largely review-ready instances, while human inspection provides a final safeguard against residual factual or specification errors.

\begin{table}[t]
\centering
\small
\setlength{\tabcolsep}{7pt}
\caption{
Composition of the LiveGeo static benchmark and human-review acceptance rates.
Dynamic Tasks are matched extensions of applicable static instances.
}
\label{tab:task_taxonomy_counts}
\begin{tabular}{llrr}
\toprule
\textbf{Task group} & \textbf{Task category} & \textbf{Retained instances} & \textbf{Review pass} \\
\midrule
Atomic Geo Tasks
    & Grounding & 132 & 96.4\% \\
    & Discovery & 171 & 89.5\% \\
    & Mobility Planning & 175 & 92.6\% \\
\midrule
Compositional Geo Tasks
    & Multi-Entity & 176 & 88.0\% \\
    & Long-Horizon & 150 & 91.5\% \\
    & Safety-Aware & 121 & 98.4\% \\
    & Region-Aware & 163 & 94.8\% \\
\midrule
\textbf{Total} & & \textbf{1,088} & \textbf{92.5\%} \\
\midrule
Dynamic Tasks
    & Mobility Planning & 23 & 88.5\% \\
    & Multi-Entity & 32 & 84.2\% \\
    & Long-Horizon & 25 & 92.6\% \\
    & Safety-Aware & 34 & 89.5\% \\
    & Region-Aware & 35 & 92.1\% \\
\midrule
\textbf{Dynamic Total} & & \textbf{149} & \textbf{89.2\%} \\
\bottomrule
\end{tabular}

\smallskip
\end{table}

\paragraph{Atomic Geo Tasks: establishing local validity.}
At the atomic level, LiveGeo isolates three primitives that recur throughout geographic decision-making: resolving \emph{what place the available evidence refers to}, determining \emph{which place satisfies a stated need}, and deciding \emph{whether movement among places is feasible}. We refer to these primitives as \textbf{Grounding}, \textbf{Discovery}, and \textbf{Mobility Planning}, respectively. ``Atomic'' here describes the structure of the decision rather than the complexity of the trajectory: an instance may require multiple observations, searches, and tool calls, while still centering on a single dominant geographic dependency.

\textbf{Grounding} establishes the correspondence between observations and geographic entities. Street-level evidence rarely identifies the target in isolation: storefronts, road geometry, landmarks, signs, and neighboring places jointly constrain where the scene can be, while the most visually salient object need not be the entity being asked for. The task therefore requires the agent to move from perceptual clues to a geographically verified referent, separating evidence that merely locates the surrounding scene from evidence that identifies the requested place itself. This entity-level correspondence provides the anchor on which later geographic reasoning depends.

\begin{taskexample}{Grounding}
\noindent
\begin{minipage}[c]{0.42\linewidth}
    \centering
    \includegraphics[
        width=\linewidth,
        keepaspectratio
    ]{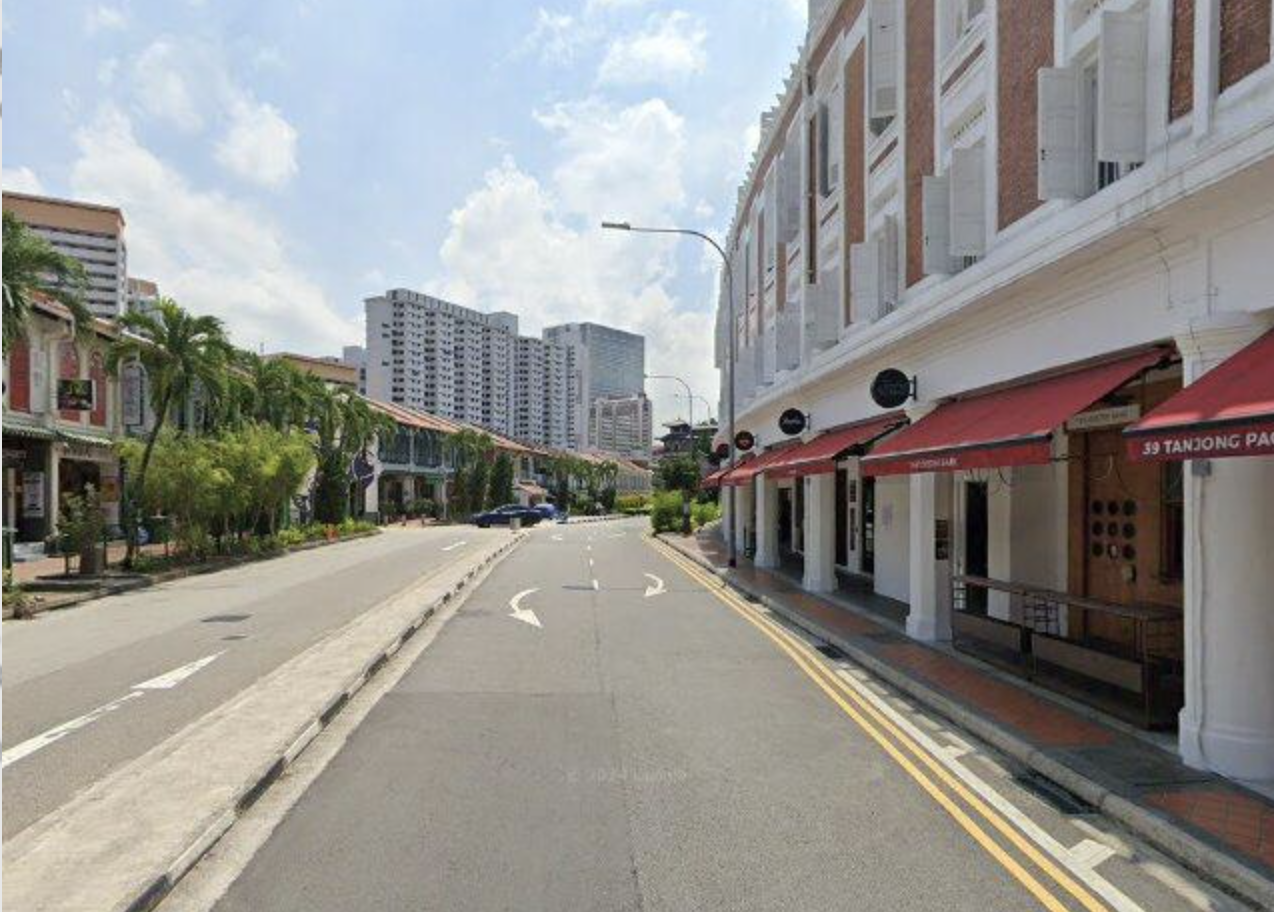}
\end{minipage}%
\hfill
\begin{minipage}[c]{0.54\linewidth}
    \justifying
    \textbf{Task:}
I am trying to identify the place in this street-view image, but my memory is fragmentary. It felt like an area with a coffee shop near something I remember as Night Hawk, though I may be mixing it up with nearby candidates. Please use the image and these hints to search the map first, then tell me the most likely location and why.
\end{minipage}
\end{taskexample}

\textbf{Discovery} shifts the problem from identifying a given place to choosing one. A user specifies a need and a set of spatial, temporal, or service-related conditions, but leaves the destination unresolved. Candidate retrieval alone does not settle the decision: many places may match the broad activity, while only a subset satisfy the particular conjunction of requirements. Discovery thus tests whether an agent can transform intent into a candidate space and, crucially, use the stated constraints to reduce that space to a defensible geographic choice.

\begin{taskexample}{Discovery}
\noindent
\textbf{Task:}
I’m blocking a short work break tomorrow near Midtown East, ideally around Second Ave, and need a café I can reserve or plan around without blowing a midrange budget. I’ll be starting from Sunglass Hut at 590 Fifth Avenue, and I won’t walk more than about ten minutes. I need vegetarian-friendly options and would rather avoid a transit-hall or retail-store café if a real coffee shop is close.
\end{taskexample}

\textbf{Mobility Planning} introduces relations among already identified places. Here, validity is no longer a property of a destination alone, but of a journey through space and time. Waypoints, travel modes, dwell times, visiting order, and deadlines interact cumulatively, so individually feasible route segments need not compose into a feasible trip. The benchmark consequently treats mobility as a constraint-satisfaction problem over the complete journey, including instances in which no arrangement satisfies all requirements. What matters is not recovering a preferred route, but correctly determining whether the requested movement is executable.

\begin{taskexample}{Mobility Planning}
\noindent
\textbf{Task:}
Lwazi Ndlovu\footnotemark{}
asks: ``Can I leave Victoria Wharf Shopping Centre,
stop at Grand Pavilion for a meeting lasting about
20 minutes, and reach GOLD Restaurant within
28 minutes in total?''
Please check the routes and help plan the journey.
\end{taskexample}

\footnotetext{All participant names in LiveGeo tasks
are fictional and generated by the GeoWorld simulator.}

Together, these three tasks define the local building blocks of LiveGeo: grounding establishes the entities under discussion, discovery chooses among them, and mobility planning determines whether they can be connected through feasible action.

\paragraph{Compositional Geo Tasks: when local validity is not enough.}
Real geographic decisions rarely remain isolated. A venue that is suitable for one person may fail for a group; an activity that fits one afternoon may break a three-day itinerary; an otherwise convenient destination may become inappropriate under hazardous weather or regional constraints. Compositional Geo Tasks are designed around precisely this gap between \emph{local plausibility} and \emph{global validity}. They retain the geographic primitives above, but introduce an additional dependency whose satisfaction can invalidate an otherwise reasonable local choice.

The four categories differ in \emph{where that dependency resides}. In \textbf{Multi-Entity} tasks, it lies across participants: locations, travel times, availability windows, and preferences must intersect at a shared activity. The relevant question is therefore not whether each participant can independently reach a venue, but whether their individual feasible sets admit a common solution.

\begin{taskexample}{Multi-Entity}
\noindent
\textbf{Task:} Please help Lin Weijie, Li Jiaying, Chen Huiling, and Xu Lihua find a restaurant where they can have a meal together. They will travel separately: Lin Weijie can leave no earlier than 18:32, Li Jiaying at 18:26, Chen Huiling at 18:34, and Xu Lihua at 18:29. Everyone must leave the restaurant by 20:30. Please coordinate their journeys so that they can spend at least 60 minutes dining together.

\end{taskexample}

In \textbf{Long-Horizon} tasks, the dependency instead propagates forward through a sequence of activities. Each decision changes the temporal and geographic state from which the next one must be made. A museum, restaurant, or meeting may be perfectly suitable in isolation yet render a later commitment unreachable. The resulting challenge is to preserve feasibility across a chain of decisions, rather than repeatedly optimizing each step in isolation.

\begin{taskexample}{Long-Horizon}
\noindent
\textbf{Task:}
Please build me a 3-day London itinerary using public transit where practical. I’m Eleanor Finch, staying at Bankside Hotel, Autograph Collection in South Bank. Monday: I arrive at Oxford Circus at 10:00 and will be there for about an hour. After that, I’d like to go for brunch/coffee at a place in King’s Cross that works as both a restaurant and a cafe. Before my 14:00 check-in at Bankside Hotel, Autograph Collection, I’d like to visit a museum in Covent Garden that is about an 8-minute drive from the hotel, and definitely no more than around 11 minutes by car. After checking in, add time at a public park whose primary type is simply a park, then send me to Paddington for dinner at a Mexican restaurant. I’d like to finish the evening back near the hotel with drinks at a pub in Bankside. Tuesday: Start with breakfast at a bakery that also functions as a restaurant/cafe. Then plan a relaxed visit to a garden-type park where I can walk around and sketch. For lunch, I want a cafe/restaurant in Paddington where the price is at most £10. After that, I’d like to visit a museum in Paddington whose primary type is museum. For dinner, take me to a food-court-style restaurant in Covent Garden. End the evening at a department store where I can do some shopping and also have a drink at a bar. Wednesday: Before checking out, I’d like breakfast at an Indian restaurant in Paddington that also works as a cafe/restaurant, with time to settle the breakfast bill and make a few sketch notes. I check out from Bankside Hotel, Autograph Collection at 10:30. After checkout, fit in a tourist attraction that is also a museum, then make sure I get to Paddington Station for my 14:00 departure.
\end{taskexample}

\textbf{Safety-Aware} tasks place the additional dependency in the surrounding environment. Geographic suitability is conditional on the state of the world at the relevant place and time: weather, hazards, or other safety-relevant conditions may remove an otherwise valid option from the feasible set. Importantly, LiveGeo does not infer safety awareness merely from the final answer. The interaction trajectory must show that the relevant environmental evidence was acquired, localized to the activity or access route, and allowed to affect the decision.

\begin{taskexample}{Safety-Aware}
\noindent
\textbf{Task:}
I’m Naledi Khumalo, a master’s student at the University of Cape Town upper campus in Rondebosch. I need lodging late at night on 20 March 2026, around 11:00 PM, starting from the University of Cape Town in Cape Town. Please find a suitable hotel that is no more than about a 12-minute drive from UCT. Choose an option that is safe and practically executable at that time of night. Return the selected hotel’s name and location, the relevant safety information you found, and explain why you chose it.
\end{taskexample}

\begin{safetyevidence}{Safety Evidence}
\noindent
\begin{minipage}[t]{0.47\linewidth}
    \textbf{Event 1:} A burst internal water service line has disrupted essential hotel water supply in an approximate simulated affected area around Southern Sun Waterfront Cape Town. The disruption is active from 21:30 to 23:50 on 20 March 2026, and lodging operations in the affected premises are temporarily unavailable while repairs and safety checks are carried out.
\end{minipage}
\hfill
\begin{minipage}[t]{0.47\linewidth}
    \textbf{Event 2:} A localized electrical switchgear fault has interrupted essential building services in an approximate simulated affected area around The Capital Mirage Cape Town. The disruption is active from 22:00 to 23:55 on 20 March 2026, and lodging operations in the affected premises are temporarily unavailable until power systems are isolated and checked.
\end{minipage}
\end{safetyevidence}

Finally, \textbf{Region-Aware} tasks introduce dependencies that arise from geographic context. Places that are indistinguishable under generic service or accessibility criteria may cease to be equivalent once local conventions, schedules, practices, or other region-specific conditions are taken into account. The benchmark therefore asks whether regional understanding becomes operational: it must alter which choices remain valid, rather than appearing as contextual knowledge in the explanation.

\begin{taskexample}{Region-Aware}
\noindent
\textbf{Task:}
I’m Mpho Daniels, in Cape Town at Granger Bay Campus – CPUT. I’m looking for an aquarium that offers scuba-diving opportunities in a large tank rather than the open ocean. Please return the selected place’s name and location, confirming that it is in the Victoria \& Alfred Waterfront district.
\end{taskexample}

\begin{safetyevidence}{Regional Evidence (Wikivoyage)}
\noindent
The Two Oceans Aquarium in Cape Town offers Scuba diving opportunities in its 2.2 million-litre tank for those who prefer not to dive in the open ocean.
\end{safetyevidence}

Viewed together, these categories correspond to four qualitatively different ways in which an apparently valid local decision can fail: because it conflicts with \emph{another entity}, with a \emph{later decision}, with the \emph{current environment}, or with the \emph{regional context}. This is the central distinction between atomic and compositional evaluation in LiveGeo. The latter is not intended simply to produce longer or more complicated prompts; it tests whether agents can identify and preserve the dependencies that make a collection of local choices jointly executable.

\paragraph{Dynamic Tasks: from decision validity to decision maintenance.}
The distinction above still assumes that the evidence supporting a decision remains fixed while the task is being solved. Dynamic Tasks remove this assumption. They are constructed from Mobility Planning and the four compositional categories using matched static--dynamic pairs that share the same objective, initial constraints, and initial world state. What changes is not the task, but the world in which the task continues to unfold.

This design shifts the object of evaluation. In a static task, the central question is whether the agent can establish that a decision is valid under the observed state. In a dynamic task, the question becomes whether the agent can preserve that validity as the supporting state evolves. A traffic disruption may invalidate a timing constraint; a place-state update may eliminate a selected destination; changing weather may alter the safety of an activity; or a participant's availability may remove the overlap on which a group plan relied. Conversely, many changes may occur without touching any dependency of the current decision.

Reliable behavior therefore requires more than reacting to change. The agent must determine \emph{which} changes intersect the dependency structure of the current plan, revalidate only the decisions whose support has been affected, and revise them without unnecessarily disturbing constraints that still hold. A change that does not alter the basis of the decision should leave the decision intact; a change that does must propagate to the appropriate downstream choices.

Dynamic LiveGeo thus evaluates \emph{selective decision maintenance}: the ability to keep a geographic decision aligned with the evidence that justifies it as the world evolves. In combination with the static taxonomy, this yields the full progression studied by the benchmark---from establishing individual geographic relations, to composing them into executable decisions, to maintaining those decisions when their supporting conditions no longer remain fixed.

\begin{taskexample}{Dynamic Task -- Initial Request}
\noindent
\textbf{Task:}
I'm Naledi Khumalo, a master's student at UCT upper campus. Starting from the University of Cape Town in Rondebosch, Cape Town, at about 07:30 on 20 March 2026, please help me choose a suitable restaurant. The place should support spending around ZAR 250. Please recommend a safe and practically reachable option for the circumstances, and return the selected place's name and location, the relevant safety information you obtained, and explain why you chose it.
\end{taskexample}

\begin{safetyevidence}{Task-Relevant Events}
\noindent
\textbf{06:05 | LocalServiceApp:} Harbour Bridge Restaurant on the Foreshore is closed this morning, Friday 20 March, for urgent kitchen maintenance and will reopen at 12:00. Sorry for the short notice.
\tcblower
\textbf{06:30 | LocalServiceApp:} Honey \& Dora Rooftop is opening early this Friday, 20 March, from 06:30 to 11:00 for a rooftop breakfast service. Coffee, light plates and walk-ins welcome.
\end{safetyevidence}

\begin{otherevent}{Other World Events (Selected from the Shared Timeline)}
\noindent
\textbf{06:30 | LocalServiceApp:} Notice from Dentist on Main Rondebosch, 41 Main Rd, Rondebosch: the practice is closed for the whole of today, Friday 20 March 2026, because of an unexpected staff shortage. No appointments and no walk-in treatment are available today; booked patients are being contacted to reschedule. We are sorry for the inconvenience.
\tcblower
\textbf{07:00 | LocalServiceApp:} Notice from Dis-Chem Claremont Pharmacy, Standard Bank Galleria, 178-182 Main Rd, Claremont: we are opening our doors early today, Friday 20 March 2026, at 07:00 instead of our usual 09:00, for a community health and medicine morning. From 07:00 our pharmacists will be available for prescriptions, medicine advice and basic health checks, and normal trading continues afterwards. Please come in.
\end{otherevent}

\subsection{Task Construction Details}
\label{app:construction_pipeline}

LiveGeo constructs geographic tasks through a capability-driven task construction workflow grounded in GeoWorld. GeoWorld provides the geographic entities, attributes, routes, and world states needed to define and verify task requirements, while task-specific construction logic determines how these resources are used to instantiate different geographic capabilities. By separating the underlying world from task construction, the same environment can support new task types without requiring changes to its geographic infrastructure.

\subsubsection{GT Selection}

\paragraph{Static GT selection.}
All LiveGeo tasks are constructed from a fixed version of GeoWorld, which provides real-world POIs, their attributes, route information, and other geographic facts.
Based on these facts, we determine the ground truth (GT) for each task through task-specific procedures. For Grounding, Discovery, Multi-Entity, Safety-Aware, and Region-Aware tasks, GT is selected programmatically from real POIs according to the requirements of each task category. For Long-Horizon tasks, an LLM suggests the types of activities to include, while code selects the corresponding real POIs to form an ordered sequence as GT. Mobility Planning follows a different procedure: code first selects the locations and constructs travel constraints, and a solver then determines whether the trip is feasible, producing a Boolean GT.

Importantly, selecting a GT does not mean that it is already a verified answer.
For POI-based tasks, the selected GT serves as the starting point for constructing constraints $C_0$ using facts from GeoWorld. We then verify whether GT constitutes a feasible solution under the constructed constraints and whether these constraints are supported by world facts. Only tasks that pass the subsequent verification procedures are retained in LiveGeo. Table~\ref{tab:gt_selection} summarizes the GT selection procedures across task categories, together with illustrative examples.

\paragraph{Dynamic GT selection.}
For dynamic tasks, we start from a verified static task with its initial GT $GT_0$.
Before constructing dynamic events, we select a later GT $GT_1$ that the events are intended to make valid. For POI-based tasks, an LLM may suggest the direction of change, while code selects $GT_1$ from real POIs in GeoWorld. For example, in a Multi-Entity task in London, the initial GT is Bonne Bouche Paddington ($GT_0$), while the later GT is Hayat Paddington ($GT_1$). A temporary closure of the original venue is then introduced as an event, and subsequent verification checks whether the updated world state supports $GT_1$. As with static tasks, the selected GT becomes a verified reference answer only after the corresponding construction and verification procedures.

\begin{table}[t]
\centering
\small
\renewcommand{\arraystretch}{1.15}
\caption{GT selection across LiveGeo task categories, with illustrative examples.}
\label{tab:gt_selection}
\begin{tabularx}{\textwidth}{lXX}
\toprule
\textbf{Task category} & \textbf{GT selection} & \textbf{Example GT} \\
\midrule

Grounding
& Sample a target POI from GeoWorld.
& Hackett London -- Regent Street (London). \\

Discovery
& Sample a target POI from GeoWorld.
& Bread Street Kitchen \& Bar (London). \\

Mobility Planning
& Compute a Boolean GT using route facts and a solver.
& True (feasible journey, London). \\

Multi-Entity
& Sample a POI as the shared activity location.
& Shinjuku Takashimaya (Tokyo). \\

Long-Horizon
& Use an LLM to suggest activity types and code to select real POIs in sequence.
& Tokyo Skytree Town $\rightarrow$ Shinjuku Takashimaya $\rightarrow$ Onigiri Bongo (partial). \\

Safety-Aware
& Set the task time and sample an open POI.
& President Hotel (London). \\

Region-Aware
& Sample a POI from candidates matching a regional criterion.
& Royal Opera House (London). \\

\bottomrule
\end{tabularx}
\end{table}

\begin{figure}[t]
\centering
\includegraphics[width=0.99\linewidth]{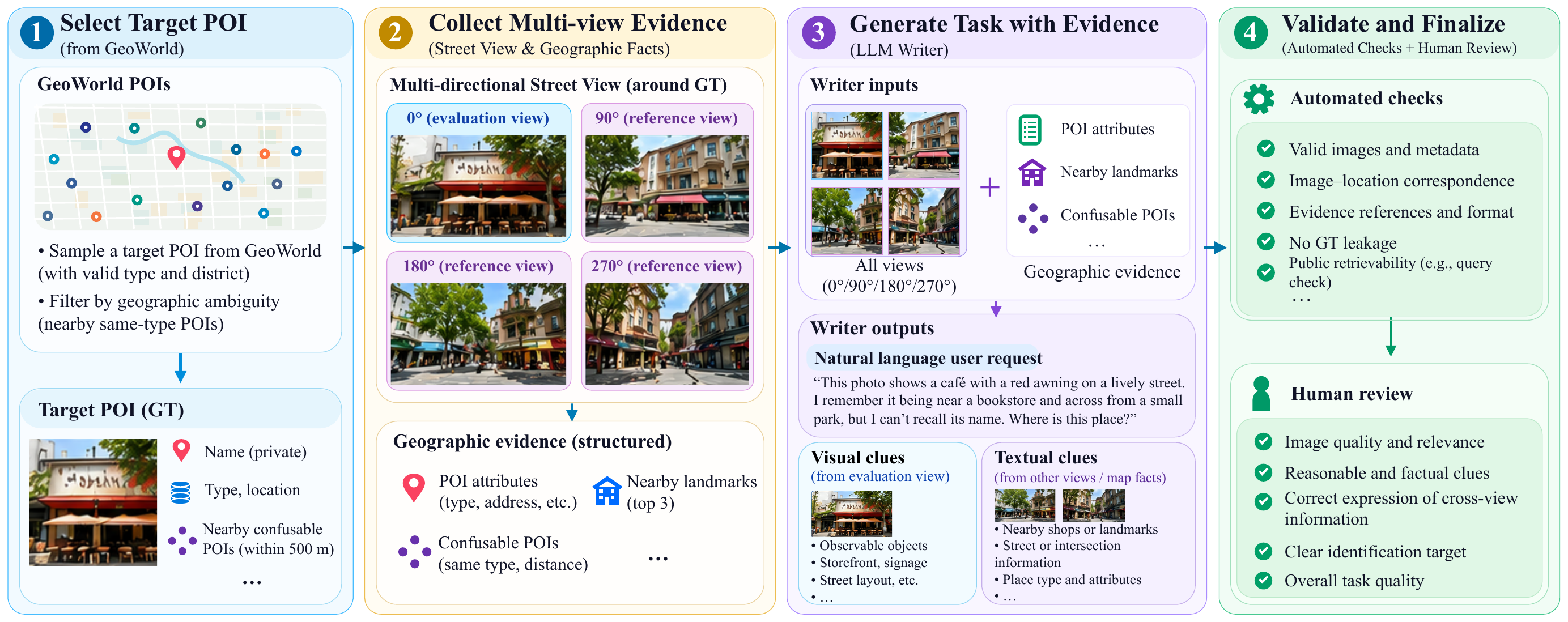}
\caption{\textbf{Grounding task construction.}
A target POI is selected from GeoWorld, and multi-view Street View images and geographic evidence are collected to support task generation.
The LLM writer distinguishes visual clues available in the evaluation image from contextual information provided through the user request.
Generated tasks undergo automated checks and human review before inclusion in LiveGeo.}
\label{fig:grounding_construction}
\end{figure}

\subsubsection{Constraint and Task Construction}

\paragraph{Static constraint and task construction.}
We construct task constraints $C_0$ by translating relevant facts in GeoWorld into explicit user requirements.
World facts describe actual conditions, such as opening hours and travel times, while constraints specify what a valid solution must satisfy, such as being open from 18:00 to 19:00 or reachable within 20 minutes.
Not every available fact becomes a constraint.
For POI-based tasks, we establish a candidate set, derive possible constraints from their attributes and route information, and examine how individual constraints and their combinations affect the GT and other candidates.
The selected constraints are recorded with their values, units, and supporting facts for subsequent verification.

The construction procedure depends on what each task is designed to evaluate.
For \emph{Grounding}, we first select a target POI and collect multi-directional images from Google Street View\footnote{\url{https://www.google.com/streetview/}}.
As illustrated in Figure~\ref{fig:grounding_construction}, an LLM writer uses these images, geographic attributes, and nearby landmarks to generate a natural-language identification request. Visual clues are drawn from the evaluation image, while information from other views is incorporated into the request as textual clues.
The writer records the supporting evidence for each clue. The generated task undergoes automated checks for image validity, spatial correspondence, and answer leakage, followed by human review of its factual and visual consistency.

\begin{promptbox}{Grounding Task Generation Prompt}
\textbf{Input:} A fixed target POI (GT); the final evaluation image; additional Street View reference images; and geographic evidence, including place attributes and nearby landmarks.
\medskip
\\
\textbf{Instruction:} Generate a natural English user request asking an agent to identify the target location using a single photograph and geographic information. Use the \emph{evaluation image} to identify what the agent can directly observe and clarify which place the request refers to. Leave useful visual details for the agent to interpret rather than describing the entire image. Use the \emph{reference images} and geographic evidence to provide relevant contextual clues. Information drawn from reference images must be expressed as textual clues in the request, not as details visible in the evaluation image. Write a natural, contextualized request of 2--4 sentences. Use only supported evidence. Do not invent place facts, visual details, or spatial relationships. Do not reveal the GT name, full address, coordinates, or internal IDs.
\medskip
\\
\textbf{Output:} A structured JSON object containing the user request, visual clues from the evaluation image, textual clues included in the request, and clue-to-evidence mappings for verification and human review.
\end{promptbox}

\begin{figure}[t]
\centering
\includegraphics[width=\linewidth]{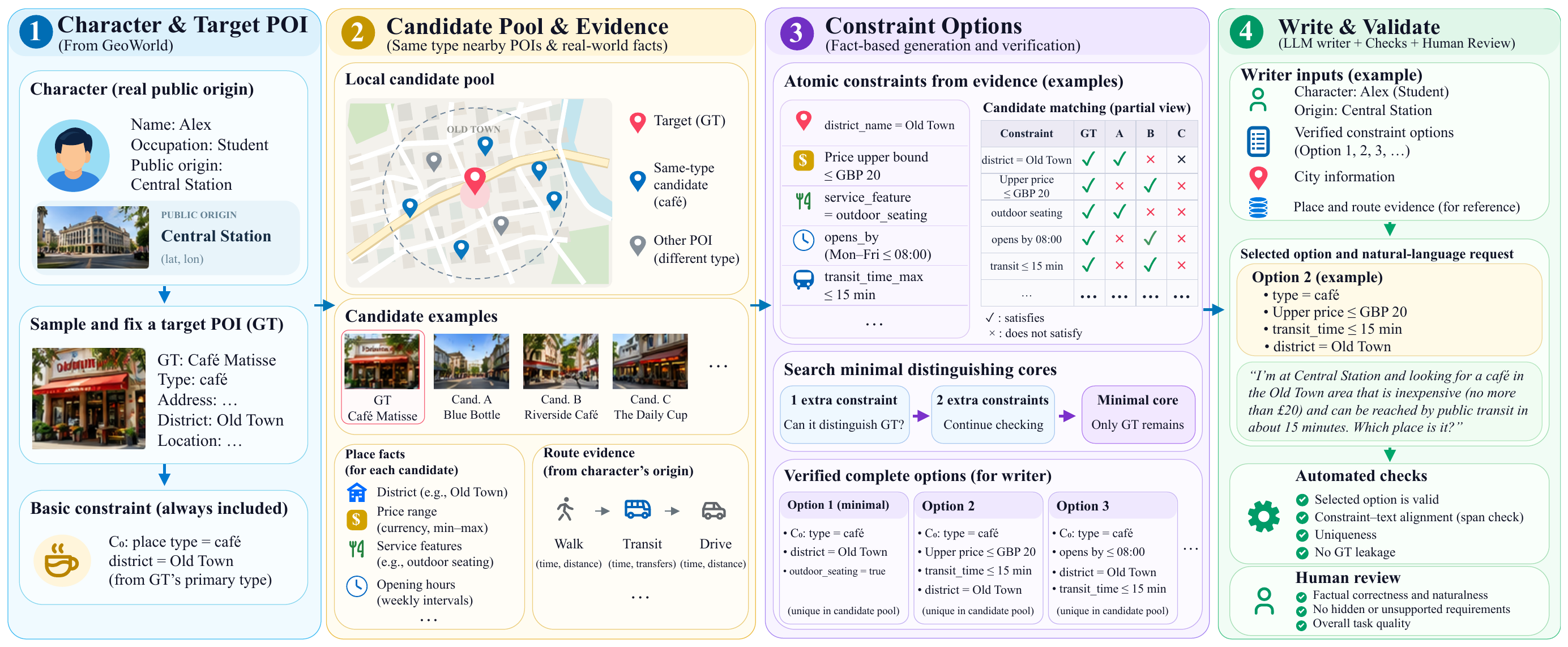}
\caption{\textbf{Discovery task construction.}
A character and target POI are selected from GeoWorld, followed by the construction of a local candidate pool and the collection of geographic and routing evidence. Script-generated constraints are evaluated against the candidates and combined into verified options that distinguish the GT within the pool. An LLM writer selects one option and converts its conditions into a natural-language request. Automated checks and human review are conducted before the task is included in LiveGeo.}
\label{fig:discovery_construction}
\end{figure}

For \emph{Discovery}, the goal is to construct constraints that distinguish the GT from nearby candidates. We first select a character with a public starting location and fix a target POI (GT). The GT's place type defines the base constraint $C_0$, which is retained in all subsequent constraint combinations. As illustrated in Figure~\ref{fig:discovery_construction}, we then construct a local candidate pool of nearby POIs matching $C_0$ and collect their geographic attributes and routing information from the character's starting location. Code generates fact-supported additional constraints, evaluates them against the candidates, and searches for combinations with the fewest additional constraints that distinguish the GT within this pool.  The LLM writer selects one verified option and expresses all its conditions in a natural-language request without modifying them. The generated task undergoes automated checks and human review before inclusion in LiveGeo.

\begin{promptbox}{Discovery Task Generation Prompt}
\textbf{Input:} A city, a character with a public starting location, and a set of script-verified constraint options.
\medskip
\\
\textbf{Instruction:} Select one of the provided constraint options and write a natural English request for finding a place. Express every condition in the selected option, including the required place type. Do not add, remove, or modify any conditions, numerical thresholds, or other specified requirements. Character information may provide context but must not introduce additional requirements. Clearly state the city and starting location. Express travel times and distances naturally using the provided actual values rather than copying internal verification ranges. Preserve the meaning of each condition and do not reveal the target POI or private identifiers. Integrate the conditions into a coherent, realistic user request.
\medskip
\\
\textbf{Output:} A structured JSON object containing the selected constraint option, a natural-language user request, and exact text corresponding to each selected condition.
\end{promptbox}

\begin{figure*}[t]
    \centering
    \includegraphics[width=0.99\textwidth]
    {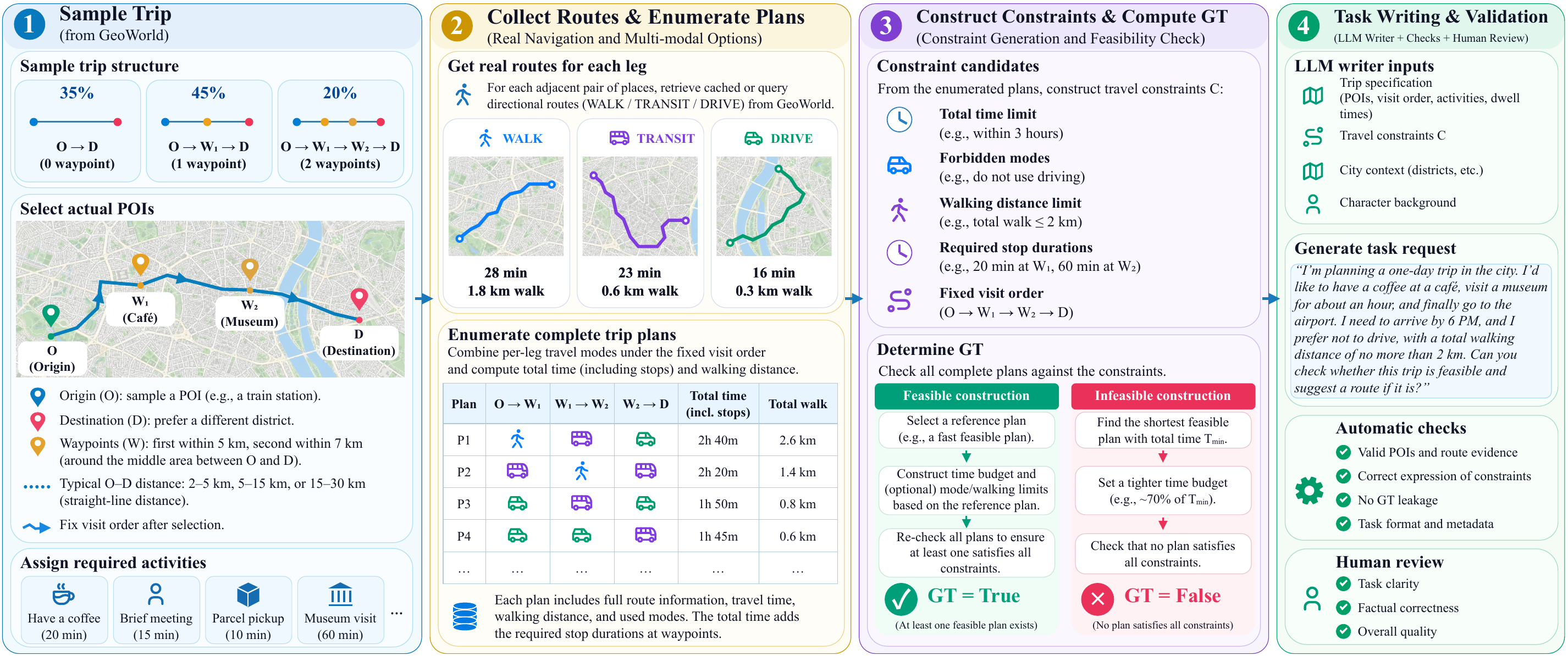}
    \caption{
    \textbf{Construction pipeline for Mobility Planning tasks.}
    The pipeline samples trip structures and required activities,
    retrieves real routes and enumerates complete travel plans,
    constructs constraints and computes feasibility ground truth,
    and generates natural-language requests through LLM writing
    and validation.
    }
    \label{fig:mobility_planning_construction}
\end{figure*}

For \emph{Mobility Planning}, we first sample an origin, a destination, and optional waypoints with a fixed visit order and required activities. We then retrieve real routes for each trip leg across different travel modes and enumerate complete trip plans. Based on these plans, we construct travel constraints, including time limits, prohibited travel modes, walking distance limits, and required stop durations, and compute the feasibility ground truth (GT) by checking whether any complete plan satisfies all constraints. Finally, an LLM writer converts the fixed trip specification and constraints into a natural-language request, followed by automatic validation and human review. The construction process is illustrated in Figure~\ref{fig:mobility_planning_construction}.

\begin{promptbox}{Mobility Task Generation Prompt}
\textbf{Input:} A traveler and a predefined trip specifying the origin, destination, optional waypoints, visit order, and total stop duration, together with travel constraints on time, transport modes, and walking distance, and an optional infeasibility hint.
\medskip
\\
\textbf{Instruction:} Write a natural English request describing the specified trip and its travel constraints. Include all provided requirements and integrate them into a coherent, realistic user request rather than listing them mechanically. Request a feasibility assessment, a detailed route plan, and an explanation of the reasoning.
\medskip
\\
\textbf{Output:} A plain-text user request describing the specified trip, its travel constraints, and the feasibility question.
\end{promptbox}

\begin{figure*}[t]
    \centering
    \includegraphics[width=0.99\textwidth]
    {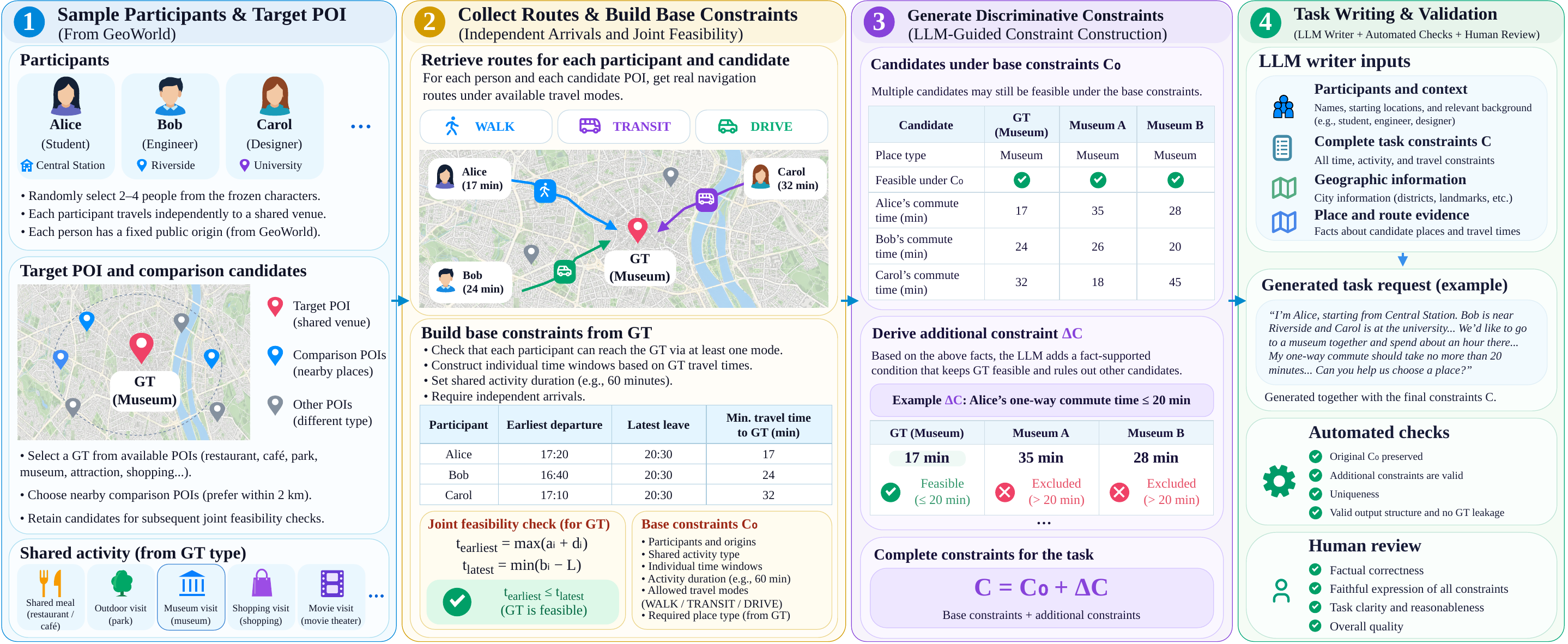}
    \caption{
    \textbf{Multi-Entity task construction.}
    Participants, a target POI, and comparison candidates are selected from GeoWorld. Real routes and participant availability are used to construct base constraints $C_0$ and check joint feasibility. An LLM adds fact-supported constraints $\Delta C$ to make the fixed GT uniquely feasible and generates a corresponding user request. The resulting task undergoes automated verification and human review.
    }
    \label{fig:multi_entity_construction}
\end{figure*}

\begin{figure*}[t]
    \centering
    \includegraphics[width=0.99\textwidth]
    {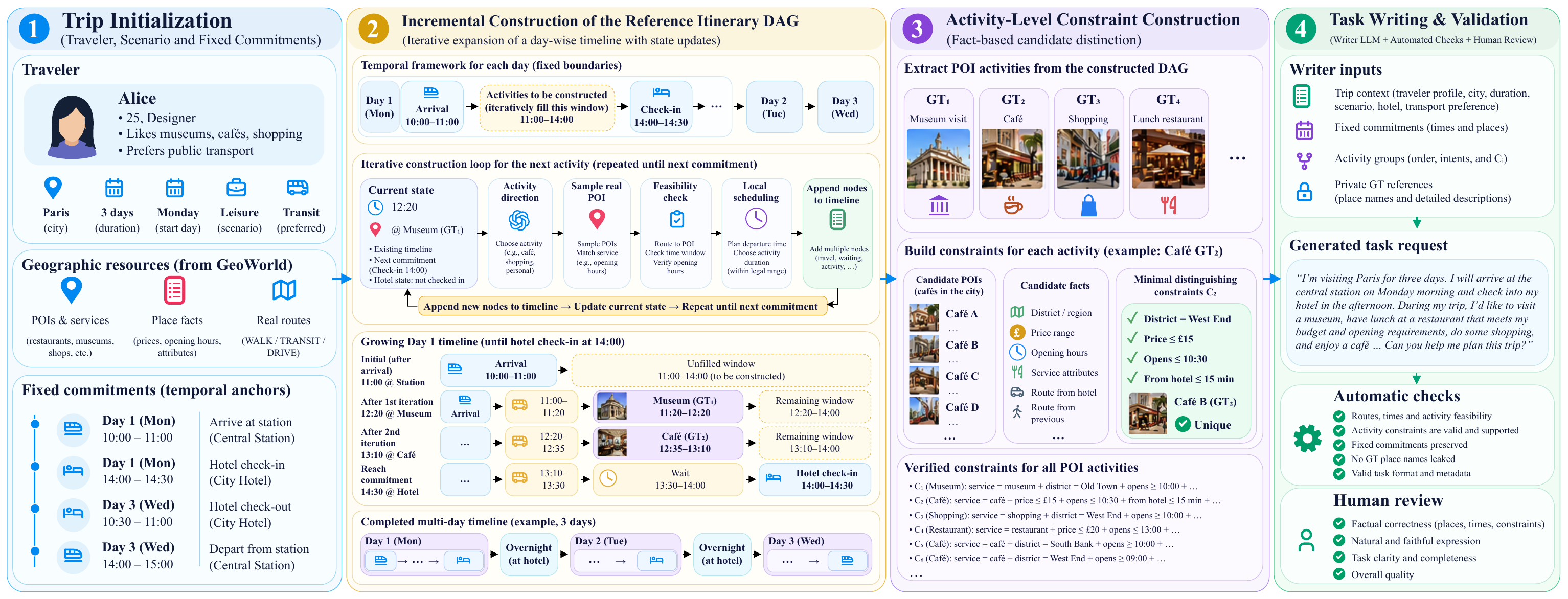}
    \caption{
    \textbf{Long-Horizon task construction.}
    Trip context and fixed commitments initialize the task.
    A reference itinerary DAG is incrementally constructed using real POIs, routes, and temporal feasibility checks.
    Activity-level constraints are then derived from geographic facts to distinguish the target POIs.
    Finally, an LLM writer generates the natural-language request, followed by automated validation and human review.
    }
    \label{fig:long_horizon_construction}
\end{figure*}

For \emph{Multi-Entity}, constraints must account for both individual requirements and the feasibility of a shared activity. We first select participants, a target POI, and comparison candidates from GeoWorld. Using real travel routes and participant availability, we construct the initial constraints $C_0$ and verify the feasibility of the shared activity. An LLM then adds fact-supported constraints $\Delta C$ to distinguish the fixed GT from other feasible candidates and generates a corresponding natural-language request. The construction process is illustrated in Figure~\ref{fig:multi_entity_construction}.

\begin{promptbox}{Multi-Entity Task Generation Prompt}
\textbf{Input:} Participants and their departure locations, availability windows, shared activity requirements, the complete task constraints, and relevant geographic and route information.
\medskip
\\
\textbf{Instruction:} Write a natural English user request asking for a suitable place where all participants can complete the shared activity. Clearly express each participant's starting location and availability, the activity type and duration, and all additional travel and place requirements. Preserve all specified conditions and numerical thresholds without adding, removing, or modifying any requirements. Integrate these details into a coherent, realistic request rather than listing structured fields. Do not reveal the target POI, other candidate places, or verification results. Use only the provided information.
\medskip
\\
\textbf{Output:} A structured JSON object containing the original participant and activity requirements, additional constraints, and the corresponding natural-language user request.
\end{promptbox}

For \emph{Long-Horizon}, a feasible multi-day reference itinerary is constructed by iteratively filling the time windows between fixed commitments. At each step, activities are selected, real POIs are sampled when needed, and route and time feasibility are checked before new nodes are appended to the timeline.The completed itinerary is then used to derive constraints for each POI activity, as illustrated in Figure~\ref{fig:long_horizon_construction}. Constraints operate at two levels: individual activities must satisfy their location, service, and time requirements, while the full itinerary must respect activity order, travel times, durations, and fixed commitments such as hotel check-in and departure. Importantly, detailed schedules stored in the reference itinerary serve as feasibility evidence and are not automatically included as user constraints.

\begin{promptbox}{Long-Horizon Task Generation Prompt}
\textbf{Input:} A traveler profile and trip context, fixed commitments with their locations and times, activity groups containing activity intents and selected place constraints, and private GT references.
\medskip
\\
\textbf{Instruction:} Convert the provided trip information, fixed commitments, and activity requirements into a natural, realistic English user request. Write in the first person or on behalf of the traveler, organizing the trip by day and actual itinerary order. Integrate each activity's purpose, inbound travel, preceding and follow-on personal activities, and return travel into a coherent description. Express every provided place constraint without omitting, relaxing, strengthening, or inventing any conditions. Include route durations as user requirements only when explicitly specified in the place constraints, using natural and verifiable approximations. When an activity requires multiple services, express them as a combined need. Use personal activities only to provide context, without introducing additional place-selection requirements. Preserve monetary limits without inventing an overall budget, and use the 24-hour format for clock times. Do not reveal the target POIs in the private GT references, but retain the specified locations and times of fixed commitments.
\medskip
\\
\textbf{Output:} A structured JSON object containing a complete natural-language user request, covering the trip context, fixed commitments, activities in chronological order, and all selected place constraints.
\end{promptbox}

\begin{figure*}[t]
    \centering
    \includegraphics[width=0.99\textwidth]
    {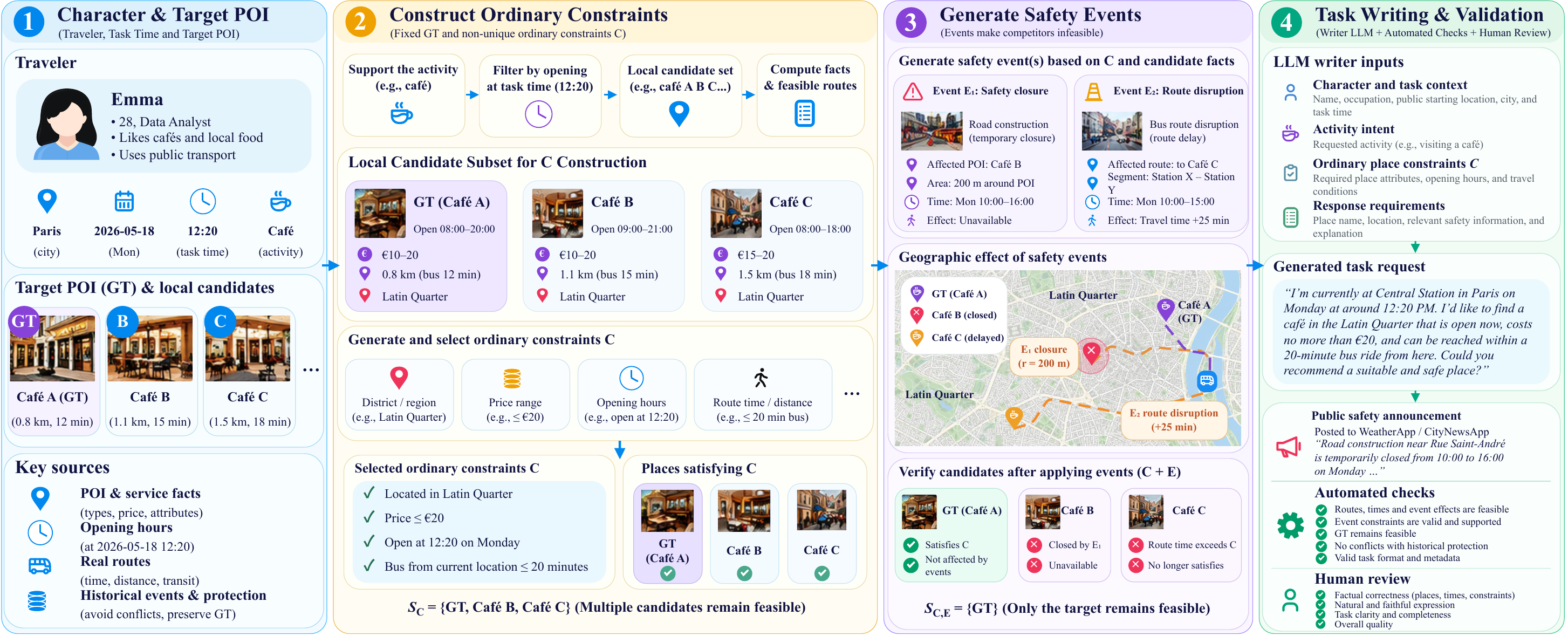}
    \caption{
    \textbf{Safety-Aware task construction.}
    A fixed GT and competing POIs are selected, and ordinary constraints $C_0$ are constructed to retain multiple feasible candidates.
    Safety events are then generated from geographic and route evidence to make the competing candidates infeasible while preserving the GT.
    Finally, an LLM writer generates the user request without disclosing the events, followed by automated validation and human review.
    }
    \label{fig:safety_aware_construction}
\end{figure*}

\emph{Safety-Aware} follow a different construction principle: ordinary constraints $C_0$ must not independently determine the GT.
For Safety-Aware tasks, $C_0$ is constructed to retain both the GT and at least one competing location, so that an additional safety condition $E$ is necessary to distinguish the safe choice.
For example, two hotels may satisfy the same travel and availability requirements, but only one remains suitable under the given safety conditions. This ensures that identifying the correct location requires considering current safety information rather than relying on ordinary place constraints alone. The construction process is illustrated in Figure~\ref{fig:safety_aware_construction}.

\begin{promptbox}{Safety Event Construction Prompt}
\textbf{Input:} An activity intent, selected ordinary constraints $C$, a fixed target POI and competing candidates, verified place and route evidence, geographic references and maps, the task time, an allowed event-generation window, and existing spatial protection information.
\medskip
\\
\textbf{Instruction:} Construct realistic simulated safety events that make the competing candidates infeasible while preserving the fixed target under the unchanged constraints $C$. Generate exactly one event per competing candidate, allowing each event to affect multiple candidates when supported by the evidence. Choose appropriate safety scenarios and define their impacts using either an approximate POI-centered area or a supplied navigation segment. Ensure that each event covers the task time, falls within the permitted time window, and does not conflict with existing spatial protection. For every event, explain its exposure and effects on all candidates using their respective geographic or route evidence. Specify whether affected places or routes become unavailable or whether travel times increase. Generate a public English notice consistent with each event's location, timing, and practical consequences without revealing the target or candidate roles. Separately specify the protection needed to preserve the target POI and any routes required by $C$. If no supported construction is possible, explain why rather than inventing a valid outcome.
\medskip
\\
\textbf{Output:} A structured JSON object containing the safety events, their types, time windows, geographic scopes, effects, candidate-specific exposure evidence, and public notices, together with a separate GT protection specification covering the target location, required routes, and active time window.
\end{promptbox}

\begin{promptbox}{Safety-Aware Task Generation Prompt}
\textbf{Input:} A traveler profile, task context including the city, starting location and task time, an activity intent, selected ordinary place constraints, and the required response content.
\medskip
\\
\textbf{Instruction:} Write a natural English user request from the traveler's perspective or on their behalf. Integrate the activity purpose, starting location, task time, and all specified place requirements into a coherent and realistic request. Preserve every condition and numerical threshold without adding, removing, relaxing, or strengthening any requirements. Express travel times and distances naturally rather than reproducing precise route measurements mechanically. Use the traveler's personal background only as context, without introducing additional place-selection conditions. Use the supplied English names or appropriate transliterations for public locations. Ask for a suitable and safe place that remains practically accessible under current conditions, including its name and location, relevant safety information, and an explanation of the choice.
\medskip
\\
\textbf{Output:}  A structured JSON object containing a complete natural-language user request, covering the activity intent, traveler context, starting location, time requirements, ordinary place constraints, and the requested safety-informed recommendation and justification.
\end{promptbox}

\begin{figure*}[t]
    \centering
    \includegraphics[width=0.99\textwidth]
    {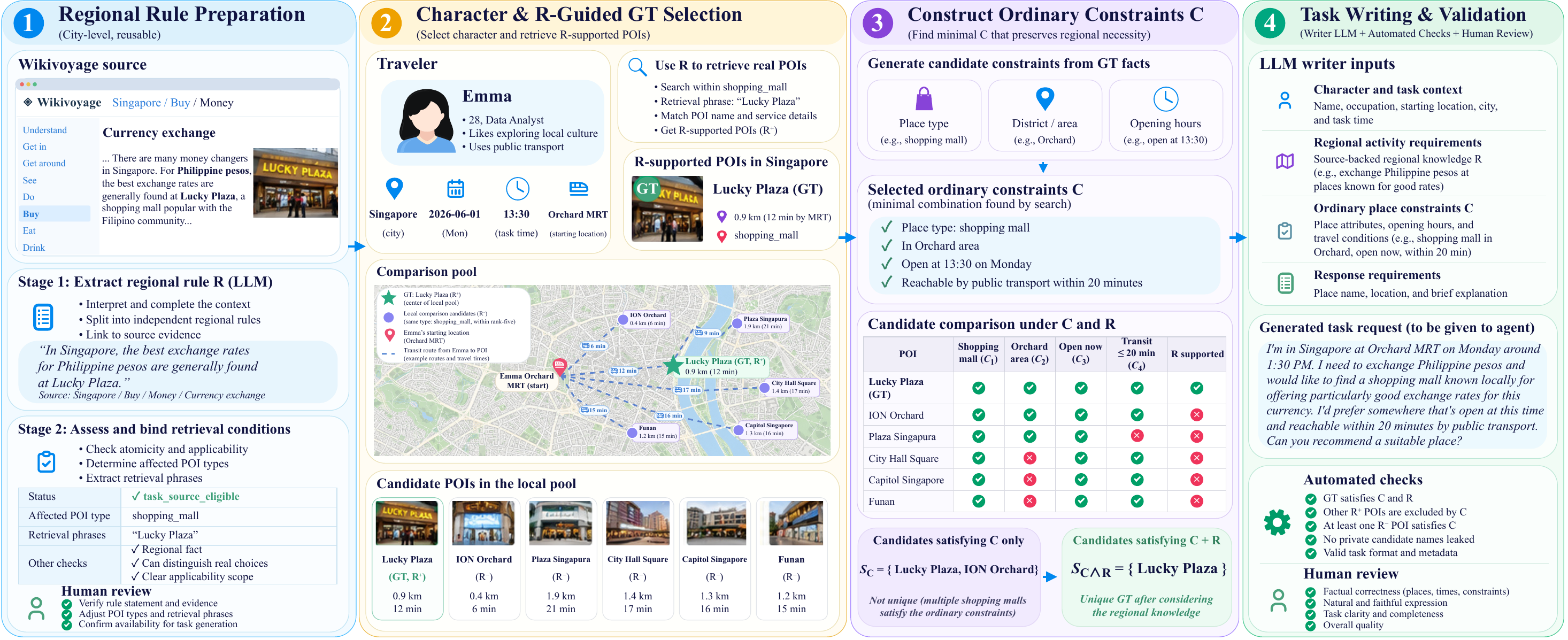}
    \caption{
    \textbf{Region-Aware task construction.}
    Source-backed regional criteria $R$ are extracted from Wikivoyage and assessed for POI selection.
    An $R$-supported GT and comparison candidates are selected.
    Ordinary constraints $C_0$ are constructed to exclude other $R$-supported POIs while retaining at least one competing POI, making $R$ necessary to uniquely identify the GT.
    Finally, an LLM writer generates the task request, followed by automated validation and human review.
    }
    \label{fig:region_aware_construction}
\end{figure*}

For \emph{Region-Aware} tasks, regional criteria $R$ are derived from Wikivoyage\footnote{\url{https://en.wikivoyage.org/}}. We use GPT-5.6 Sol to extract region-specific constraints from Wikivoyage pages, select those that can affect POI selection, and verify them through human review. The GT must satisfy both the ordinary constraints $C_0$ and the regional criterion $R$.
We construct $C_0$ to exclude other candidates satisfying $R$, while retaining at least one candidate that satisfies $C_0$ but not $R$.
This ensures that the regional criterion is necessary to identify the GT, rather than being an optional detail that does not affect the decision.
The construction process is illustrated in Figure~\ref{fig:region_aware_construction}.

\begin{promptbox}{Region-Aware Task Generation Prompt}
\textbf{Input:} A character and task context, including the city, starting location, and task time; a source-backed regional activity context and place requirements; verified ordinary place constraints $C$; and response requirements.
\medskip
\\
\textbf{Instruction:} Write a natural English user request from the traveler's perspective or on their behalf. Use the regional context to explain the activity's purpose naturally, but only express concrete regional place requirements supported by the provided retrieval phrases. Do not directly recite regional rules, source recommendations, or conclusions, or introduce requirements based on unconfirmed examples. Include every ordinary place constraint, preserving all specified conditions and numerical thresholds without adding, removing, relaxing, or strengthening any requirements. Express place types, areas, services, prices, opening hours, travel times, and distances naturally, integrating related conditions into a coherent request. Clearly state the city, public starting location, and task time without revealing raw coordinates. Do not reveal the target POI or internal construction information. Request the selected place's name and location.
\medskip
\\
\textbf{Output:} A structured JSON object containing a complete natural-language user request, including the regional activity need, task context, all selected ordinary place constraints, and the required place recommendation
\end{promptbox}

\paragraph{Dynamic constraint and task construction.}
Dynamic tasks are constructed from verified static tasks, inheriting the user request, initial constraints $C_0$, ground truth $GT_0$, and world state $W_0$. A Direction Agent first identifies the dimensions along which the target outcome should change, based on the original task and available geographic facts. For place-selection tasks, these dimensions guide the filtering of a local candidate pool, from which a new target $GT_1$ is sampled and fixed before event generation. When changes to the task requirements are needed, target constraints $C_1$ are constructed using the facts of $GT_1$ and comparisons with eligible candidates, while unaffected requirements from $C_0$ are preserved.

\begin{figure*}[t]
    \centering
    \includegraphics[width=0.99\textwidth]{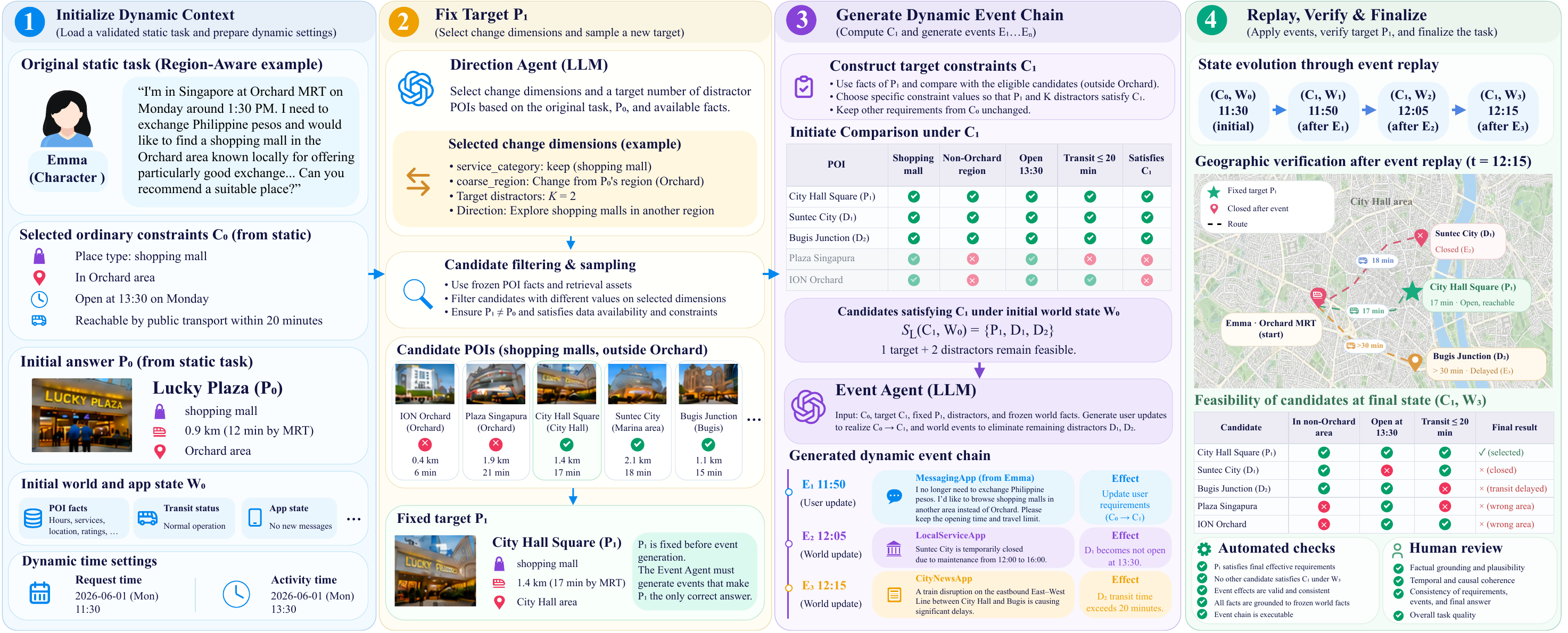}
    \caption{
    \textbf{Dynamic task construction.} Starting from a validated static task, a Direction Agent identifies change dimensions and a target number of distractors. A new target $GT_1$ is sampled from the filtered candidates, and target constraints $C_1$ are constructed to retain $GT_1$ and the selected distractors. An Event Agent generates timestamped public messages and structured updates to realize the constraint changes and make the distractors infeasible. The event sequence is then replayed and automatically verified, followed by human review of event consistency and plausibility.}
    \label{fig:dynamic_event_construction}
\end{figure*}

In the example shown in Figure~\ref{fig:dynamic_event_construction}, the target constraints retain $GT_1$ and two competing POIs before the events are applied. These target constraints guide event construction. Given the fixed $GT_1$, the target constraint changes, and relevant world information, a construction LLM, referred to as the \emph{Event Author}, generates a complete sequence of timestamped events. Each event contains a public message, observable through the corresponding application, and a structured update specifying its effects on the world.
The number and types of events depend on the task. In Figure~\ref{fig:dynamic_event_construction}, the first event updates the user's activity and regional requirements, while the subsequent events temporarily close one competing POI and disrupt travel to another. Together, these events allow the fixed $GT_1$ to become the unique feasible choice. Events may also involve weather, participant availability, or other changes to the environment. The Event Author cannot alter the fixed $GT_1$ or the underlying geographic facts. Each event records its publication time, effective interval, affected entities, and supporting references, allowing announcements and their actual effects to be distinguished.

The generated events are replayed chronologically in a temporary world state. Both user messages and environmental changes update the evolving world, while the effective task constraints reflect any explicit changes to user or participant requirements. Verification checks the validity of event references, temporal consistency, the feasibility of the resulting state changes, and whether the final outcome matches the fixed $GT_1$. For place-selection tasks, this includes checking the remaining candidates against the effective constraints and updated world state. Human reviewers further assess the plausibility of the events and the consistency between public messages and structured updates. Only validated event sequences that do not conflict with existing events or protected task dependencies are incorporated into the shared world timeline.

\begin{promptbox}{Dynamic Event Chain Generation Prompt}
\textbf{Input:} The original static task and its constraints $C_0$, the reference answer $P_0$, a fixed target $P_1$, verified POI and route facts, target constraint changes and remaining competing candidates, editable constraint specifications, the task's temporal context, and geographic evidence with historical spatial protection. 
\medskip
\\
\textbf{Instruction:} Generate a complete, plausible causal chain of events that makes the fixed $P_1$ the correct outcome under the updated requirements and evolving world state. Use the provided target constraint changes to guide event construction where supplied, while preserving the original request, unchanged requirements, and frozen facts. For place-selection tasks, generate events that render the remaining competing candidates infeasible under the effective constraints while preserving the feasibility of the fixed $P_1$. Construct events through supported App operations, including updates to user or participant requirements, service availability, traffic, transit, weather, and safety conditions. Each event must specify its activation time, effective interval, affected scope, structured state changes, public message, causal role, and dependencies on preceding events. Public messages must naturally communicate the corresponding changes through their respective Apps without revealing $P_0$, $P_1$, or verification results.
\medskip
\\
\textbf{Output:} A structured JSON object containing the complete event chain, including each event's App, event type, activation time, affected scope, structured state changes, public message, causal role, and causal dependencies, together with private target-protection declarations when required.
\end{promptbox}

Notably, for \emph{Mobility Planning}, the ground truth is a Boolean feasibility judgment. We sample whether $GT_1$ preserves or reverses $GT_0$ with equal probability, then generate events to realize the target outcome under the effective travel requirements and route conditions. For example, a traffic disruption may cause a previously feasible journey to exceed its travel-time limit, making it infeasible, whereas a less severe delay may leave its feasibility unchanged. After event replay, we recompute journey feasibility to verify $GT_1$.

\paragraph{Event protection.} As dynamic tasks are progressively constructed and incorporated into the shared world timeline, newly generated events may conflict with existing events in space and time or affect the world states on which the ground truths of previous tasks depend. To prevent interference between tasks while allowing diverse events to coexist in the same geographic world, we introduce an event protection mechanism. This mechanism maintains two types of historical records: (1) the affected geographic scopes and effective intervals of committed events, and (2) the resources on which the ground truths of previously validated tasks depend, including places, routes, usage times, and participant or user request identities.

\begin{figure*}[t]
    \centering
    \includegraphics[width=0.99\textwidth]{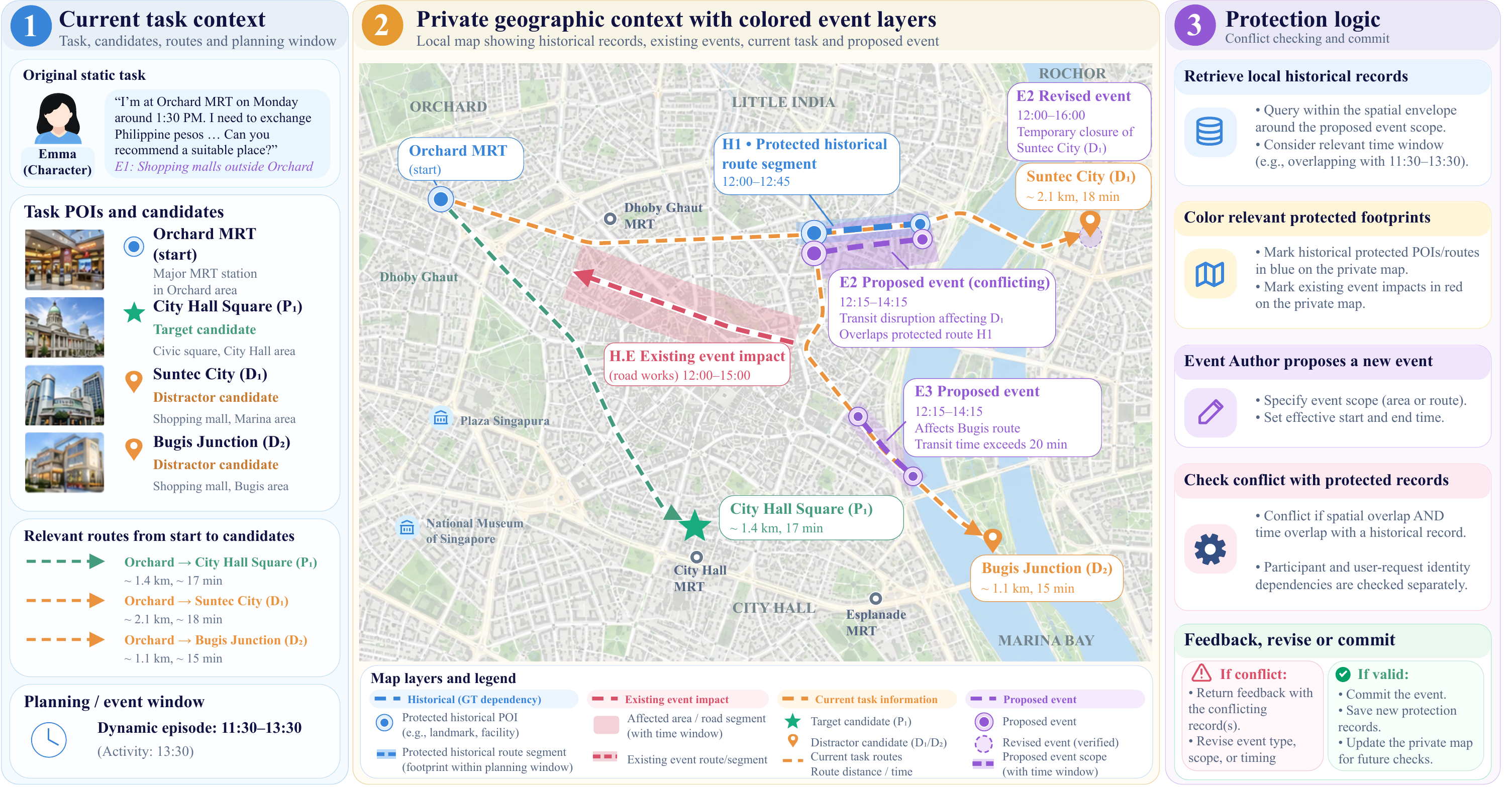}
    \caption{Event protection mechanism in the shared geographic world. Historical event footprints and protected task dependencies are retrieved according to the current task's spatial and temporal scope and visualized on the Event Author's private map. Conflicting events are revised and reverified, while validated events and their protection records are incorporated into the shared world timeline.}
    \label{fig:event_protection}
\end{figure*}

As illustrated in Figure~\ref{fig:event_protection}, before generating new dynamic events, we define a local spatiotemporal scope based on the places, routes, and time window relevant to the current task. We then retrieve historical event footprints and protected task dependencies relevant to this scope. Both types of records are filtered based on spatial intersection and temporal relevance, and the results are visualized using distinct colors and markers on a private map provided to the Event Author. The map displays the affected areas of existing events alongside protected places and routes, allowing the Event Author to identify potential spatiotemporal conflicts and select appropriate event locations and times. Participant and user request identity dependencies are not visualized on the map but are checked separately during subsequent verification.

After event generation, we determine the actual affected scopes and effective intervals from the structured updates and check them against the complete historical records. For geographically relevant records, a spatiotemporal conflict occurs only when a new event spatially intersects a historical protection record and their effective intervals overlap. Spatial intersections are determined using the original geographic geometries, including boundary contact. Temporal overlap is determined using half-open intervals: two intervals $[s_1,e_1)$ and $[s_2,e_2)$ overlap if $s_1<e_2$ and $s_2<e_1$. For identity-related records, we further check whether a new event affects the same participant or user request during an overlapping time interval. When a conflict is detected, the relevant records and reasons for the conflict are returned to the Event Author to revise the event sequence, followed by another round of verification.

Only events that pass the complete historical conflict checks are committed to the shared world timeline. Their corresponding event records and task protection information are also retained for subsequent tasks. Notably, overlapping protection scopes do not inherently constitute a conflict. For example, two tasks may depend on the same route or use the same place at different times. The mechanism restricts new events that conflict with existing protection records, rather than preventing tasks from sharing geographic resources. In this way, GeoWorld preserves the validity of historical tasks and supports the continued accumulation of dynamic events within the same geographic environment without requiring tasks to occupy entirely separate spatial or temporal regions.

\subsubsection{Task Verification}

\paragraph{Static task verification.}
We verify each constructed task through programmatic checks before generating the final user request.
The verification covers three main aspects.
\textit{(1) Fact Support:} All entities, place attributes, and route information used in the task must be supported by records from the corresponding GeoWorld version.
Missing information is not treated as a negative fact; for example, an unavailable route record cannot be used to conclude that a destination is unreachable.
\textit{(2) Constraint Consistency:} Each constraint must specify its applicable entity, condition, value, unit, and relevant temporal or spatial context.
We check that the constraints are well-defined, mutually consistent, and supported by the available facts.
\textit{(3) GT Validity:} We determine whether the GT constitutes a valid solution under the complete task requirements and whether the required distinctions between candidates hold.
Since different task categories have different validity conditions, we apply task-specific verification rules, summarized in Table~\ref{tab:task_verification}.
Here, $E$ denotes the safety-related environmental conditions in Safety-Aware tasks, while $R$ denotes the region-specific criteria in Region-Aware tasks.

\begin{table}[t]
\centering
\small
\renewcommand{\arraystretch}{1.2}
\caption{Task-specific verification criteria and supporting evidence in LiveGeo.}
\label{tab:task_verification}
\begin{tabularx}{\textwidth}{lXX}
\toprule
\textbf{Task category} & \textbf{Verification criteria} & \textbf{Supporting evidence} \\
\midrule

Grounding
& Check whether the image, clues, and target location are consistent and publicly identifiable.
& Image-source records, landmarks, and retrieval evidence. \\

Discovery
& Verify that the GT satisfies $C_0$ and is distinguished from other candidates within the defined comparison set.
& Candidate-constraint satisfaction records and fact sources. \\

Mobility Planning
& Determine whether a complete journey satisfies all travel requirements, or establish infeasibility.
& Route records, feasible route chains, or supported infeasibility reasons. \\

Multi-Entity
& Verify that all participants can reach the location and complete the shared activity within their time windows.
& Individual routes, arrival times, and shared activity windows. \\

Long-Horizon
& Check individual activity requirements and the feasibility of the complete itinerary.
& Activity matching, travel times, schedules, and reference arrangements. \\

Safety-Aware
& Verify that $C_0$ retains competing candidates, while $C_0$ together with $E$ identifies the GT.
& Risk effects, spatial and temporal relations, and candidate outcomes. \\

Region-Aware
& Verify that $C_0$ retains candidates not satisfying $R$, while $C_0$ together with $R$ identifies the GT.
& Regional evidence, candidate comparisons, and counterfactual checks. \\

\bottomrule
\end{tabularx}
\end{table}

Beyond these checks, we verify \textit{Public Observability}, ensuring that the information needed to solve the task can be obtained through the user request or available agent tools, rather than relying on hidden construction records.
This does not require exposing internal GT labels or region-specific criteria directly to the agent; instead, the available geographic facts and task context must provide sufficient information for the intended decision.

After programmatic verification, the LLM writer converts the structured task into a natural-language request.
We check whether the generated request preserves all selected constraints without changing their values, omitting requirements, introducing unsupported conditions, or revealing the GT.
Human reviewers further examine the request, supporting evidence, and potential ambiguities that programmatic checks may not fully capture.
Tasks that fail these checks are revised or discarded according to the corresponding construction rules.

\paragraph{Dynamic task verification.}
Dynamic verification extends the static procedure to changes in world states and task requirements.
We first check that the dynamic task preserves the initial world state, user request, constraints $C_0$, and GT $GT_0$ of its static counterpart.
We then verify the generated events, checking that their affected entities, timestamps, observable messages, and corresponding state updates are valid and consistent.

Next, we apply the events in chronological order to reconstruct the evolving world states and effective task constraints.
Environmental changes update the corresponding world states, while explicit changes to user or participant requirements update the affected constraints without altering the remaining requirements.
Based on these updated conditions, we verify whether the predefined $GT_1$ becomes a valid solution and whether the intended change from $GT_0$ to $GT_1$ is supported by the resulting world state.
For example, when Restaurant A closes and Restaurant B extends its opening hours, we check that these events produce the corresponding state updates and that B becomes a valid choice under the original dining requirements.

We also verify that each event has a valid and observable effect, although an individual event need not independently change the final GT.
When events are integrated into a shared world timeline, additional checks ensure that they do not conflict with existing events or invalidate protected task dependencies.
Finally, human reviewers examine the event sequence, its effects on the task, and the supporting evidence.
Only tasks that pass the required programmatic checks and human review are retained in LiveGeo.

\subsection{Evaluation Details}
\label{app:bench_evaluation}

This section details the evaluation indicators and correctness criteria for each LiveGeo task category. Beyond determining overall task success, we use these indicators to further analyze model performance and failure patterns across task categories in  Appendix~\ref{app:failure_analysis}.

\subsubsection{Grounding}
We evaluate grounding at two complementary levels: whether the agent \emph{selects the correct geographic entity} through tool interaction, and whether it \emph{communicates the correct grounded result} in its final response. This distinction mirrors everyday wayfinding: when helping someone locate a place, one must not only identify the correct location, but also communicate it clearly to the person asking. Accordingly, we define two indicators, \textbf{Tool-grounded} and \textbf{User-facing}, as summarized in Table~\ref{tab:grounding_eval_metrics}. Together, they capture the two stages of grounding: selecting the correct location in the environment and communicating that result to the user.

\begin{table}[t]
\centering
\small
\renewcommand{\arraystretch}{1.18}
\caption{Evaluation indicators for Grounding.}
\label{tab:grounding_eval_metrics}
\begin{tabularx}{\linewidth}{@{}p{2.15cm} X p{4.0cm}@{}}
\toprule
\textbf{Indicator} & \textbf{Criterion} & \textbf{What it measures} \\
\midrule

\textbf{Tool-grounded}
&
\textbf{(i)} The agent invokes the location-selection tool;
\textbf{(ii)} the selected location is correct.
&
Whether the agent correctly selects the target location through tool interaction.
\\

\textbf{User-facing}
&
\textbf{(i)} The agent submits a final response;
\textbf{(ii)} the response contains the correct place name.
&
Whether the correct location is explicitly communicated to the user.
\\

\bottomrule
\end{tabularx}
\end{table}

\paragraph{Overall grounding correctness.}
A test instance is considered correctly grounded only when both indicators pass. Formally, let $T$ denote the \textbf{Tool-grounded} indicator and $U$ denote the \textbf{User-facing} indicator. We define the overall grounding outcome as
\[
G = T \land U.
\]
Thus, an instance is counted as successful only if the agent both selects the correct geographic entity through the tool and communicates the correct place name in its final response. A failure at either stage results in an incorrect grounding outcome.

\subsubsection{Discovery}
We evaluate discovery by whether the agent \emph{recommends an acceptable place for the user's request}. In everyday place search, exploring multiple plausible options is useful only if the process ultimately leads to an acceptable destination. Accordingly, we use \textbf{Correctness} as the primary indicator, as summarized in Table~\ref{tab:discovery_eval_metrics}. The evaluator compares the submitted place name and coordinates against the instance's private acceptable-answer set.

\begin{table}[t]
\centering
\small
\renewcommand{\arraystretch}{1.18}
\caption{Evaluation indicator for Discovery.}
\label{tab:discovery_eval_metrics}
\begin{tabularx}{\linewidth}{@{}p{2.15cm} X p{4.0cm}@{}}
\toprule
\textbf{Indicator} & \textbf{Criterion} & \textbf{What it measures} \\
\midrule

\textbf{Correctness}
&
\textbf{(i)} The agent submits a final selection containing a place name and coordinates;
\textbf{(ii)} both match the same entry in the acceptable-answer set.
&
Whether the agent ultimately recommends an acceptable destination.
\\

\bottomrule
\end{tabularx}
\end{table}

\paragraph{Overall discovery correctness.}
A test instance is considered \textbf{correct} if the agent submits a complete final selection whose place name and coordinates match the same entry in the private acceptable-answer set. Intermediate search and comparison steps are retained for trajectory analysis, but do not affect the final Discovery outcome.

\subsubsection{Mobility Planning}
We evaluate mobility planning by whether the agent \emph{correctly determines whether a requested trip is feasible} and, when it is feasible, \emph{submits a route that satisfies the user's travel constraints}. In everyday travel planning, reaching a destination is not enough: the journey must connect the required places, visit necessary stops, use permitted transport modes, and remain within the user's time and walking limits. Accordingly, we define seven indicators, summarized in Table~\ref{tab:mobility_eval_metrics}. Together, they capture the agent's feasibility judgment, the validity and constraint compliance of the submitted route, and the efficiency of its tool interaction.

\begin{table}[t]
\centering
\small
\renewcommand{\arraystretch}{1.18}
\caption{Evaluation indicators for Mobility Planning.}
\label{tab:mobility_eval_metrics}
\begin{tabularx}{\linewidth}{@{}>{\raggedright\arraybackslash}p{2.15cm} X p{4.0cm}@{}}
\toprule
\textbf{Indicator} & \textbf{Criterion} & \textbf{What it measures} \\
\midrule

\textbf{Feasibility consistency}
&
The submitted Boolean feasibility judgment matches the private ground truth.
&
Whether the agent correctly determines whether the requested trip is feasible.
\\

\textbf{Route structure}
&
The submitted route starts at the required origin, ends at the required destination, and forms a connected sequence of supported route legs.
&
Whether the selected legs form a valid journey.
\\

\textbf{Waypoint coverage}
&
All required intermediate places are visited in the specified order.
&
Whether the journey includes the required stops in the correct sequence.
\\

\textbf{Mode compliance}
&
No selected route leg uses a prohibited transport mode.
&
Whether the journey respects the user's transport restrictions.
\\

\textbf{Time compliance}
&
Travel time plus required stop durations does not exceed the specified time budget.
&
Whether the journey fits within the available time.
\\

\textbf{Walking compliance}
&
The cumulative walking distance does not exceed the specified limit.
&
Whether the journey respects the user's walking allowance.
\\

\textbf{Tool efficiency}
&
The number of tool calls remains within the call budget.
&
Whether the agent completes the planning process without excessive tool interaction.
\\

\bottomrule
\end{tabularx}
\end{table}

\paragraph{Overall mobility planning correctness.}
A test instance is considered \textbf{correct} only when \textbf{Feasibility consistency} passes and at least five of the seven indicators pass. Formally, let $F$ denote the \textbf{Feasibility consistency} indicator and let $I_1,\ldots,I_7$ denote the seven binary indicators, with $I_1=F$. We define
\[
M = F \land \left(\sum_{j=1}^{7} I_j \geq 5\right).
\]
Thus, a correct feasibility judgment is mandatory, while the remaining indicators jointly determine whether the submitted plan is sufficiently valid and constraint-compliant. Indicators that cannot be verified do not count as passes.

\subsubsection{Multi-Entity}
We evaluate multi-entity tasks by whether the agent \emph{selects the correct shared destination} while accounting for the participants' coordination requirements. In everyday group planning, choosing a meeting place is only part of the task: participants may differ in their budgets, dietary needs, and availability, and everyone must be included in the communication. Accordingly, we report six indicators, summarized in Table~\ref{tab:multi_entity_eval_metrics}. Place selection determines overall correctness, while the remaining five indicators provide auxiliary diagnostics of the submitted plan and coordination process.

\begin{table}[t]
\centering
\small
\renewcommand{\arraystretch}{1.18}
\caption{Evaluation indicators for Multi-Entity.}
\label{tab:multi_entity_eval_metrics}
\begin{tabularx}{\linewidth}{@{}>{\raggedright\arraybackslash}p{2.15cm} X p{4.0cm}@{}}
\toprule
\textbf{Indicator} & \textbf{Criterion} & \textbf{What it measures} \\
\midrule

\textbf{Place selection}
&
The submitted place matches the private target by name and, when provided, coordinates.
&
Whether the agent selects the correct shared destination.
\\

\textbf{Feasibility consistency}
&
The submitted feasibility judgment agrees with the private ground truth.
&
Whether the agent correctly determines whether a shared plan is feasible.
\\

\textbf{Budget compliance}
&
The selected place's per-person price does not exceed the specified budget.
&
Whether the destination respects the group's budget constraint.
\\

\textbf{Dietary compatibility}
&
The selected place accommodates all specified dietary requirements.
&
Whether the destination satisfies the group's dietary needs.
\\

\textbf{Time-window compliance}
&
The submitted meeting interval lies within the shared availability window.
&
Whether the proposed meeting time fits the participants' common availability.
\\

\textbf{Notification completeness}
&
The recorded messaging actions collectively include all required participants.
&
Whether every participant is included in the communication.
\\

\bottomrule
\end{tabularx}
\end{table}

\paragraph{Overall multi-entity correctness.}
A test instance is considered \textbf{correct} when \textbf{Place selection} passes. The remaining five indicators are auxiliary diagnostics and do not impose additional conditions on overall correctness. They instead characterize how the agent handles aspects of group coordination, such as accommodating participant constraints and communicating the resulting plan.

\subsubsection{Long-Horizon}

We evaluate long-horizon planning by whether the agent \emph{constructs a complete and acceptable itinerary across multiple days}. Unlike individual place selection, long-horizon planning requires the agent to coordinate multiple geographic decisions within a single plan. A destination may satisfy the requirements of an individual activity, yet its inclusion may render the overall itinerary infeasible due to travel connections, opening hours, activity order, or fixed commitments. The key challenge therefore lies not only in selecting suitable places, but in ensuring that these selections form a jointly feasible sequence. Accordingly, we adopt \textbf{Itinerary correctness} as the primary indicator, as summarized in Table~\ref{tab:long_horizon_eval_metrics}. Rather than evaluating places independently, this indicator assesses the correctness of the complete ordered itinerary.

\begin{table}[t]
\centering
\small
\renewcommand{\arraystretch}{1.18}
\caption{Evaluation indicator for Long-Horizon.}
\label{tab:long_horizon_eval_metrics}
\begin{tabularx}{\linewidth}{
@{}>{\raggedright\arraybackslash}p{2.15cm} X p{4.0cm}@{}
}
\toprule
\textbf{Indicator} & \textbf{Criterion} & \textbf{What it measures} \\
\midrule

\textbf{Itinerary correctness}
&
\textbf{(i)} A valid final submission with the correct feasibility judgment;
\textbf{(ii)} complete daily selections with the required number and order of activities, where each selected place is uniquely identified by its name and coordinates;
\textbf{(iii)} the complete itinerary matches a saved acceptable itinerary.
&
Whether the selected places form a complete and acceptable ordered itinerary across multiple days.
\\

\bottomrule
\end{tabularx}
\end{table}

\paragraph{Overall long-horizon correctness.}
We evaluate itinerary correctness at the level of the complete plan rather than individual place selections. A successful itinerary must include the required activities in the correct order, with each selected place uniquely identified, while satisfying the constraints imposed by opening hours, travel connections, activity durations, and fixed commitments. Formally, we define
\[
L = V \land S \land (A \lor E),
\]
where $V$ denotes a valid submission with the correct feasibility judgment, $S$ indicates a complete and correctly ordered selection of places, $A$ denotes a match with a saved acceptable itinerary, and $E$ indicates a verified feasible alternative. This formulation captures the central challenge of long-horizon planning: individually suitable destinations do not necessarily form a workable journey, and success depends on whether all selections remain compatible across activities and days.

\subsubsection{Safety-Aware}

We evaluate safety-aware planning by whether the agent \emph{selects the correct destination}, \emph{satisfies the task and safety constraints}, and \emph{grounds its recommendation in safety information obtained through tool interaction}. In real-world geographic decision-making, a destination that appears suitable may become unsafe or inaccessible due to closures, access restrictions, or travel disruptions. Making a reliable recommendation therefore requires more than identifying a suitable place: the agent must account for relevant safety conditions and support its choice with evidence actually obtained from the environment. Accordingly, we evaluate three complementary components of \textbf{Safety choice and grounding}: \textbf{Place correctness}, \textbf{Constraint compliance}, and \textbf{Evidence grounding}, as summarized in Table~\ref{tab:safety_eval_metrics}.

\begin{table}[t]
\centering
\small
\renewcommand{\arraystretch}{1.18}
\caption{Evaluation components for Safety-Aware.}
\label{tab:safety_eval_metrics}
\begin{tabularx}{\linewidth}{
@{}>{\raggedright\arraybackslash}p{2.15cm} X p{4.0cm}@{}
}
\toprule
\textbf{Indicator} & \textbf{Criterion} & \textbf{What it measures} \\
\midrule

\textbf{Place correctness}
&
The submitted place name and coordinates uniquely identify the target destination in the task world.
&
Whether the agent selects the correct destination.
\\

\textbf{Constraint compliance}
&
The selected destination passes all applicable task and safety constraints, including event-adjusted availability and travel-time conditions.
&
Whether the destination satisfies the user's requirements under the relevant safety conditions.
\\

\textbf{Evidence grounding}
&
\textbf{(i)} The agent obtains all required safety notices through the designated Apps;
\textbf{(ii)} reports their complete descriptions verbatim;
\textbf{(iii)} provides no safety information unsupported by recorded tool responses.
&
Whether the recommendation is grounded in safety evidence actually obtained and faithfully reported by the agent.
\\

\bottomrule
\end{tabularx}
\end{table}

\paragraph{Overall safety-aware correctness.}
We evaluate safety-aware correctness by jointly considering the destination choice, its compliance with safety constraints, and the evidence supporting the recommendation. A successful response must not only identify the correct destination and satisfy all applicable constraints, but also obtain and faithfully report the required safety notices through tool interaction. Formally, we define
\[
S = F \land P \land C \land E,
\]
where $F$ denotes a valid positive feasibility declaration, $P$ indicates Place correctness, $C$ indicates Constraint compliance, and $E$ indicates Evidence grounding. This formulation distinguishes making a correct destination choice from supporting that choice with verifiable safety information: neither selecting the correct place without the required evidence nor reporting accurate safety information for an incorrect destination is sufficient for success.

\subsubsection{Region-Aware}

We evaluate region-aware decision-making by whether the agent \emph{selects an acceptable destination under the user's requirements and the relevant regional context}. In everyday place search, a destination that satisfies the user's general preferences may still be inappropriate due to local customs, practices, or regional conditions. Thus, making a suitable choice requires the agent to consider not only the attributes of a place, but also the regional context in which it is situated. Accordingly, we use \textbf{Contextual choice correctness} as the primary indicator, as summarized in Table~\ref{tab:region_eval_metrics}.

\begin{table}[t]
\centering
\small
\renewcommand{\arraystretch}{1.18}
\caption{Evaluation indicator for Region-Aware.}
\label{tab:region_eval_metrics}
\begin{tabularx}{\linewidth}{@{}>{\raggedright\arraybackslash}p{2.15cm} X p{4.0cm}@{}}
\toprule
\textbf{Indicator} & \textbf{Criterion} & \textbf{What it measures} \\
\midrule

\textbf{Contextual choice correctness}
&
\textbf{(i)} The agent submits \texttt{feasible: true} and a final destination identified by its name and coordinates;
\textbf{(ii)} the selected destination matches an entry in the private acceptable-answer set.
&
Whether the selected destination satisfies both the user's requirements and the relevant regional context.
\\

\bottomrule
\end{tabularx}
\end{table}

\paragraph{Overall region-aware correctness.}
We evaluate region-aware correctness based on whether the agent selects a destination that satisfies both the user's requirements and the relevant regional context. Formally, we define
\[
R = F \land (p \in \mathcal{A}),
\]
where $F$ denotes a positive feasibility declaration, $p$ represents the selected place identified by its name and coordinates, and $\mathcal{A}$ is the private acceptable-answer set established during task construction.

\subsubsection{Dynamic}
We evaluate dynamic decision-making at two complementary levels: whether the agent \emph{obtains and correctly reports relevant world changes}, and whether it \emph{makes a correct decision after those changes}. For example, learning that a restaurant has closed is not enough if the agent still recommends it; conversely, choosing a valid alternative does not establish that the agent actually observed the closure. Accordingly, we define two indicators, \textbf{Event grounding} and \textbf{Replanning}, as summarized in Table~\ref{tab:dynamic_eval_metrics}. Together, they distinguish awareness of world changes from the ability to incorporate those changes into valid decisions. The same criteria apply to Task-Only and Shared Dynamics, although the latter introduces additional world changes that may be unrelated to the current task.

\begin{table}[t]
\centering
\small
\renewcommand{\arraystretch}{1.18}
\caption{Evaluation indicators for Dynamic.}
\label{tab:dynamic_eval_metrics}
\begin{tabularx}{\linewidth}{@{}>{\raggedright\arraybackslash}p{2.7cm} X@{}}
\toprule
\textbf{Indicator} & \textbf{Criterion} \\
\midrule

\textbf{Event grounding}
&
The agent observes all required task events after the initial checkpoint and faithfully reports their complete public messages using evidence obtained through tool interactions.
\\

\textbf{Replanning}
&
The agent's final decision matches the verified post-event reference answer for the updated world state.
\\

\bottomrule
\end{tabularx}
\end{table}

\paragraph{Overall dynamic correctness.}
Dynamic evaluation proceeds through a sequence of decisions within a virtual planning window. The agent first submits an intermediate plan and specifies a later query time. As intermediate submissions advance virtual time, world changes take effect, and the agent may query the environment again, revise its plan, and submit a final answer. The initial checkpoint records the agent's starting decision but is not required to match the construction-side initial reference answer.

The two indicators assess distinct requirements. \textbf{Event grounding} requires the agent to observe every task-bound event through its corresponding App after the initial checkpoint and reproduce each complete public message verbatim in \texttt{plan.event\_information}. All reported messages must be supported by recorded tool observations; paraphrases or unsupported statements do not count. Under Shared Dynamics, only events bound to the current task are required, rather than every change occurring in the shared world. \textbf{Replanning} evaluates whether the final decision matches the verified post-event reference answer. For Mobility Planning, this comparison checks the Boolean feasibility judgment; for other task categories, it checks the selected place or ordered place sequence. This indicator therefore measures agreement with the verified target rather than independently rechecking every public constraint.

Overall dynamic success requires both indicators to pass, together with the necessary submission and evaluation checks. Let $O$, $E$, and $P$ denote Structured output, Event grounding, and Replanning, respectively, and let $q$ denote the evaluator's rounded percentage of passed checks among those applicable to the run. These additional checks cover task-transition validation, required notification attempts, and tool-call efficiency. Task-transition validation verifies the constructed task rather than the agent's capability. For a normally evaluated run, overall correctness is defined as
\[
D = O \land E \land P \land (q \geq 70).
\]
Thus, an agent must both obtain the necessary world updates and produce the correct final decision.

\subsection{Benchmark Statistics}
\label{app:benchmark_statistics}

\paragraph{Geographic and task coverage.}
Table~\ref{tab:benchmark_city_distribution} complements the aggregate statistics in the main text with the city-level composition of the benchmark. The completed task collection contains 1,088 static instances---478 Atomic Geo Tasks and 610 Compositional Geo Tasks---and 149 matched dynamic instances.

Every static category is represented in all six cities, although the distribution is not uniform: static coverage ranges from 140 instances in Tokyo to 230 in Cape Town. Mobility Planning and Multi-Entity are relatively evenly distributed across cities, whereas Long-Horizon and Safety-Aware have larger concentrations in Cape Town. We report these counts explicitly to distinguish capability coverage from geographic balance; aggregate task scores should be interpreted alongside this composition.

\begin{table*}[t]
\centering
\small
\setlength{\tabcolsep}{4pt}
\caption{City-level distribution of LiveGeo instances. Dynamic instances are counted once, irrespective of their event-scope setting.}
\label{tab:benchmark_city_distribution}
\begin{tabular}{lrrrrrrr}
\toprule
\textbf{Task category} & \textbf{Shanghai} & \textbf{Tokyo} & \textbf{Singapore} & \textbf{London} & \textbf{New York} & \textbf{Cape Town} & \textbf{Total} \\
\midrule
Grounding & 21 & 18 & 40 & 20 & 10 & 23 & 132 \\
Discovery & 42 & 27 & 26 & 28 & 18 & 30 & 171 \\
Mobility Planning & 30 & 26 & 30 & 30 & 30 & 29 & 175 \\
Multi-Entity & 30 & 30 & 30 & 27 & 29 & 30 & 176 \\
Long-Horizon & 30 & 6 & 26 & 18 & 19 & 51 & 150 \\
Safety-Aware & 8 & 3 & 16 & 27 & 20 & 47 & 121 \\
Region-Aware & 26 & 30 & 30 & 27 & 30 & 20 & 163 \\
\midrule
Static subtotal & 187 & 140 & 198 & 177 & 156 & 230 & 1,088 \\
Dynamic & 24 & 26 & 21 & 27 & 22 & 29 & 149 \\
\midrule
\textbf{Total} & \textbf{211} & \textbf{166} & \textbf{219} & \textbf{204} & \textbf{178} & \textbf{259} & \textbf{1,237} \\
\bottomrule
\end{tabular}
\end{table*}

\paragraph{Variation within task categories.}
The collection also spans different decision structures within each category. Discovery includes 63 restaurant, 42 hotel, 33 caf\'e, and 33 attraction requests. Mobility Planning includes 104 driving, 33 walking, and 38 mixed-mode instances, according to the stored task-level mode labels. Multi-Entity involves two to eight participants, while Long-Horizon spans two to five days. Region-Aware covers 55 requested place types. These descriptors characterize the range of instantiated requests rather than assigning difficulty: a larger group or longer itinerary does not, by itself, establish that an instance is harder.

\begin{figure*}[t]
    \centering
    \includegraphics[width=\textwidth]{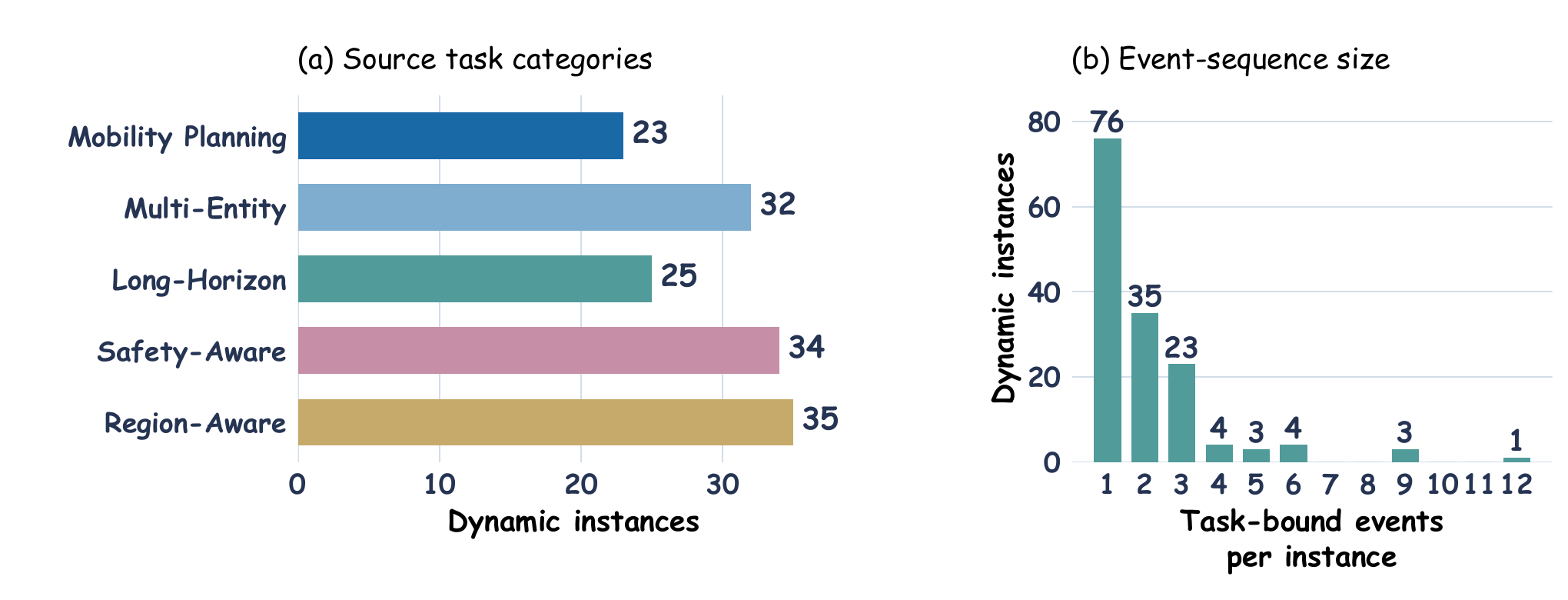}
    \caption{Composition of the dynamic task set: (a) source static-task categories and (b) task-bound event counts per instance. Both panels count task instances, not evaluation runs; additional shared-world events are excluded from panel (b).}
    \label{fig:benchmark_dynamic_composition}
\end{figure*}

\paragraph{Composition of dynamic episodes.}
Figure~\ref{fig:benchmark_dynamic_composition} details the 149 dynamic instances: 23 extend Mobility Planning, 32 Multi-Entity, 25 Long-Horizon, 34 Safety-Aware, and 35 Region-Aware. Together, they contain 309 task-bound event records, with one to twelve events per instance; 111 instances contain one or two events. The public observation channels are MessagingApp (192 records), LocalServiceApp (95), and CityNewsApp (22). User-constraint updates account for 188 records, making changes to the requested decision conditions more prevalent than other event types in the current collection. These figures describe task-bound events, not all background events available in the shared world. The same 149 instances are reused in both dynamic event-scope settings: \texttt{task\_only} retains their own bound events, whereas \texttt{shared\_world} loads these same events together with other events in the bound world snapshot. In both settings, event effects become observable as virtual time advances, and the agent must actively query Apps; neither setting supplies the event list directly. The settings therefore change the event environment, not the membership or size of the dynamic task set.

\section{Experiment}

\subsection{Experimental Setup Details}
\label{app:experimental_setup}

\paragraph{Agent Interaction Protocol.}
We evaluate all models using a unified ReAct-based agent scaffold. At initialization, the agent receives the task-specific system prompt, the available App catalog, and the user's original request. The agent interacts with GeoWorld through a text-based Thought--Action--Observation loop. At each turn, it generates a visible \texttt{Thought} followed by an \texttt{Action} containing a JSON-formatted tool name and its arguments. Each turn executes at most one tool call. The returned tool result is appended to the interaction history as an \texttt{Observation}, which the agent uses to determine its next action. Tool schemas are progressively exposed as the agent opens the corresponding Apps. We retain the complete visible interaction history throughout execution without automatic context truncation. Task completion requires an explicit submission through the designated \texttt{final\_answer} interface, rather than simply ending the model's response. The final submission is subsequently evaluated by the corresponding task-specific Judge.

\paragraph{Task-Specific Output Contracts.}
We use task-specific output contracts to standardize the required decision content across models evaluated on the same task. Although most tasks share a common submission schema, their semantic requirements differ according to the geographic decision being evaluated. Table~\ref{tab:output_contracts} summarizes the required outputs for the seven task categories.

\begin{table}[t]
\centering
\small
\renewcommand{\arraystretch}{1.3}
\setlength{\tabcolsep}{6pt}
\caption{Task-specific output requirements in LiveGeo.}
\label{tab:output_contracts}
\begin{tabularx}{\linewidth}{
    @{}
    >{\raggedright\arraybackslash}p{2.55cm}
    >{\raggedright\arraybackslash}X
    @{}
}
\toprule
\textbf{Task Category} & \textbf{Required Output} \\
\midrule
Grounding
& Explicitly select a grounded location candidate before submitting the final answer. \\

Discovery
& Report feasibility and explanation, together with the selected place and its geographic coordinates. \\

Mobility Planning
& Report trip feasibility, selected route identifiers, transportation mode, and supporting explanation. \\

Multi-Entity
& Provide the selected places and the required multi-entity coordination decisions. \\

Long-Horizon
& Provide a day-wise itinerary with selected POIs and their activity order. \\

Safety-Aware
& Provide the selected place and supporting safety information from observed messages. \\

Region-Aware
& Select a place and explain its suitability under the specified regional context and user requirements. \\
\bottomrule
\end{tabularx}
\end{table}

\paragraph{Dynamic Interaction and Information Access.}
For dynamic tasks, the agent operates within the same ReAct-based interaction framework, but must additionally maintain and update its decisions as the world evolves. In the current dynamic interface, the agent first submits an intermediate answer containing its initial plan (P0) and a proposed next query time. GeoWorld then advances the virtual clock and applies the scheduled events that have become effective. The agent can subsequently query the Apps to acquire updated information and revise its decisions before submitting the final answer (P1). The final submission includes the relevant event information supporting its updated decision.

We further distinguish two information access modes. In our main dynamic settings, both \textit{Task-Only Dynamics} and \textit{Shared Dynamics} use a \textit{Pull-based} mode, in which agents must actively query GeoWorld Apps to discover available updates. As an additional setting, we introduce \textit{Explicit Update}, in which publicly available event notifications are directly appended to the system context as they become available, allowing agents to receive updates without having to discover them through App queries.

\paragraph{Model Integration and Runtime Configuration.}
We evaluate nine representative models through their corresponding API interfaces. Table~\ref{tab:model_config} lists their displayed names, actual API model identifiers, and request interfaces.

\begin{table}[t]
\centering
\small
\setlength{\tabcolsep}{5pt}
\renewcommand{\arraystretch}{1.15}
\caption{Model identifiers and API interfaces.}
\label{tab:model_config}
\begin{tabular}{@{}l l c@{}}
\toprule
\textbf{Model} & \textbf{API Model ID} & \textbf{Interface} \\
\midrule
GPT-5.6 Sol & \texttt{gpt-5.6} & Responses \\
Claude Opus 5 & \texttt{claude-opus-5} & Chat \\
Claude Sonnet 5 & \texttt{claude-sonnet-5} & Chat \\
Gemini 3.1 Pro Preview & \texttt{gemini-3.1-pro-preview} & Chat \\
Gemini 3.7 Flash & \texttt{gemini-3.7-flash} & Chat \\
Grok 4.5 & \texttt{grok-4.5} & Chat \\
Qwen3.8-Max & \texttt{qwen3.8-max} & Chat \\
Kimi K3 & \texttt{kimi-k3} & Chat \\
DeepSeek V4 Flash Exp & \texttt{deepseek-v4-flash-vision-exp} & Chat \\
\bottomrule
\end{tabular}
\end{table}

\subsection{Failure Attribution}
\label{app:failure_attribution}

\begin{figure}[t]
    \centering
    \includegraphics[width=\linewidth]{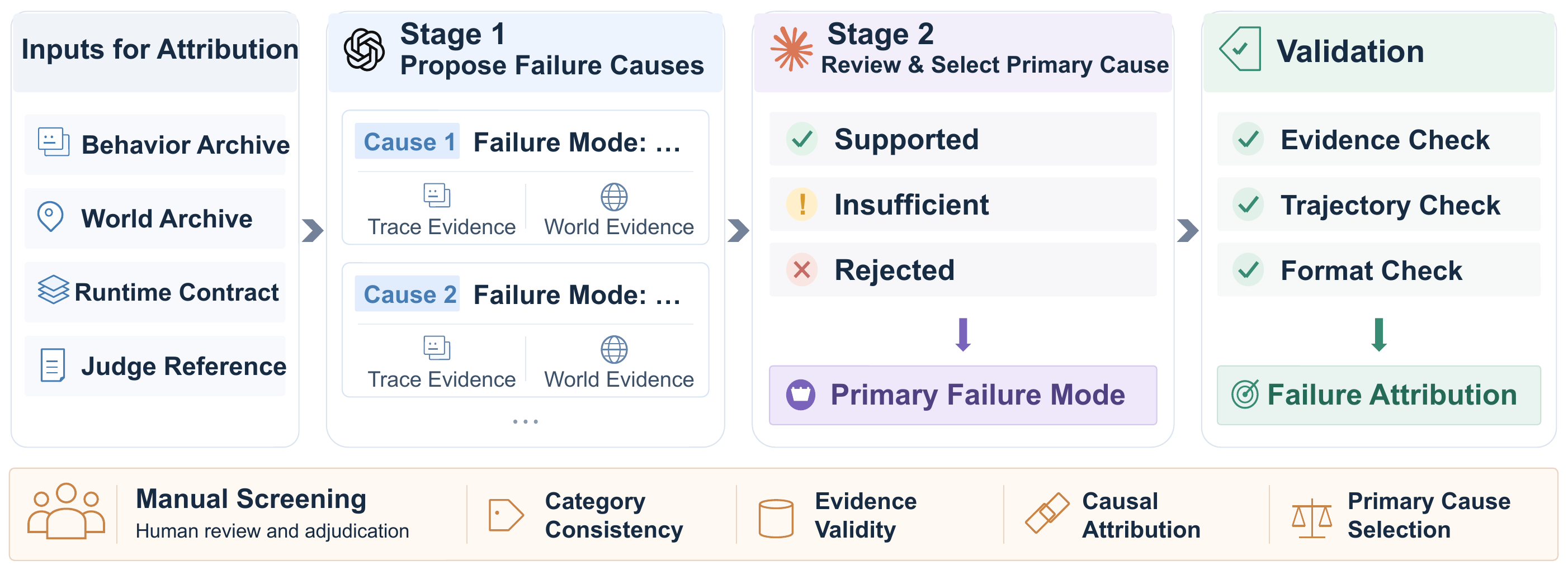}
    \caption{
        Overview of the failure attribution process.
        Stage 1 proposes candidate failure causes based on
        trace and world evidence. Stage 2 reviews the
        candidates and selects the primary failure mode.
        Validation checks evidence, trajectory completeness,
        and output format.
    }
    \label{fig:failure_attribution}
\end{figure}

We attribute each failed execution by examining the agent's
interaction trajectory against the geographic information
available during execution. As illustrated in
Figure~\ref{fig:failure_attribution}, the process consists
of two stages: proposing candidate failure causes and
reviewing them to identify the primary cause, followed
by validation of the attribution outputs. We use seven
predefined failure categories for Static settings and
five for Dynamic settings. Table~\ref{tab:failure_categories} defines these categories
and the behavioral errors they explain.

\paragraph{Failure Categories.}
Table~\ref{tab:failure_categories} summarizes the behavioral criteria
for the seven Static and five Dynamic failure categories.

\begin{table*}[t]
\centering
\small
\renewcommand{\arraystretch}{1.15}
\caption{Operational definitions of the failure categories.}
\label{tab:failure_categories}
\begin{tabularx}{\textwidth}{@{}>{\raggedright\arraybackslash}p{0.28\textwidth} >{\raggedright\arraybackslash}X@{}}
\toprule
\textbf{Category} & \textbf{Explanation} \\
\midrule
\multicolumn{2}{@{}l}{\textit{Static Settings}} \\
Landmark Bias
& Prefers prominent landmarks over candidates better supported by task evidence. \\

Spatial Reasoning Failure

& Makes incorrect inferences about distance, direction, region, connectivity, or reachability. \\

Visual Perception Failure
& Misinterprets visual features or associates image evidence with the wrong geographic entity. \\

Intent-to-Search Failure
& Translates user requirements into inappropriate search, filtering, or verification conditions. \\

Discriminative Verification Failure
& Misses or misapplies decisive checks needed to distinguish between candidate alternatives. \\

Evidence Integration Failure
& Fails to combine relevant facts, producing a decision inconsistent with their joint implications. \\

Evidence Hallucination
& Bases decisions on factual claims unsupported or contradicted by available evidence. \\

\midrule
\multicolumn{2}{@{}l}{\textit{Dynamic Settings}} \\
Change Monitoring Failure
& Misses accessible task-relevant updates and continues reasoning from outdated information. \\

Task-Relevance Confusion
& Misjudges an observed event's relevance to the task, leading to inappropriate decisions. \\

Event Applicability Misjudgment
& Misapplies events to the wrong entity, time, mode, or participant. \\

Decision Inertia
& Retains a prior decision despite recognizing that a world change has invalidated it. \\

Constraint-Preserving Replanning Failure
& Revises a plan but violates remaining constraints or fails to produce a executable plan. \\
\bottomrule
\end{tabularx}
\end{table*}

\paragraph{Inputs for Attribution.}
For each failed execution, we prepare four complementary materials to establish what the agent did, what information was available, and how its outcome should be evaluated. The \textbf{Behavior Archive} records the agent's interaction trajectory, including its actions, observations, and tool responses. The \textbf{World Archive} reconstructs the corresponding geographic environment, including publicly queryable information and time-dependent changes. Together, they allow us to distinguish information actually observed by the agent from information it could have queried or that is only available for retrospective analysis. The \textbf{Runtime Contract} preserves the prompts and tools provided to the agent, while the \textbf{Judge Reference} provides the ground truth and evaluation criteria. Keeping these materials separate ensures that post-hoc reference information is not mistaken for evidence available to the agent during execution.

\paragraph{Stage 1: Propose Failure Causes.}
In the first stage, GPT-5.6 sol with codex examines the attribution materials and explores the behavior and world archives through three read-only tools (\texttt{archive\_list}, \texttt{archive\_read}, and \texttt{public\_query\_at}). Guided by the predefined failure categories, it identifies potential causes by examining the agent's interaction trajectory and the corresponding geographic evidence. For each proposed cause, it specifies the failure category, describes the error mechanism, and provides supporting evidence from the original records, together with an explanation of how the error contributed to the final failure. The analysis distinguishes information actually observed by the agent from information that was publicly queryable or available only for retrospective reference. When the evidence is insufficient to establish a cause, the model retains it as an uncertain hypothesis and explicitly states the evidence gap. The prompt template used for this stage is provided below.

\begin{promptbox}{Stage 1: Failure Cause Proposal Prompt}
\small
\textbf{Task.}
Analyze why the geographic decision-making agent failed and
propose candidate causes.

\medskip
\textbf{Inputs.}\\
Failure categories: \texttt{\{failure\_categories\}}\\
Behavior archive: \texttt{\{behavior\_archive\}}\\
World archive: \texttt{\{world\_archive\}}\\
Runtime contract: \texttt{\{runtime\_contract\}}\\
Judge reference: \texttt{\{judge\_reference\}}

\medskip
\textbf{Instructions.}
Inspect the complete interaction trajectory, including all images.
Use \texttt{archive\_list} and \texttt{archive\_read} to examine
the original records, and \texttt{public\_query\_at} to check
public information at the relevant execution time.

The failure categories specify their names, definitions, and
distinguishing criteria. Apply these definitions rather than
matching labels superficially. For each candidate, identify
the specific behavior that justifies its category.

Distinguish information actually observed by the agent from
information publicly queryable at that time and information
available only for retrospective reference. Do not treat
queryable or retrospective information as evidence that the
agent observed it.

For each candidate, provide its category, error mechanism,
archive references with verbatim quotations, and an explanation
of how the error contributed to the final failure. State any
evidence gaps or counterevidence, retaining plausible but
unverified causes as uncertain hypotheses.
Do not invent evidence or select a primary cause.

\medskip
\textbf{Output.}
Return the candidate causes according to
\texttt{\{output\_schema\}}.
\end{promptbox}

\paragraph{Stage 2: Review \& Select Primary Cause.}
In the second stage, Claude Opus 5 independently reviews the original attribution materials and the candidate causes proposed in Stage 1, using the same read-only archive tools. Rather than taking the proposed causes at face value, it revisits the original evidence and examines the full interaction trajectory and relevant world information to assess each candidate. Each candidate is assigned one of three verdicts: \textit{supported}, \textit{insufficient}, or \textit{rejected}. Based on this review, the model selects exactly one candidate as the primary failure cause, favoring the explanation with the strongest supporting evidence and clearest connection to the final failure. It must select from the original candidates without introducing new causes or changing their categories.

If one or more candidates are supported, the primary cause is selected from them. If none is supported, the model still selects the most plausible original candidate, but marks the attribution as low-confidence and explicitly indicates that it lacks sufficient evidential support. The original verdicts are preserved, so the selection does not imply that an unsupported candidate has been validated. The prompt template used for this stage is provided below.

\begin{promptbox}{Stage 2: Primary Cause Review Prompt}
\small
\textbf{Task.}
Independently review the proposed failure causes and select
exactly one as the primary cause.

\medskip
\textbf{Inputs.}\\
Failure categories: \texttt{\{failure\_categories\}}\\
Stage 1 candidates: \texttt{\{candidate\_causes\}}\\
Behavior archive: \texttt{\{behavior\_archive\}}\\
World archive: \texttt{\{world\_archive\}}\\
Runtime contract: \texttt{\{runtime\_contract\}}\\
Judge reference: \texttt{\{judge\_reference\}}

\medskip
\textbf{Instructions.}
Treat the Stage 1 candidates as hypotheses, not established
evidence. Use \texttt{archive\_list}, \texttt{archive\_read},
and \texttt{public\_query\_at} to revisit the cited records,
inspect the complete trajectory and all images, and seek
supporting evidence and counterevidence.

Evaluate each candidate against its category definition,
the original records, and its causal connection to the final
failure. Distinguish information actually observed by the
agent from information publicly queryable at the relevant time
and information available only for retrospective reference.

Assign each candidate one verdict:
\emph{supported} if the evidence establishes the proposed
mechanism and its contribution to the failure;
\emph{insufficient} if the cause remains plausible but
necessary evidence is missing;
or \emph{rejected} if the records contradict the proposed
mechanism or causal explanation.
Justify each verdict using archive references and verbatim
quotations, explicitly noting unresolved evidence gaps.

Select exactly one original candidate as the primary cause.
Favor direct evidence, a clear causal connection to the failed
outcome, and an explanation that withstands counterevidence;
do not automatically select the earliest error.
Do not introduce new causes or change candidate categories.

If any candidates are supported, select from those candidates.
Otherwise, select the most plausible original candidate,
mark the attribution as low-confidence, and state that it
lacks sufficient evidential support. Preserve all candidate
verdicts; selection must not upgrade an unsupported verdict.
Explain why the selected cause is preferred over the alternatives.

\medskip
\textbf{Output.}
Return the candidate verdicts, primary candidate identifier,
selection rationale, and confidence according to
\texttt{\{output\_schema\}}.
\end{promptbox}

\paragraph{Validation.}
We apply three validation checks to the outputs of both attribution stages before accepting them. First, \textit{Evidence Check} verifies that every quoted citation appears verbatim in the archive record it references; paraphrased or fabricated quotations are not accepted. Second, \textit{Trajectory Check} requires the attribution model to read the complete text of every required trajectory record and inspect all original images before declaring its analysis complete. If the model declares completion while required materials remain unread, it receives a correction prompt identifying the materials that still need to be inspected. Third, \textit{Format Check} verifies compliance with the required output schema: each candidate must include its reasoning and supporting evidence, and the selected primary cause must correspond to an original Stage 1 candidate with its category unchanged.

Outputs that fail any of these checks receive up to two correction attempts. Outputs that remain invalid are recorded as \textit{Analysis Errors} and excluded from the attribution results. These errors are distinguished from low-confidence attributions, in which the analysis is completed successfully but the available evidence does not sufficiently support the selected cause.

\paragraph{Manual Screening.}
Beyond automated validation, we conduct manual screening of all validated attribution outputs by three reviewers. Two reviewers independently examine each attribution, assessing category consistency, evidence validity, the causal connection between the identified behavior and task failure, and the justification for the selected primary cause. Disagreements are resolved through discussion, with the third reviewer serving as an adjudicator when necessary.

\subsection{Failure Analysis Details}
\label{app:failure_analysis}
To further understand model performance and identify the sources of failure, we conduct a detailed analysis of each LiveGeo task category. We examine task-specific results and agent behaviors to investigate where models encounter difficulties and how their failure patterns differ across tasks.

\subsubsection{Grounding}
\paragraph{Grounding behavior across models.}
\begin{figure*}[t]
    \centering
    \begin{minipage}[t]{0.38\textwidth}
        \centering
        \includegraphics[width=\linewidth]{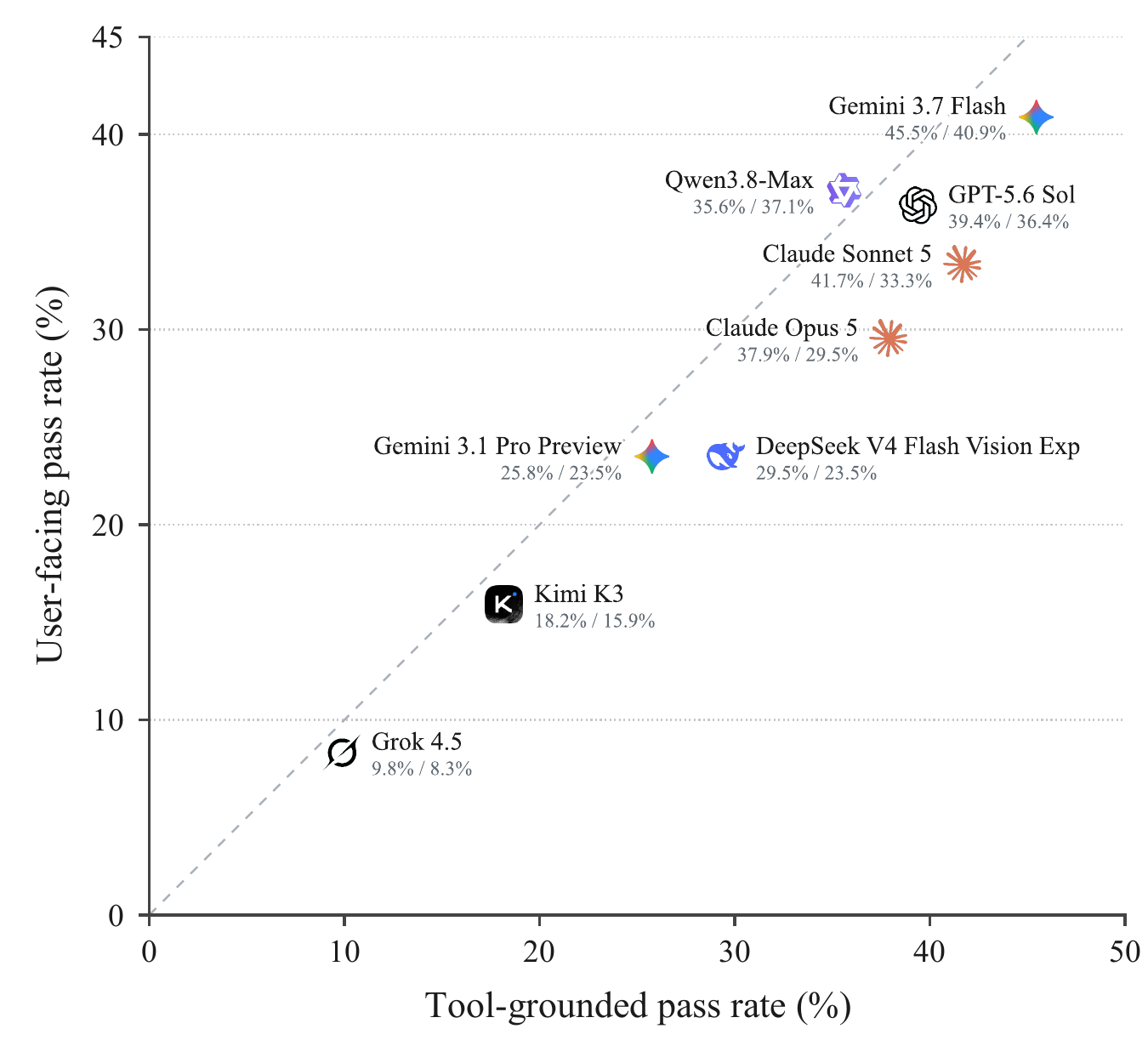}
        \captionof{figure}{Grounding performance across models.}
        \label{fig:grounding_results}
    \end{minipage}
    \hfill
    \begin{minipage}[t]{0.58\textwidth}
        \centering
        \includegraphics[width=\linewidth]{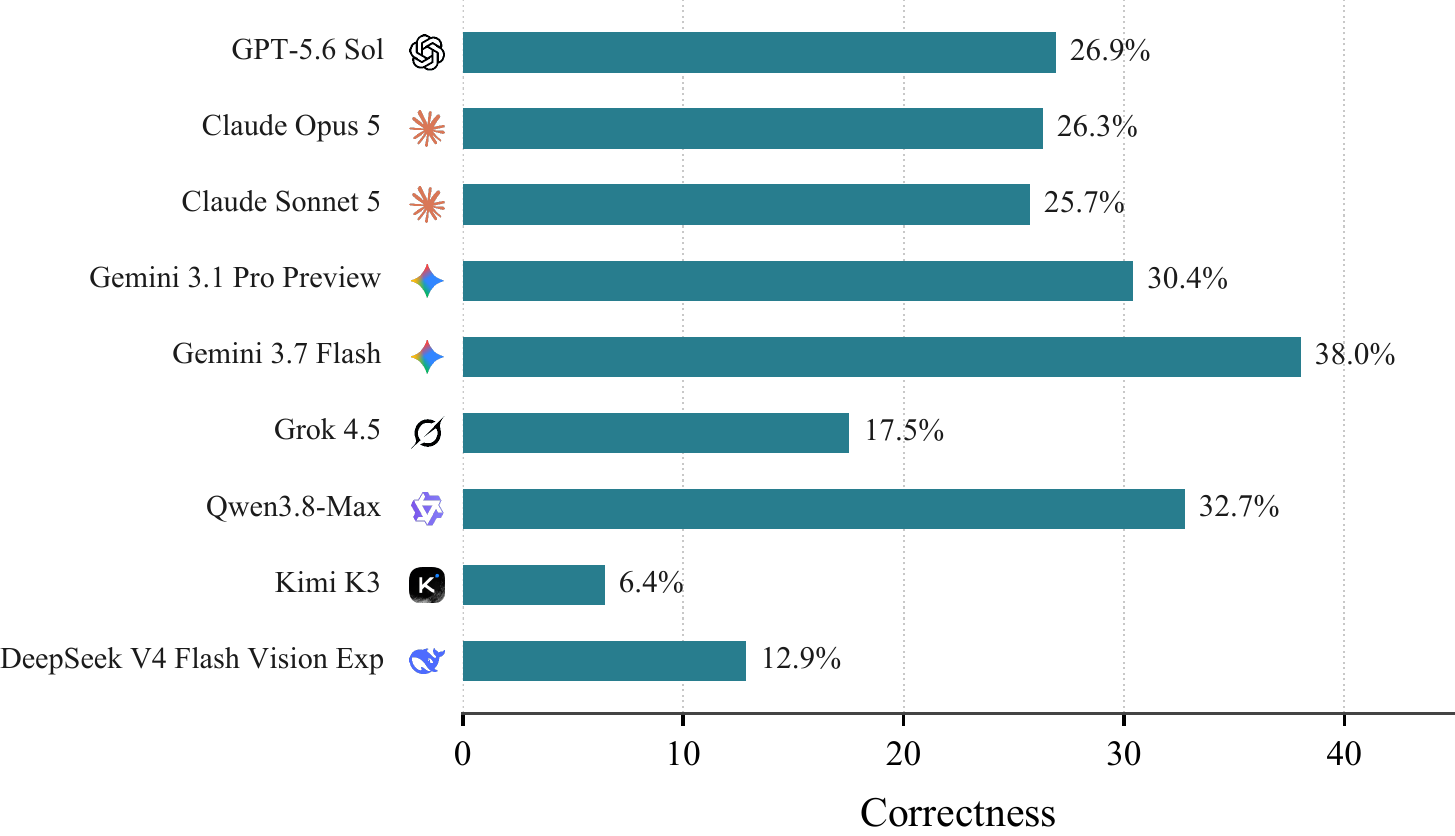}
        \captionof{figure}{Discovery performance across models.}
        \label{fig:discovery_results}
    \end{minipage}
\end{figure*}


Figure~\ref{fig:grounding_results} reveals clear differences in both grounding strength and failure patterns across models. Gemini 3.7 Flash performs best on both Tool-grounded (45.5\%) and User-facing (40.9\%), while Grok 4.5 is lowest on both (9.8\% and 8.3\%). Most models lie below the diagonal, suggesting that selecting the correct location is generally easier than faithfully communicating it to the user. This gap is especially pronounced for Claude Sonnet 5 (41.7\% vs.\ 33.3\%) and Claude Opus 5 (37.9\% vs.\ 29.5\%), which resemble a person who often finds the right place but is less reliable when relaying it. GPT-5.6 Sol and Gemini 3.1 Pro Preview are more balanced across the two stages, while Qwen3.8-Max is the notable exception whose User-facing score (37.1\%) slightly exceeds its Tool-grounded score (35.6\%). Overall, these patterns show that grounding failures can occur at different stages: some models struggle more with selecting the right place, others with communicating it, and successful grounding requires both.

\subsubsection{Discovery}
\paragraph{Discovery behavior across models.}

Figure~\ref{fig:discovery_results} shows a clear separation in end-to-end discovery performance. Gemini 3.7 Flash achieves the highest correctness at 38.0\%, followed by Qwen3.8-Max (32.7\%) and Gemini 3.1 Pro Preview (30.4\%). GPT-5.6 Sol, Claude Opus 5, and Claude Sonnet 5 form a tightly clustered middle group at around 26\%, while Grok 4.5, DeepSeek V4 Flash Vision Exp, and Kimi K3 fall substantially lower, with Kimi K3 reaching only 6.4\%. In everyday terms, stronger models are more often able to carry an open-ended place search through to a concrete, acceptable destination, whereas weaker models frequently fail to complete that process. Their lower scores are also accompanied by frequent protocol terminations in the evaluation trajectories. Since the metric evaluates the full discovery pipeline rather than accuracy conditional on a valid submission, even the best-performing model succeeds on only 65 of 171 instances, highlighting the difficulty of reliably carrying an open-ended place request through to a valid final recommendation.

\subsubsection{Mobility Planning}

\begin{figure*}[t]
    \centering
    \includegraphics[width=\textwidth]{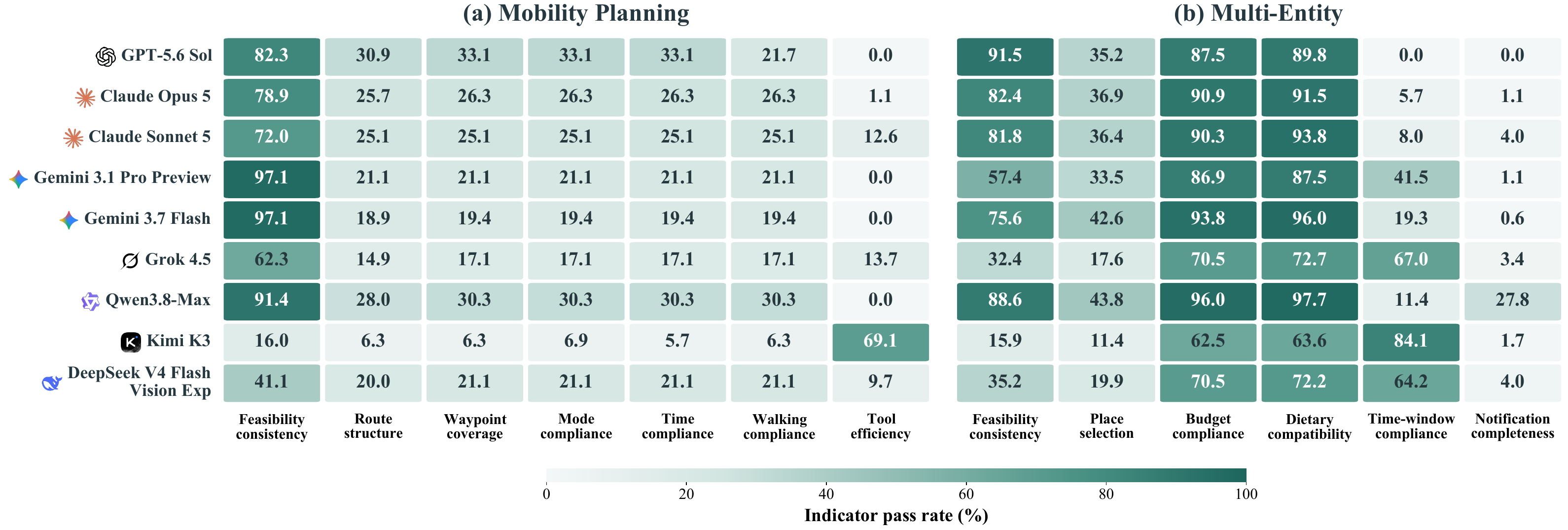}
    \caption{Indicator pass rates for (a) Mobility Planning and (b) Multi-Entity.}
    \label{fig:planning_multi_entity_results}
\end{figure*}

\paragraph{Mobility planning behavior across models.}
Figure~\ref{fig:planning_multi_entity_results}(a) reveals a clear gap between judging whether a trip is feasible and turning that judgment into a valid, constraint-aware journey. Gemini 3.1 Pro Preview and Gemini 3.7 Flash achieve the highest \textbf{Feasibility consistency} at 97.1\%, yet remain around 19--21\% on most route-related indicators, resembling a traveler who can often tell whether a journey should be possible but struggles to turn that judgment into a complete plan. GPT-5.6 Sol and Qwen3.8-Max show more balanced profiles, combining strong feasibility judgments with comparatively higher route-related pass rates. Kimi K3 displays the opposite pattern, with the highest \textbf{Tool efficiency} (69.1\%) but only 16.0\% feasibility consistency and roughly 6\% on most route-related indicators. This contrast shows that using few tool calls does not by itself imply effective planning: successful mobility planning requires both recognizing what is possible and translating that judgment into a verifiable journey that respects the user's constraints.

\subsubsection{Multi-Entity}
\paragraph{Coordination behavior across models.}
Figure~\ref{fig:planning_multi_entity_results}(b) shows substantial variation in how models coordinate multiple participants' requirements. Qwen3.8-Max achieves the highest \textbf{Place selection} rate at 43.8\%, closely followed by Gemini 3.7 Flash at 42.6\%, while Kimi K3 is lowest at 11.4\%. Qwen3.8-Max also combines high feasibility and budget pass rates with the highest \textbf{Notification completeness} (27.8\%), resembling a more attentive group organizer that is comparatively better at carrying a shared decision through to the participants. For most other models, notification completeness remains very low, suggesting that communication is often left unfinished even when other parts of the plan are addressed. The auxiliary indicators should, however, be interpreted cautiously: \textbf{Dietary compatibility} ranges from 63.6\% to 97.7\% across models, while \textbf{Time-window compliance} may receive a default pass when no final answer is submitted. The high time-window rates of Kimi K3, Grok 4.5, and DeepSeek V4 Flash Vision Exp therefore should not be read as evidence of stronger scheduling ability. Overall, even the strongest models select the correct shared destination in fewer than half of the instances, indicating that reconciling multiple people's requirements into a reliable shared decision remains challenging.

\subsubsection{Long-Horizon}
\begin{figure}[t]
    \centering
    \includegraphics[width=\columnwidth]{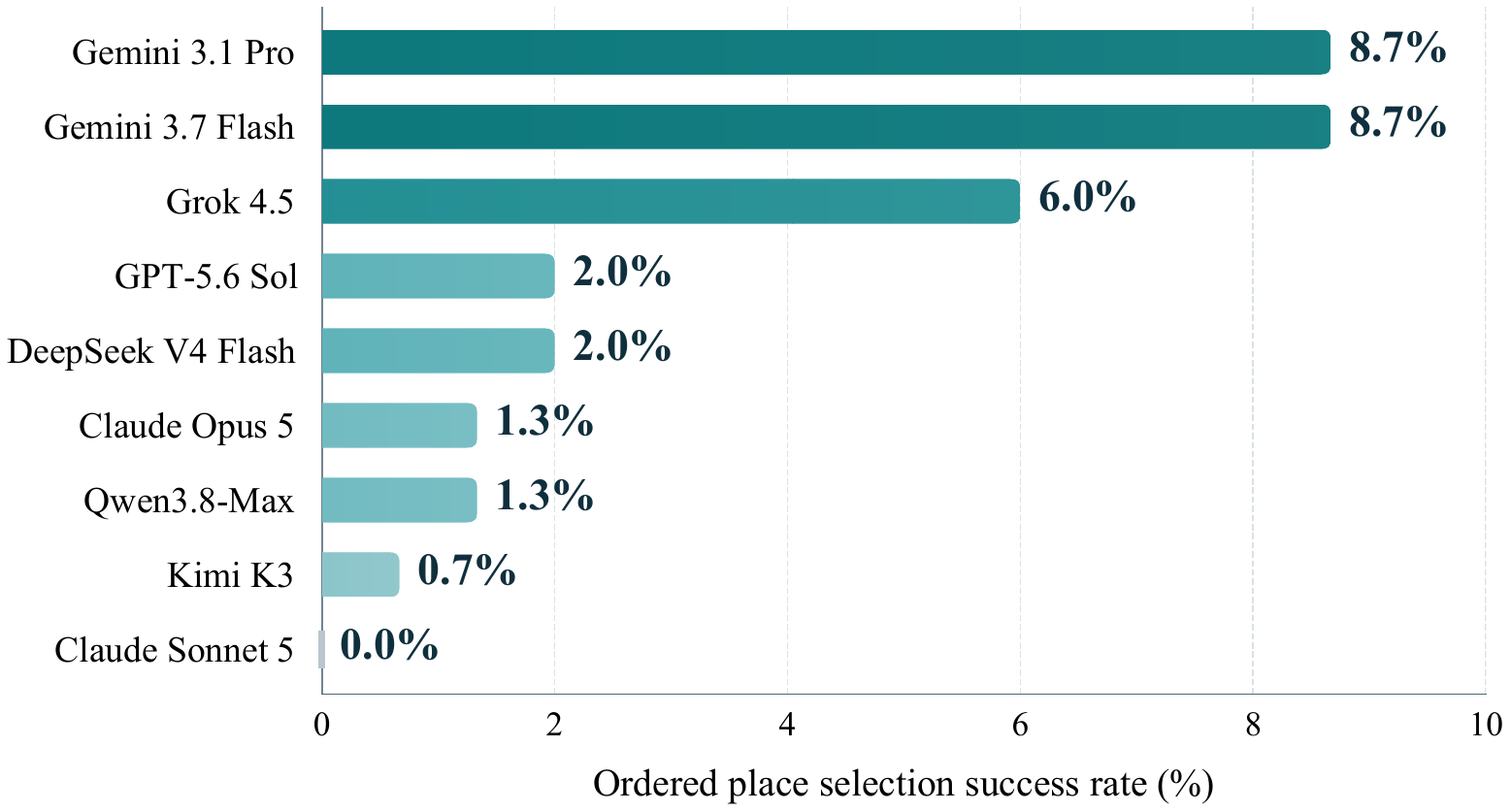}
    \caption{Long-horizon planning performance across nine models. The figure reports the itinerary correctness rate on 150 instances, measuring whether agents construct complete and acceptable multi-day itineraries rather than merely selecting individually suitable places.}
    \label{fig:long_horizon_results}
\end{figure}
\paragraph{Long-horizon behavior across models.}
As shown in Figure~\ref{fig:long_horizon_results}, long-horizon planning remains highly challenging across all evaluated models, with none achieving an itinerary correctness rate above 9\%. Gemini 3.1 Pro and Gemini 3.7 Flash achieve the highest success rate of 8.7\%, followed by Grok 4.5 at 6.0\%. Performance then drops sharply, with the remaining six models achieving no more than 2.0\% correctness. This substantial gap reveals clear differences in long-horizon planning capability, yet the consistently low success rates point to a more fundamental challenge shared across models. Unlike individual place selection, where each decision can be evaluated independently, long-horizon planning requires all selected places to form a mutually compatible sequence across multiple activities and days. A single incompatible choice may invalidate an otherwise reasonable itinerary, making overall correctness considerably more demanding than the suitability of individual destinations. The results thus highlight a critical limitation of current agents: generating plausible geographic choices does not necessarily translate into constructing a complete and acceptable multi-day plan.

\subsubsection{Safety-Aware}
\begin{figure*}[t]
    \centering
    \includegraphics[width=\textwidth]{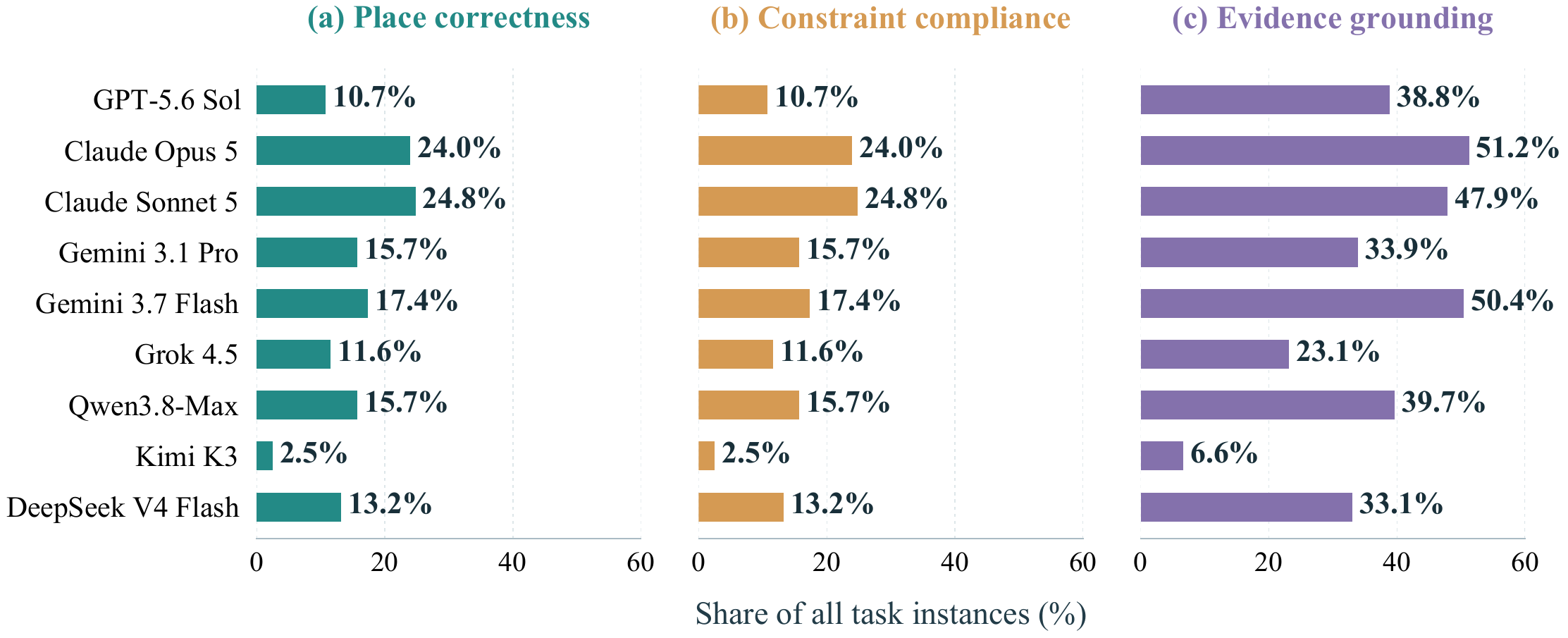}
    \caption{Component-wise performance on Safety-Aware tasks across nine models. We report the proportion of task instances satisfying (a) Place correctness, (b) Constraint compliance, and (c) Evidence grounding. Place correctness and Constraint compliance coincide across all evaluated instances, while Evidence grounding consistently achieves higher rates.}
    \label{fig:safety_components}
\end{figure*}

\paragraph{Safety-aware behavior across models.}
Figure~\ref{fig:safety_components} shows that obtaining and reporting safety evidence is easier for current agents than selecting the correct destination. Across all nine models, Evidence grounding exceeds Place correctness, with Gemini 3.7 Flash showing the largest gap (50.4\% vs. 17.4\%). Place correctness and Constraint compliance coincide because the ground-truth destination is the unique choice satisfying all task and safety constraints. Thus, the central gap is between grounding the required safety evidence and selecting a destination that satisfies those constraints. The reverse mismatch also occurs: agents can select the correct destination without correctly grounding the required evidence. For example, Claude Opus 5 selects the correct destination in 29 instances, but achieves overall success in only 21. Success therefore requires both components, rather than a high score on either alone. Claude Sonnet 5 achieves the highest overall success rate of 19.8\%, despite not leading in Evidence grounding. Together, these results show that reliable safety-aware recommendations require agents both to ground relevant safety information and to use it in selecting a valid destination.

\subsubsection{Region-Aware}
\paragraph{Region-aware behavior across models.}
\begin{figure*}[t]
    \centering
    \includegraphics[width=0.9\textwidth]{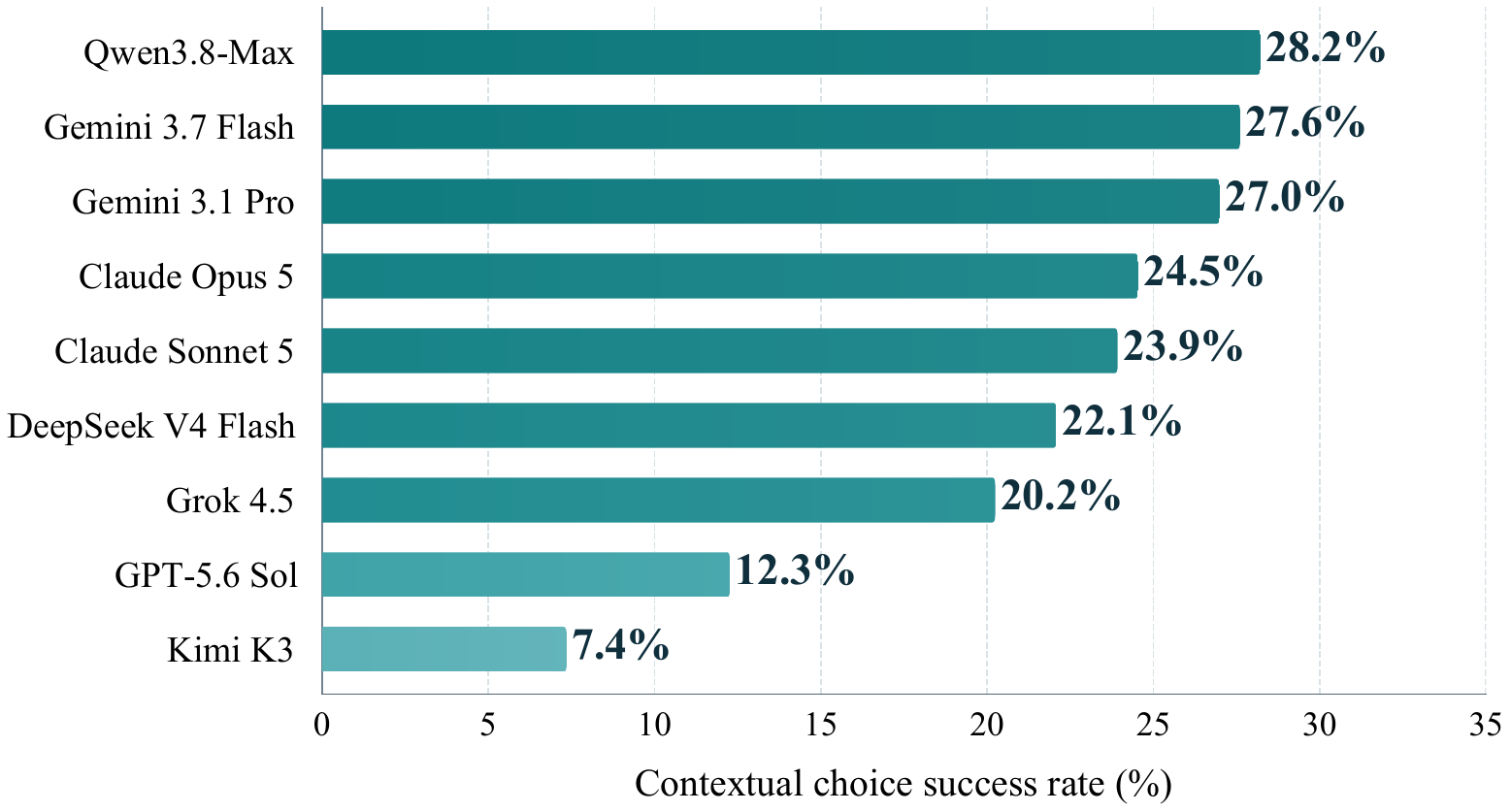}
    \caption{Region-Aware performance across nine models. Bars report Contextual choice correctness rates over 163 task instances per model, measuring whether the selected destinations satisfy both ordinary requirements and regional context.}
    \label{fig:region_aware_results}
\end{figure*}

Figure~\ref{fig:region_aware_results} shows that region-aware decision-making remains challenging across all nine models, with none achieving a Contextual choice correctness rate above 30\%. Qwen3.8-Max achieves the highest success rate at 28.2\%, closely followed by Gemini 3.7 Flash (27.6\%) and Gemini 3.1 Pro (27.0\%), while GPT-5.6 Sol and Kimi K3 fall substantially behind at 12.3\% and 7.4\%, respectively. Despite these performance differences, the low success rates across models reveal a broader difficulty in selecting destinations that satisfy both ordinary requirements and regional context. Notably, our evaluation does not require agents to explain local customs or explicitly report regional knowledge; it only assesses whether their final choices are acceptable. Even under this outcome-focused criterion, more than 70\% of instances remain unsolved by every model. These results show that current agents still struggle to make reliable decisions that account for regional context.

\subsubsection{Dynamic}
\label{sec:dynamic}
\begin{figure*}[t]
    \centering
    \includegraphics[width=\textwidth]{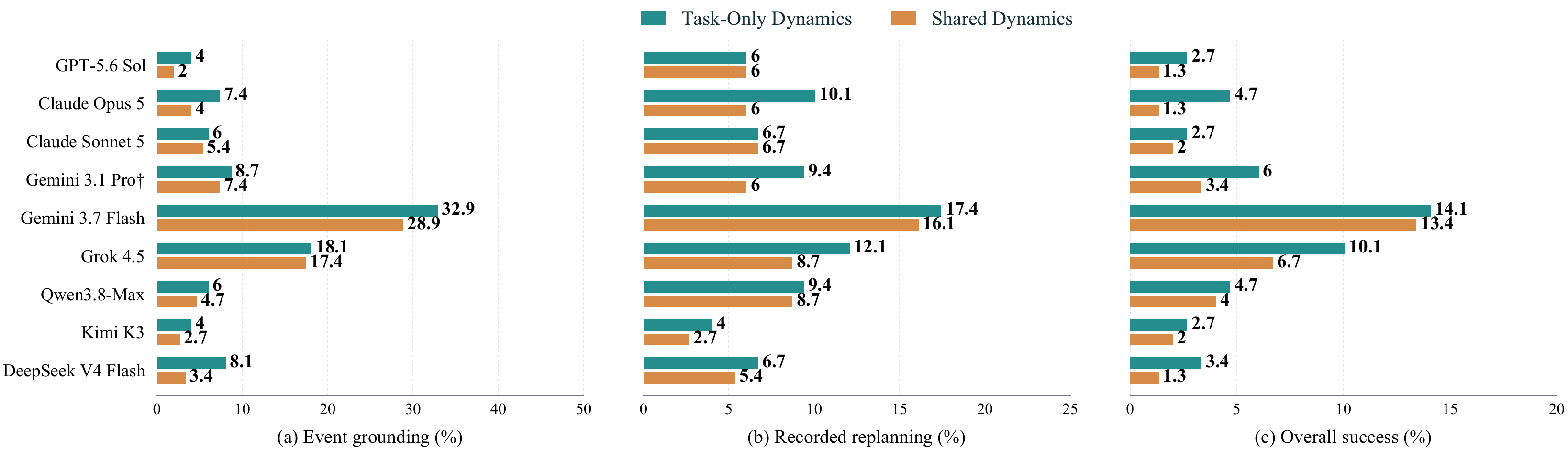}
    \caption{
        \textbf{Event grounding, replanning, and overall success under world dynamics.}
        Comparison of nine models under Task-Only and Shared Dynamics.
        (a) Event grounding measures whether agents observe and faithfully report all required task events.
        (b) Recorded replanning measures whether the final decision matches the verified post-event target, with an initial checkpoint recorded.
        (c) Overall success additionally requires all mandatory evaluation conditions to be satisfied.
        All values are percentages.
    }
    \label{fig:dynamic_evaluation_breakdown}
\end{figure*}

\paragraph{Event grounding and replanning remain challenging under world dynamics.}
Figure~\ref{fig:dynamic_evaluation_breakdown} compares Event grounding, Replanning, and overall success across models under the two dynamic settings. Both indicators reveal substantial limitations in dynamic decision-making. Even Gemini 3.7 Flash, the strongest model, achieves only 32.9\% and 28.9\% Event grounding under Task-Only and Shared Dynamics, respectively, while its Replanning rates are considerably lower at 17.4\% and 16.1\%. Its overall success rates further decrease to 14.1\% and 13.4\%. Most other models perform substantially worse on both indicators. Moving from Task-Only to Shared Dynamics reduces Event grounding across all nine models, while Replanning declines for seven models and remains unchanged for the other two. Notably, the relative performance of the two indicators differs across models: Gemini 3.7 Flash performs considerably better on Event grounding than Replanning, whereas GPT-5.6 Sol and Qwen3.8-Max exhibit the opposite pattern. These results indicate that acquiring and reporting world changes and producing correct post-event decisions present distinct challenges. Since overall success requires both indicators to pass, improving either capability alone does not guarantee reliable decision-making as the world evolves.

\paragraph{Event Types and Task-Level Performance Changes}

To understand why world dynamics affect different task categories to varying degrees, we examine the types of changes encountered by each category and their associated performance declines. 

We classify recorded updates into \textit{qualitative} and \textit{quantitative} changes based on how they modify world states or task constraints. Qualitative changes alter categorical states or requirements, or introduce, remove, or restructure conditions. Quantitative changes adjust numerical values within existing conditions. For example, changing the permitted travel mode is qualitative, whereas modifying an existing travel-time limit is quantitative. Table~\ref{tab:event_change_taxonomy} summarizes the subtypes represented in our dataset.

\begin{table*}[t]
\centering
\small
\setlength{\tabcolsep}{5pt}
\renewcommand{\arraystretch}{1.12}
\caption{Event taxonomy used in the analysis, with illustrative examples. Labels may co-occur within an event or task.}
\label{tab:event_change_taxonomy}
\begin{tabular}{@{}p{0.14\textwidth}p{0.25\textwidth}p{0.55\textwidth}@{}}
\toprule
\textbf{Class} & \textbf{Subtype} & \textbf{Example event} \\
\midrule
Qualitative
 & Place availability & The selected restaurant temporarily closes for maintenance. \\
 & Safety conditions & A new safety notice identifies the planned activity area as unsafe. \\
 & Activity / place type & The user requests a museum instead of a shopping mall. \\
 & Area requirements & The user changes the requested area from downtown to the waterfront. \\
 & Dietary requirements & A participant now requires vegetarian food. \\
 & Service requirements & The user adds a requirement for wheelchair access. \\
 & Travel modes & A participant can no longer drive and must use public transport. \\
 & Opening hours & The user requires the venue to remain open throughout the visit. \\
 & Travel-time limits & The user adds a maximum travel-time requirement to the request. \\
 & Budget requirements & The user removes the spending limit from the request. \\
 & Distance requirements & The user adds a maximum distance requirement from the meeting point. \\
 & Rating requirements & The user adds a minimum rating requirement for the venue. \\
\midrule
Quantitative
 & Traffic delays & A traffic disruption adds 12 minutes to the affected road segment. \\
 & Travel-time limits & The user reduces the existing travel-time limit from 30 to 20 minutes. \\
 & Activity duration & The planned meeting is extended from 20 to 40 minutes. \\
 & Opening hours & The venue moves its closing time from 21:00 to 19:00. \\
 & Budget requirements & The user reduces the existing budget from \$50 to \$30 per person. \\
 & Distance requirements & The user reduces the existing distance limit from 5 km to 2 km. \\
\bottomrule
\end{tabular}
\end{table*}

\begin{figure*}[t]
\centering
\includegraphics[width=\textwidth]{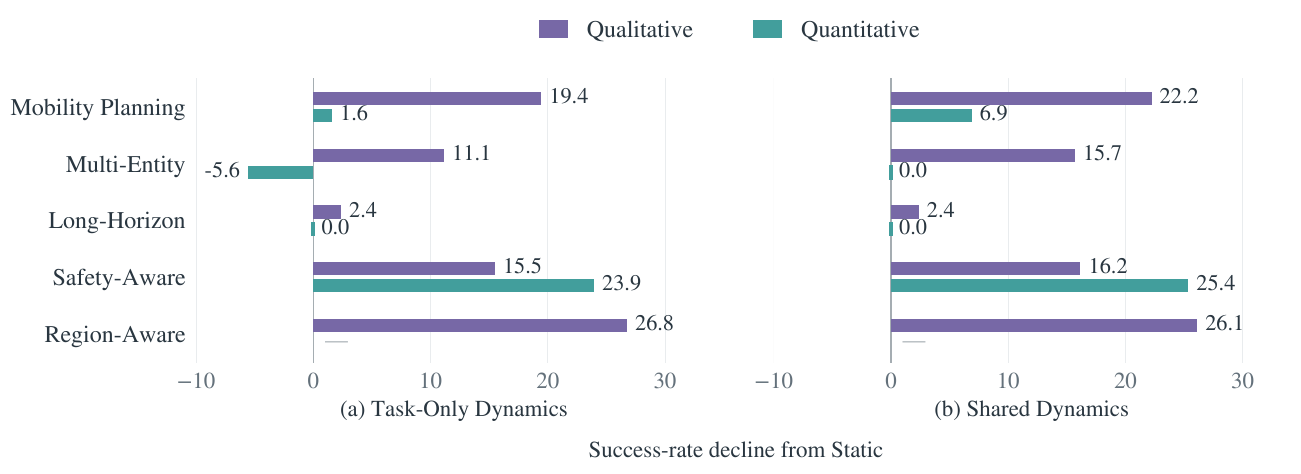}
\caption{\textbf{Performance changes associated with qualitative and quantitative events.} The y-axis shows task categories, and the x-axis shows success-rate decline from Static (pp). Purple and teal bars represent qualitative and quantitative changes, respectively.}
\label{fig:event_class_success_decline}
\end{figure*}

\begin{figure*}[t]
\centering
\includegraphics[width=\textwidth]{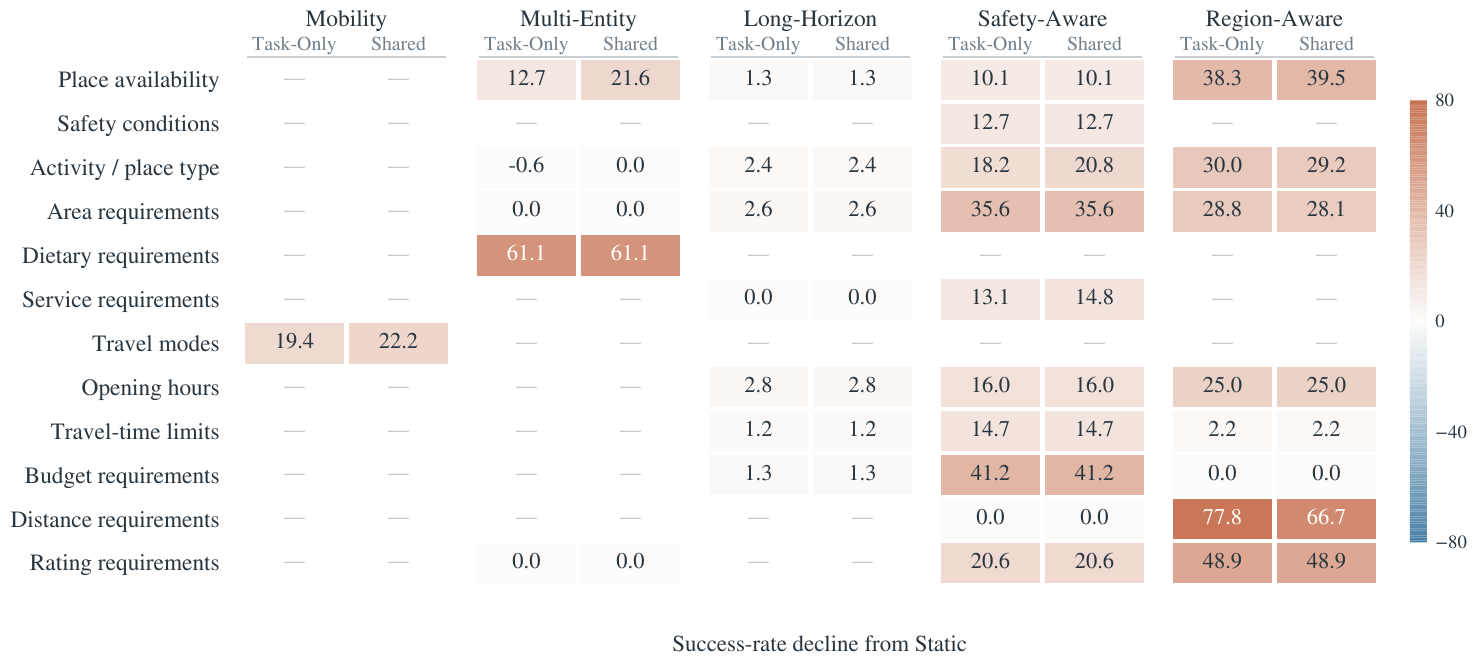}
\caption{\textbf{Performance changes by qualitative event subtype.} The y-axis shows event subtypes, and the x-axis shows task categories under Task-Only and Shared Dynamics. Color intensity indicates the magnitude of success-rate change, with warm and cool colors representing declines and improvements, respectively.}
\label{fig:qualitative_event_subtype_decline}
\end{figure*}

\begin{figure*}[t]
\centering
\includegraphics[width=\textwidth]{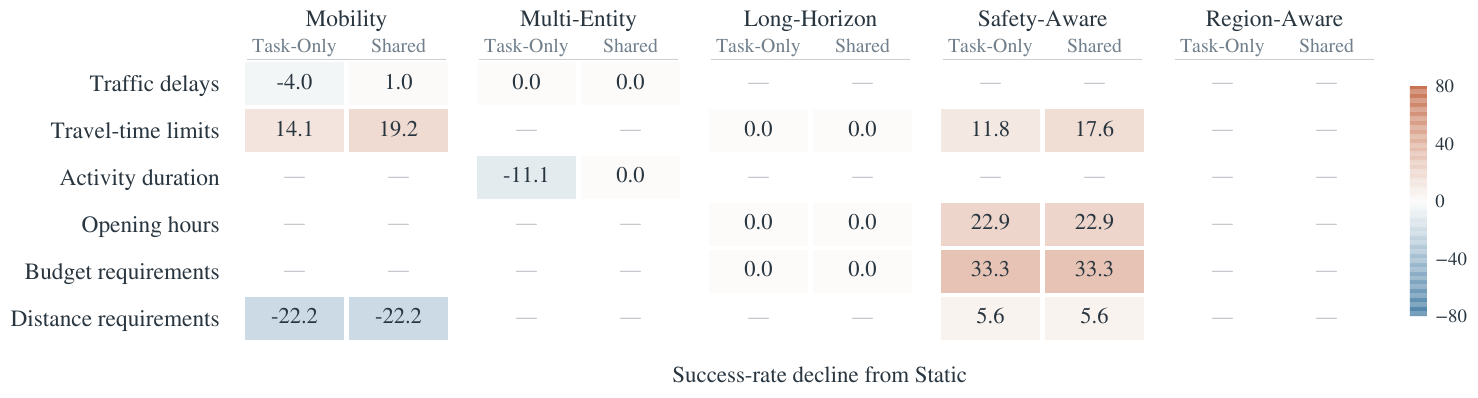}
\caption{\textbf{Performance changes by quantitative event subtype.} The y-axis shows event subtypes, and the x-axis shows task categories under Task-Only and Shared Dynamics. Color intensity indicates the magnitude of success-rate change, with warm and cool colors representing declines and improvements, respectively.}
\label{fig:quantitative_event_subtype_decline}
\end{figure*}

The performance patterns vary with the types of changes encountered by different tasks. Mobility Planning exhibits relatively small declines for tasks containing quantitative changes (1.6 percentage points under Task-Only Dynamics and 6.9 percentage points under Shared Dynamics), compared with substantially larger declines for the four tasks containing qualitative changes (19.4 and 22.2 percentage points, respectively; Figure~\ref{fig:event_class_success_decline}). For traffic-delay events, the corresponding declines are only $-4.0$ and 1.0 percentage points (Figure~\ref{fig:quantitative_event_subtype_decline}). This may be because numerical updates, such as traffic delays, can often be handled by adjusting existing travel-time estimates, whereas changes to permitted travel modes may require new route searches and supporting evidence. By contrast, changes to place availability or user requirements can invalidate previously suitable candidates and alter which alternatives remain feasible. This is particularly evident in Region-Aware, where changes to place availability, activity or place type, and geographic area are associated with declines of 38.3 and 39.5, 30.0 and 29.2, and 28.8 and 28.1 percentage points, respectively, under Task-Only and Shared Dynamics (Figure~\ref{fig:qualitative_event_subtype_decline}). Such changes require agents to reconsider candidate selection and verify alternatives against updated requirements, rather than simply adjust an existing decision.

However, quantitative changes are not necessarily easier to handle. In Safety-Aware, the eight tasks containing quantitative changes exhibit larger declines (23.9 percentage points under Task-Only Dynamics and 25.4 percentage points under Shared Dynamics) than the category average (15.5 and 16.2 percentage points, respectively), although all eight also contain qualitative changes, preventing the effects from being separated. Taken together, these patterns suggest that the impact of world dynamics depends not simply on the form of an update, but on how it interacts with the task's decision structure.

\clearpage
\begingroup
\raggedbottom
\subsection{Case Study}
\label{app:case_study}

\newsavebox{\caseprosebox}
\newsavebox{\casecaptionbox}
\newlength{\caseimageheight}
\newlength{\caseproseheight}
\newsavebox{\casecomparebox}
\newcommand{\compactcase}[6]{%
  \begin{minipage}[t]{0.485\linewidth}
  \vspace{0pt}
  \sbox{\caseprosebox}{\begin{minipage}{\linewidth}
    \fontsize{10}{12}\selectfont
    \textbf{#1}\par\vspace{3pt}#2\par
  \end{minipage}}%
  \sbox{\casecaptionbox}{\begin{minipage}{\linewidth}
    \captionsetup{font={footnotesize},skip=3pt}
    \captionof{figure}{#4}\label{#5}
  \end{minipage}}%
  \setlength{\caseimageheight}{\dimexpr #6-\caseproseheight-\ht\casecaptionbox-\dp\casecaptionbox-8pt\relax}%
  \begin{minipage}[t][\caseproseheight][t]{\linewidth}\usebox{\caseprosebox}\end{minipage}\par\vspace{5pt}
  \centering\includegraphics[width=\linewidth,height=\caseimageheight,keepaspectratio]{#3}\par
  \usebox{\casecaptionbox}
  \end{minipage}%
}
\noindent\begin{minipage}[t]{0.405\linewidth}
\vspace{0pt}\fontsize{10}{12}\selectfont

The aggregate attributions in Section~\ref{sec:failure_modes} identify where failed executions concentrate; the cases below show how those failures arise within an interaction. We examine seven Static and five Dynamic trajectories, tracing the connection between the user's requirements, the evidence available to the agent, and the decision it ultimately submits. The central distinction is between \emph{having information} and \emph{using it as a condition on a decision}. An agent can retrieve the correct place, repeat a relevant fact, or acknowledge a world change while still committing to an incompatible answer.

The figures retain selected executed queries, actual tool feedback, and saved model-thought or final-explanation excerpts; intervening interactions are omitted where they do not clarify the mechanism. Ground truth and explicitly marked posthoc world facts are evaluator-side references, not information presumed to have been observed by the agent. These cases illustrate mechanisms rather than estimate their prevalence. We also distinguish direct evidence of an error from uncertain causal attribution: the mixed-event example establishes an omission without proving distraction, and the replanning example retains contradictory tool feedback rather than attributing the entire failure to the model.

\subsubsection{Static Settings: Forming an Evidence-Grounded Decision}

\textbf{Visual Perception Failure: a mistaken clue persists after the target is retrieved.}\par\vspace{4pt}
In Figure~\ref{fig:case_visual_perception}, Claude Opus 5 reads the visible street-number clue as \texttt{29} rather than \texttt{39}. The consequential error is not confined to that initial reading. MapApp subsequently returns the ground-truth caf\'e, Five Oars Coffee Roasters Heritage, next to the remembered Night Hawk, and a detail lookup confirms the same candidate. Nevertheless, the model continues searching around the misread number. The visual interpretation thus remains a fixed premise instead of a hypothesis to reconcile with retrieved evidence. The run ends with \texttt{place\_lookup\_empty} and no saved final answer; the trace does not establish that the reading error alone caused termination. It does show how an incorrect perceptual clue can redirect further search even after the target has been found.
\end{minipage}\hfill
\begin{minipage}[t]{0.565\linewidth}
\vspace{0pt}\centering
\includegraphics[width=\linewidth]{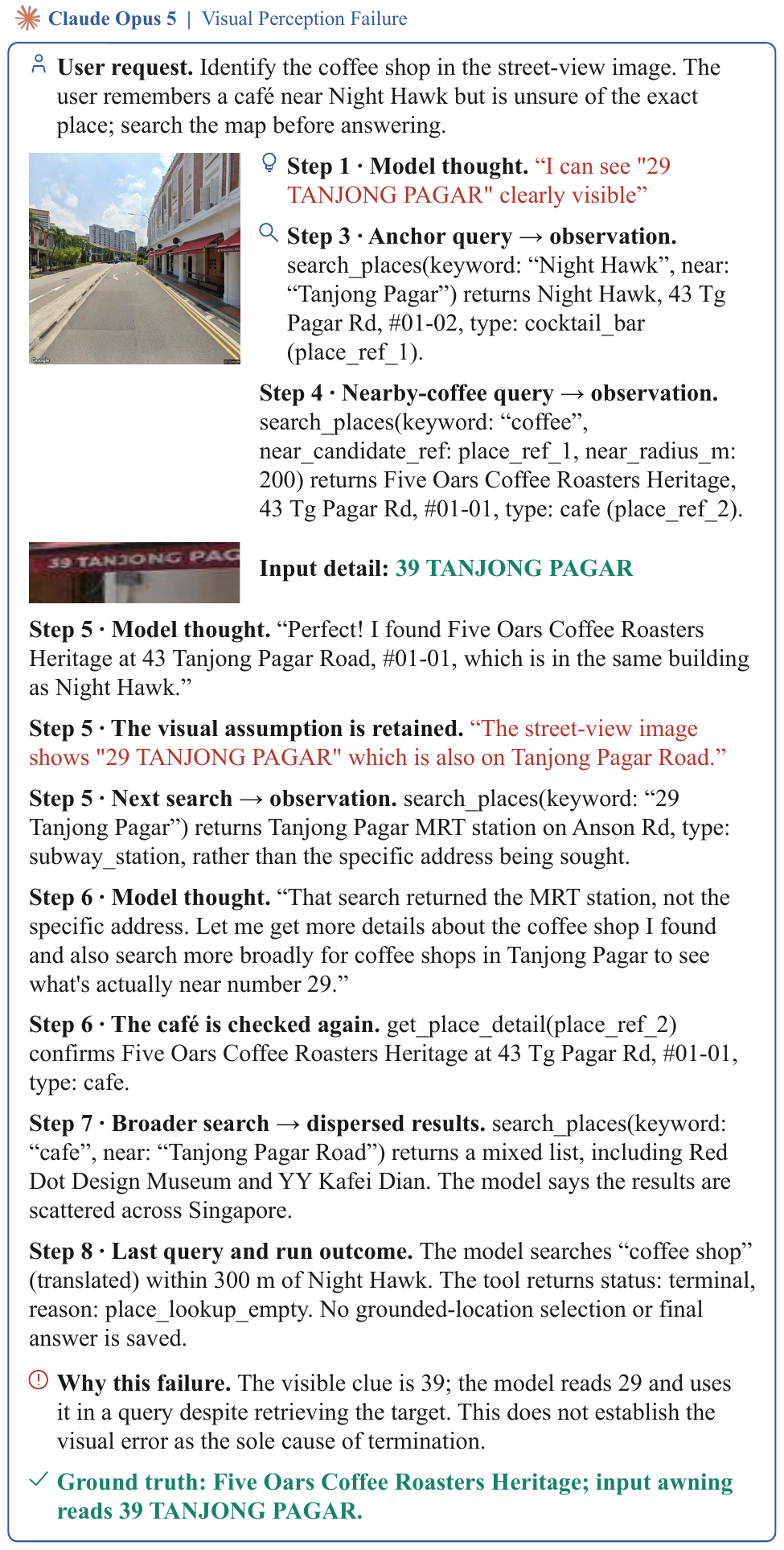}
\captionsetup{font=footnotesize,skip=4pt}
\captionof{figure}{\textbf{Visual Perception Failure} in Static Grounding (Claude Opus 5). The correct caf\'e is retrieved and inspected, but a misread street-number clue continues to direct search. No final answer is saved.}\label{fig:case_visual_perception}
\end{minipage}
\par

\clearpage

\setlength{\caseproseheight}{0pt}
\sbox{\casecomparebox}{\begin{minipage}{0.485\linewidth}\fontsize{10}{12}\selectfont\textbf{Intent-to-Search Failure: a hard requirement becomes a category-level proxy.}\par\vspace{3pt}Figure~\ref{fig:case_intent_to_search} concerns a Discovery request with a non-negotiable gluten-free requirement. Claude Opus 5 initially searches for that requirement directly. After the search returns no results and hotel details provide no explicit dietary support, it switches to ``fine dining western'' and ``Italian French,'' reasoning that these categories are more likely to accommodate dietary restrictions. The final hotel recommendation rests on that general association and asks the user to confirm gluten-free options by telephone. The failure is the weakening of a requirement during its translation into search and selection criteria: a preference for a promising category substitutes for verification of the requested service. This evidence does not prove that the recommended hotel cannot provide gluten-free food; it shows that the agent submits a recommendation while the decisive condition remains unresolved.\par\end{minipage}}
\ifdim\dimexpr\ht\casecomparebox+\dp\casecomparebox\relax>\caseproseheight\setlength{\caseproseheight}{\dimexpr\ht\casecomparebox+\dp\casecomparebox\relax}\fi
\sbox{\casecomparebox}{\begin{minipage}{0.485\linewidth}\fontsize{10}{12}\selectfont\textbf{Landmark Bias: prominence overrides an already identified entity.}\par\vspace{3pt}Figure~\ref{fig:case_landmark_bias} separates candidate retrieval from final entity selection. Claude Opus 5 retrieves both St Pancras International station and St. Pancras London, Autograph Collection, then searches nearby hotels and opens the correct hotel's detail. Its saved thought explicitly recognizes the hotel before choosing the station because it is ``the most prominent landmark in the image.'' The selection tool and final response both commit to the station. Unlike a case in which the target never appears, the relevant entity is available and acknowledged here. The failure occurs when visual prominence replaces the identity established through the hotel evidence. More retrieval alone would not address this transition; the agent must retain the task's target when resolving closely associated geographic entities.\par\end{minipage}}
\ifdim\dimexpr\ht\casecomparebox+\dp\casecomparebox\relax>\caseproseheight\setlength{\caseproseheight}{\dimexpr\ht\casecomparebox+\dp\casecomparebox\relax}\fi
\noindent\compactcase{Intent-to-Search Failure: a hard requirement becomes a category-level proxy.}{Figure~\ref{fig:case_intent_to_search} concerns a Discovery request with a non-negotiable gluten-free requirement. Claude Opus 5 initially searches for that requirement directly. After the search returns no results and hotel details provide no explicit dietary support, it switches to ``fine dining western'' and ``Italian French,'' reasoning that these categories are more likely to accommodate dietary restrictions. The final hotel recommendation rests on that general association and asks the user to confirm gluten-free options by telephone. The failure is the weakening of a requirement during its translation into search and selection criteria: a preference for a promising category substitutes for verification of the requested service. This evidence does not prove that the recommended hotel cannot provide gluten-free food; it shows that the agent submits a recommendation while the decisive condition remains unresolved.}{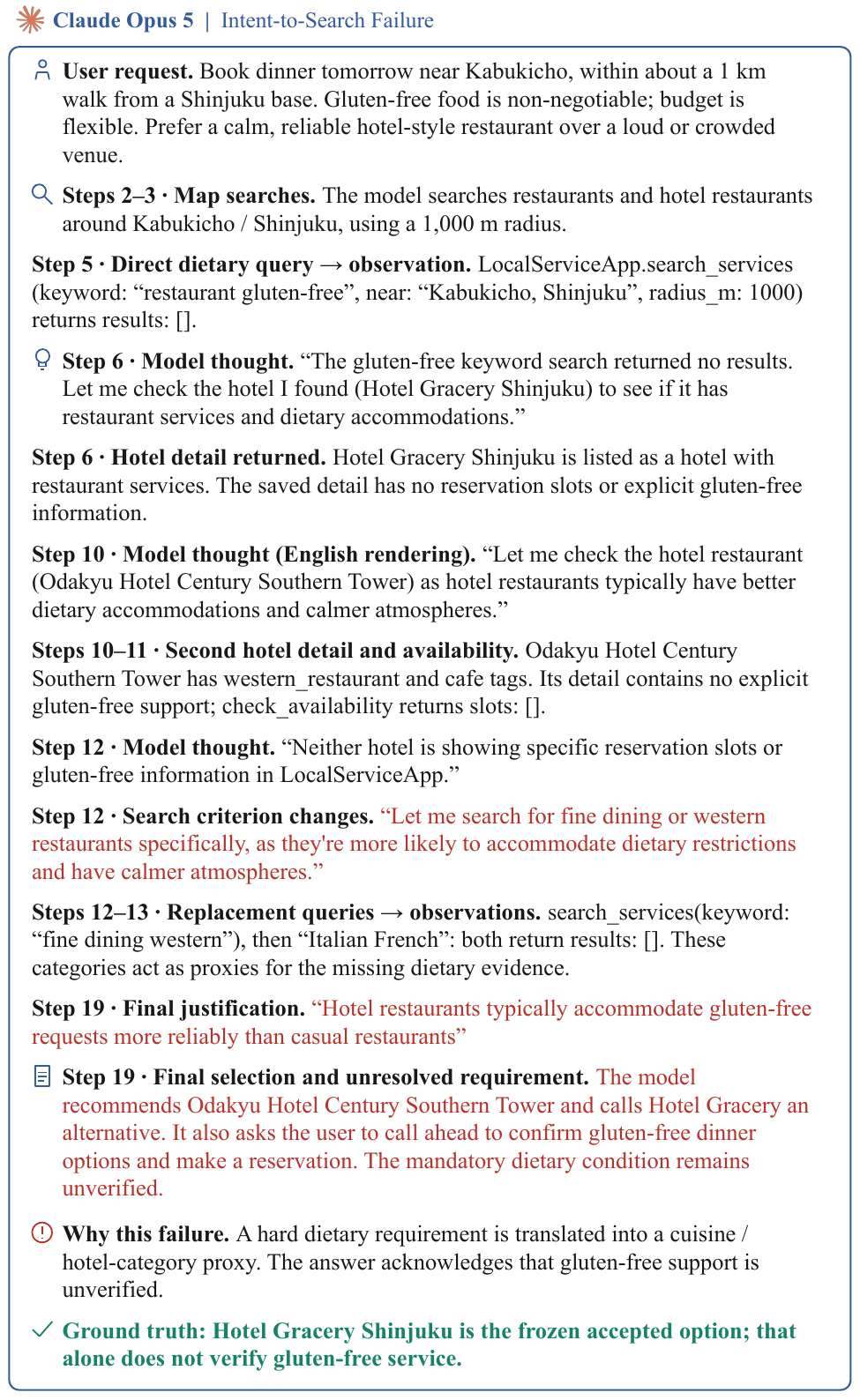}{\textbf{Intent-to-Search Failure} in Static Discovery (Claude Opus 5). Explicit dietary verification gives way to cuisine and hotel-category proxies; the recommendation leaves the gluten-free requirement unverified.}{fig:case_intent_to_search}{\dimexpr\textheight-8pt\relax}\hfill
\noindent\compactcase{Landmark Bias: prominence overrides an already identified entity.}{Figure~\ref{fig:case_landmark_bias} separates candidate retrieval from final entity selection. Claude Opus 5 retrieves both St Pancras International station and St. Pancras London, Autograph Collection, then searches nearby hotels and opens the correct hotel's detail. Its saved thought explicitly recognizes the hotel before choosing the station because it is ``the most prominent landmark in the image.'' The selection tool and final response both commit to the station. Unlike a case in which the target never appears, the relevant entity is available and acknowledged here. The failure occurs when visual prominence replaces the identity established through the hotel evidence. More retrieval alone would not address this transition; the agent must retain the task's target when resolving closely associated geographic entities.}{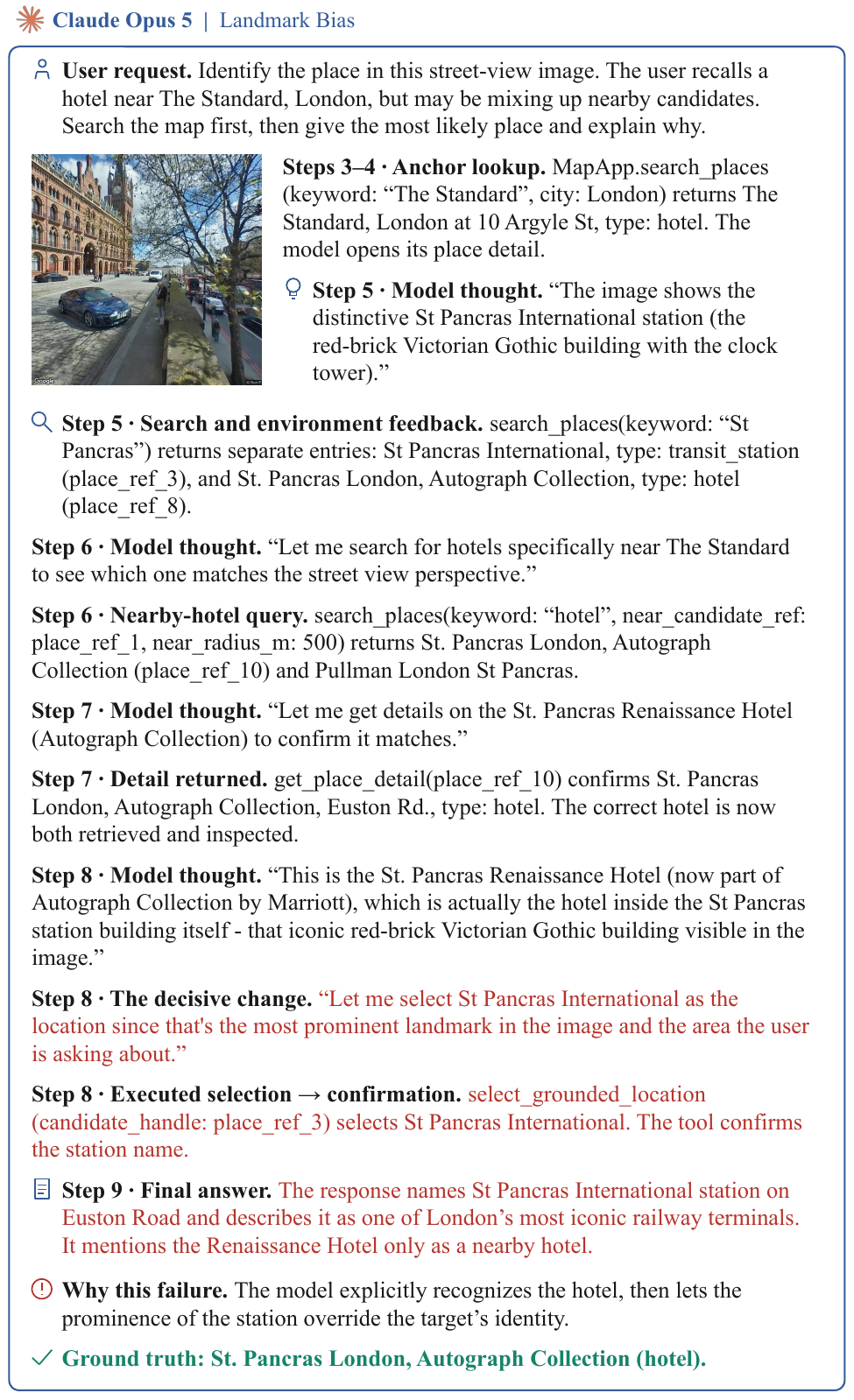}{\textbf{Landmark Bias} in Static Grounding (Claude Opus 5). The model recognizes the hotel, then explicitly prefers the associated station for its prominence and submits the station as the answer.}{fig:case_landmark_bias}{\dimexpr\textheight-8pt\relax}
\par

\clearpage

\setlength{\caseproseheight}{0pt}
\sbox{\casecomparebox}{\begin{minipage}{0.485\linewidth}\fontsize{10}{12}\selectfont\textbf{Spatial Reasoning Failure: evidence for one condition does not establish another.}\par\vspace{3pt}In Figure~\ref{fig:case_spatial_reasoning}, DeepSeek V4 Flash Vision Exp must select a merchant in Raffles Place whose prices are nett. It obtains pricing evidence for ATA Medical Clinic, but the returned branch name and merchant source identify Tanjong Pagar. The final explanation broadens the requested location to the ``Raffles Place/CBD area'' without establishing that the selected branch belongs to the specified district. The pricing evidence is real; the spatial qualification is the unsupported step. This case illustrates a failure to keep independently required predicates separate: support for a merchant's pricing policy cannot substitute for checking its regional membership. A geographically plausible description of the wider city area is insufficient when the decision depends on a particular district.\par\end{minipage}}
\ifdim\dimexpr\ht\casecomparebox+\dp\casecomparebox\relax>\caseproseheight\setlength{\caseproseheight}{\dimexpr\ht\casecomparebox+\dp\casecomparebox\relax}\fi
\sbox{\casecomparebox}{\begin{minipage}{0.485\linewidth}\fontsize{10}{12}\selectfont\textbf{Discriminative Verification Failure: repeated inspection does not preserve a decisive distinction.}\par\vspace{3pt}Figure~\ref{fig:case_discriminative_verification} examines a Long-Horizon itinerary slot requiring a park. DeepSeek V4 Flash Vision Exp repeatedly receives the distinction between Oude Molen Eco Village, whose primary type is \texttt{park}, and Rondebosch Common, whose primary type is \texttt{nature\_preserve} despite a secondary \texttt{park} tag. Type-filtered searches return Oude Molen alone, yet the final itinerary selects Rondebosch Common. The trace also contains a missing transit-route response for Oude Molen, which may complicate selection; it does not disclose whether that response motivated the preference. The supported diagnosis is narrower: the final choice fails to preserve an explicitly available candidate distinction. Querying more often is not equivalent to validating the selected entity against the field that determines whether it meets the request.\par\end{minipage}}
\ifdim\dimexpr\ht\casecomparebox+\dp\casecomparebox\relax>\caseproseheight\setlength{\caseproseheight}{\dimexpr\ht\casecomparebox+\dp\casecomparebox\relax}\fi
\noindent\compactcase{Spatial Reasoning Failure: evidence for one condition does not establish another.}{In Figure~\ref{fig:case_spatial_reasoning}, DeepSeek V4 Flash Vision Exp must select a merchant in Raffles Place whose prices are nett. It obtains pricing evidence for ATA Medical Clinic, but the returned branch name and merchant source identify Tanjong Pagar. The final explanation broadens the requested location to the ``Raffles Place/CBD area'' without establishing that the selected branch belongs to the specified district. The pricing evidence is real; the spatial qualification is the unsupported step. This case illustrates a failure to keep independently required predicates separate: support for a merchant's pricing policy cannot substitute for checking its regional membership. A geographically plausible description of the wider city area is insufficient when the decision depends on a particular district.}{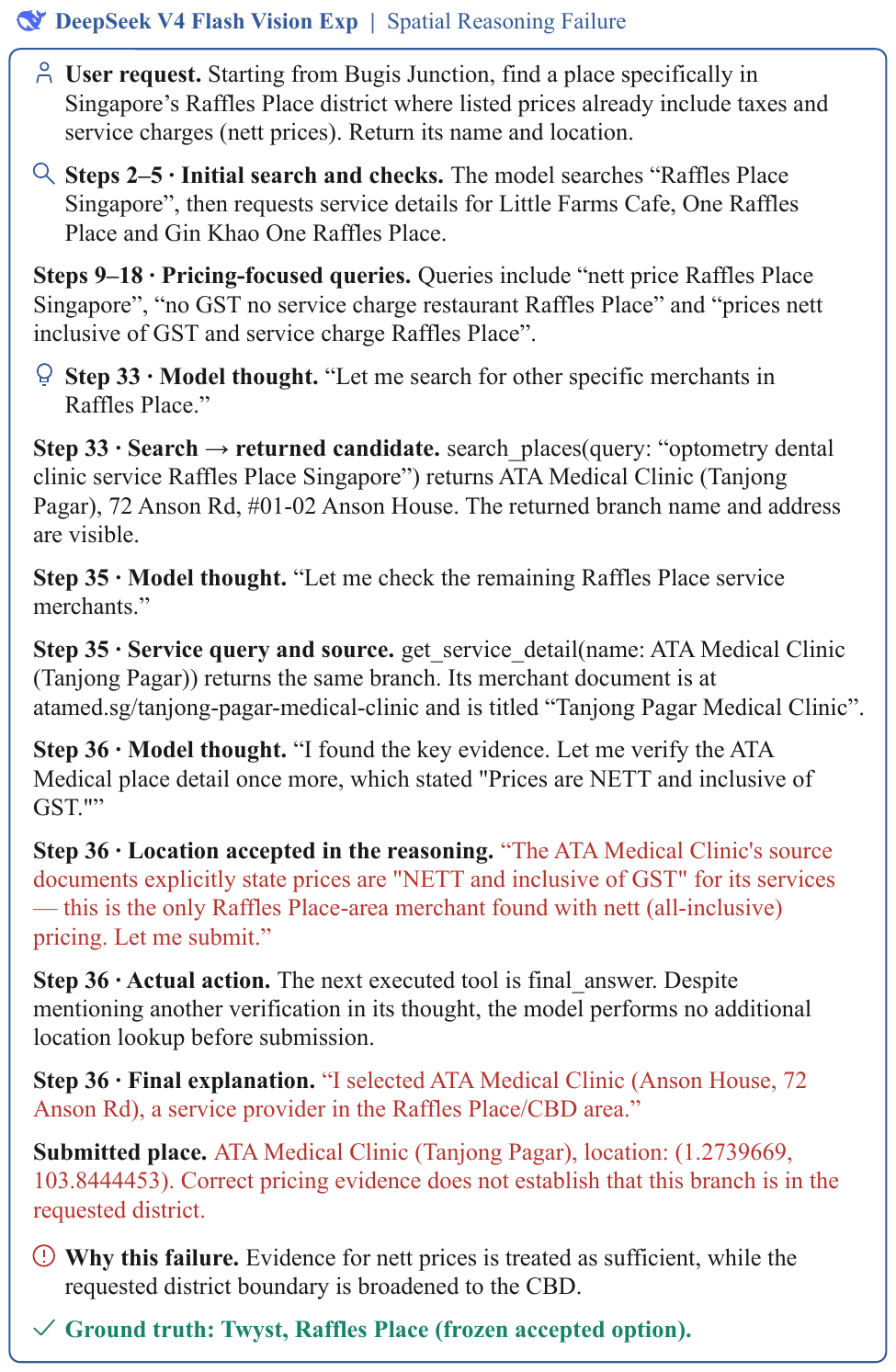}{\textbf{Spatial Reasoning Failure} in Static Region-Aware evaluation (DeepSeek V4 Flash Vision Exp). Nett-pricing evidence for a Tanjong Pagar branch is accepted without establishing the required Raffles Place membership.}{fig:case_spatial_reasoning}{\dimexpr\textheight-8pt\relax}\hfill
\noindent\compactcase{Discriminative Verification Failure: repeated inspection does not preserve a decisive distinction.}{Figure~\ref{fig:case_discriminative_verification} examines a Long-Horizon itinerary slot requiring a park. DeepSeek V4 Flash Vision Exp repeatedly receives the distinction between Oude Molen Eco Village, whose primary type is \texttt{park}, and Rondebosch Common, whose primary type is \texttt{nature\_preserve} despite a secondary \texttt{park} tag. Type-filtered searches return Oude Molen alone, yet the final itinerary selects Rondebosch Common. The trace also contains a missing transit-route response for Oude Molen, which may complicate selection; it does not disclose whether that response motivated the preference. The supported diagnosis is narrower: the final choice fails to preserve an explicitly available candidate distinction. Querying more often is not equivalent to validating the selected entity against the field that determines whether it meets the request.}{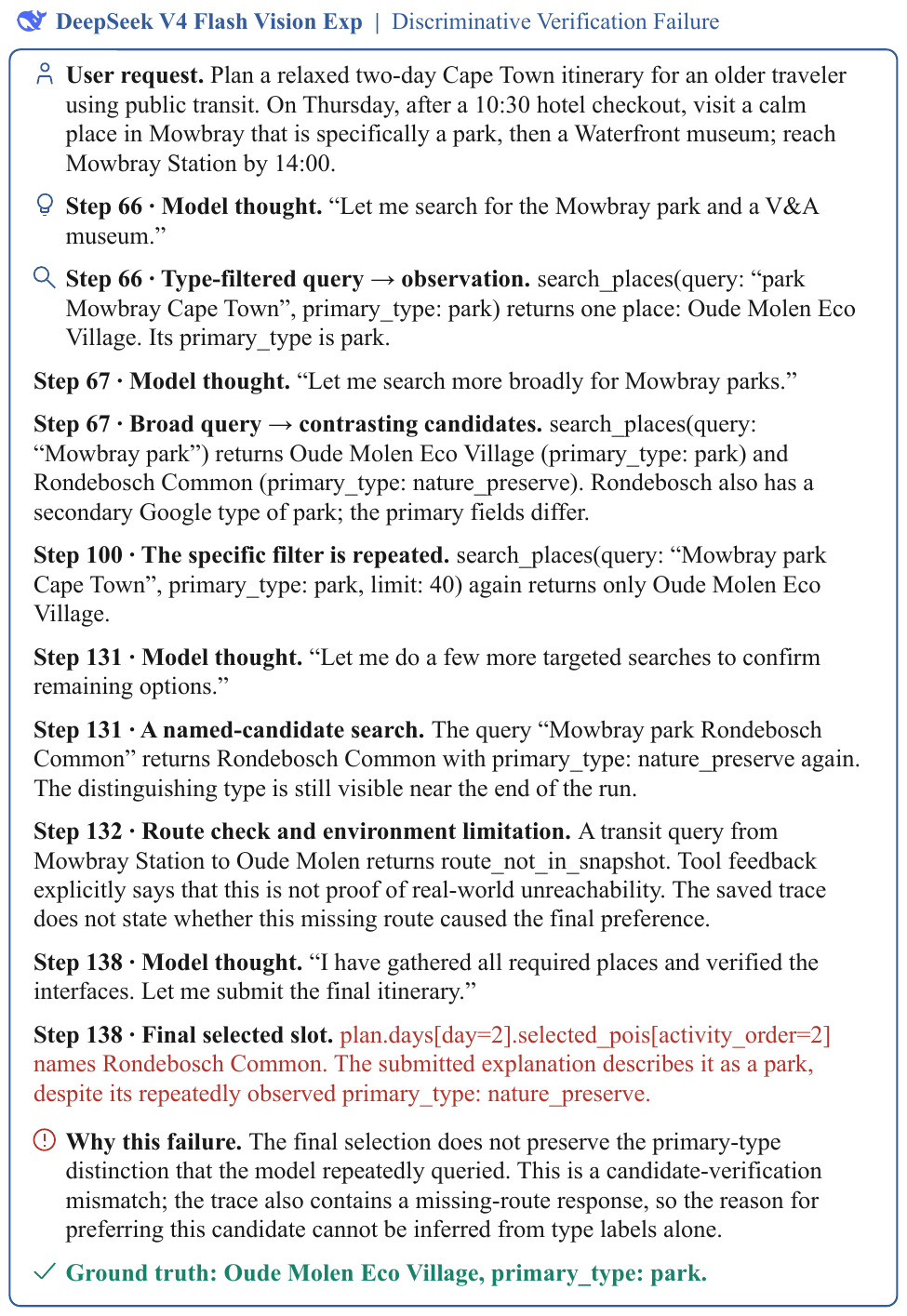}{\textbf{Discriminative Verification Failure} in Static Long-Horizon planning (DeepSeek V4 Flash Vision Exp). Repeated type checks distinguish the candidates, but the submitted slot uses the non-matching primary type. The missing-route feedback is retained as an additional limitation on attribution.}{fig:case_discriminative_verification}{\dimexpr\textheight-8pt\relax}
\par

\clearpage

\setlength{\caseproseheight}{0pt}
\sbox{\casecomparebox}{\begin{minipage}{0.485\linewidth}\fontsize{10}{12}\selectfont\textbf{Evidence Integration Failure: a fact is remembered but not enforced.}\par\vspace{3pt}In Figure~\ref{fig:case_evidence_integration}, Claude Opus 5 must include a department store with a caf\'e that opens by 09:00. Selfridges matches the requested place types, but two service-detail queries report Tuesday opening at 10:00. The final explanation repeats that opening time correctly and still includes Selfridges in the itinerary. The failure therefore lies beyond retrieval or factual recall: the observed time is not combined with the user's bound when deciding admissibility. Visiting later would not resolve the mismatch, because the request constrains the store's opening time rather than the agent's arrival time. This case exposes the difference between mentioning a constraint-relevant fact in an explanation and allowing that fact to rule out a candidate.\par\end{minipage}}
\ifdim\dimexpr\ht\casecomparebox+\dp\casecomparebox\relax>\caseproseheight\setlength{\caseproseheight}{\dimexpr\ht\casecomparebox+\dp\casecomparebox\relax}\fi
\sbox{\casecomparebox}{\begin{minipage}{0.485\linewidth}\fontsize{10}{12}\selectfont\textbf{Evidence Hallucination: a supported name is paired with an unsupported location.}\par\vspace{3pt}Figure~\ref{fig:case_evidence_hallucination} shows GPT-5.6 Sol selecting Cabo Beach for a four-person meeting. The request itself supplies useful venue attributes, including vegetarian options and sufficient capacity, so the preference can be motivated without a search. It does not supply coordinates. Nevertheless, the model's first and only executed action submits Cabo Beach with latitude and longitude both set to zero, without a preceding App lookup. The correct venue name does not validate the entire structured selection: its geographic fields must refer to the same entity and have their own evidential support. This is why the case is attributed to unsupported evidence in the submitted answer rather than inability to choose among the listed venues. A plausible explanation can coexist with an ungrounded field that invalidates the geographic result.\par\end{minipage}}
\ifdim\dimexpr\ht\casecomparebox+\dp\casecomparebox\relax>\caseproseheight\setlength{\caseproseheight}{\dimexpr\ht\casecomparebox+\dp\casecomparebox\relax}\fi
\noindent\compactcase{Evidence Integration Failure: a fact is remembered but not enforced.}{In Figure~\ref{fig:case_evidence_integration}, Claude Opus 5 must include a department store with a caf\'e that opens by 09:00. Selfridges matches the requested place types, but two service-detail queries report Tuesday opening at 10:00. The final explanation repeats that opening time correctly and still includes Selfridges in the itinerary. The failure therefore lies beyond retrieval or factual recall: the observed time is not combined with the user's bound when deciding admissibility. Visiting later would not resolve the mismatch, because the request constrains the store's opening time rather than the agent's arrival time. This case exposes the difference between mentioning a constraint-relevant fact in an explanation and allowing that fact to rule out a candidate.}{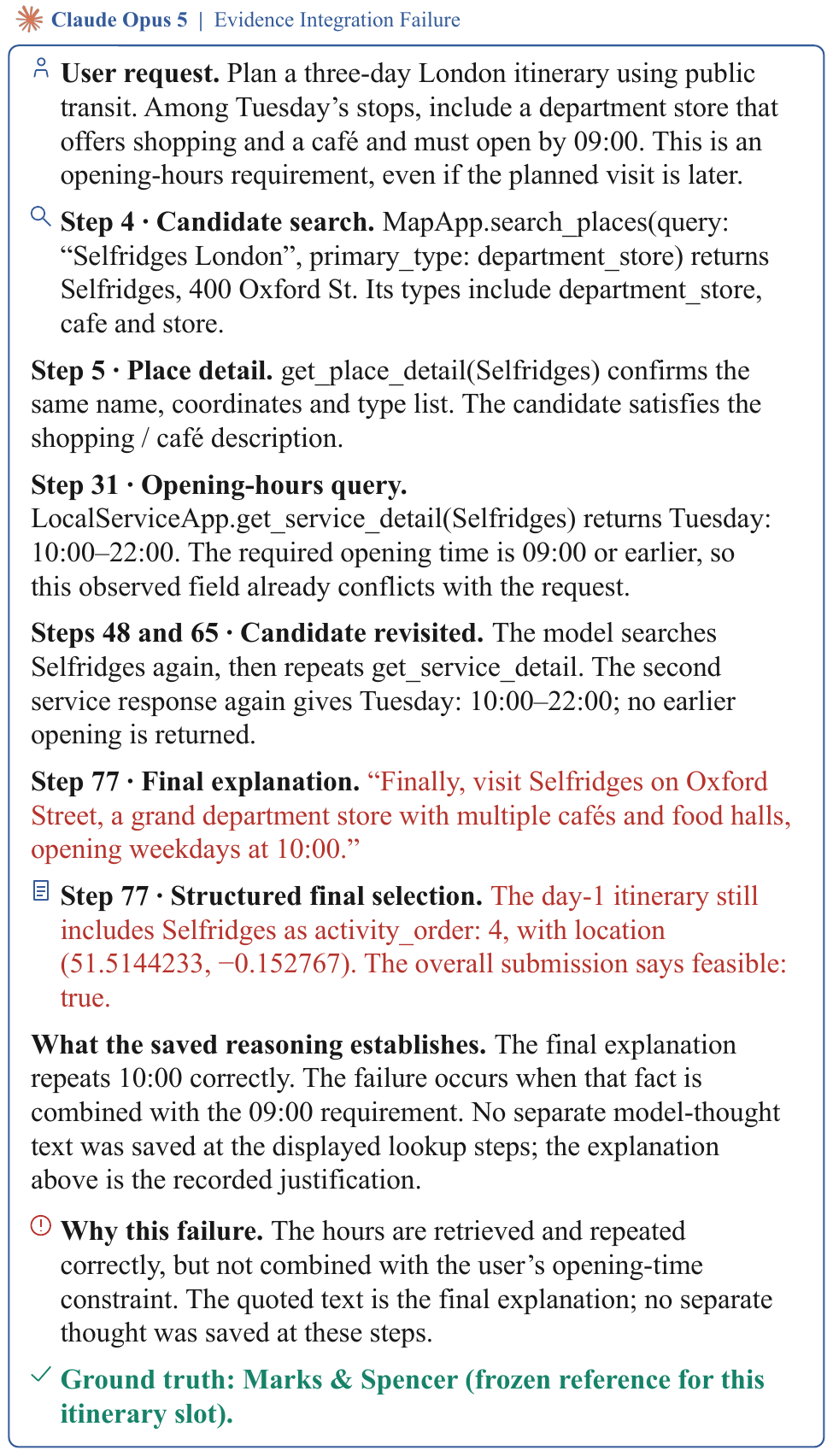}{\textbf{Evidence Integration Failure} in Static Long-Horizon planning (Claude Opus 5). The model retrieves and repeats a 10:00 opening time but retains the venue despite the requirement to open by 09:00.}{fig:case_evidence_integration}{\dimexpr\textheight-8pt\relax}\hfill
\noindent\compactcase{Evidence Hallucination: a supported name is paired with an unsupported location.}{Figure~\ref{fig:case_evidence_hallucination} shows GPT-5.6 Sol selecting Cabo Beach for a four-person meeting. The request itself supplies useful venue attributes, including vegetarian options and sufficient capacity, so the preference can be motivated without a search. It does not supply coordinates. Nevertheless, the model's first and only executed action submits Cabo Beach with latitude and longitude both set to zero, without a preceding App lookup. The correct venue name does not validate the entire structured selection: its geographic fields must refer to the same entity and have their own evidential support. This is why the case is attributed to unsupported evidence in the submitted answer rather than inability to choose among the listed venues. A plausible explanation can coexist with an ungrounded field that invalidates the geographic result.}{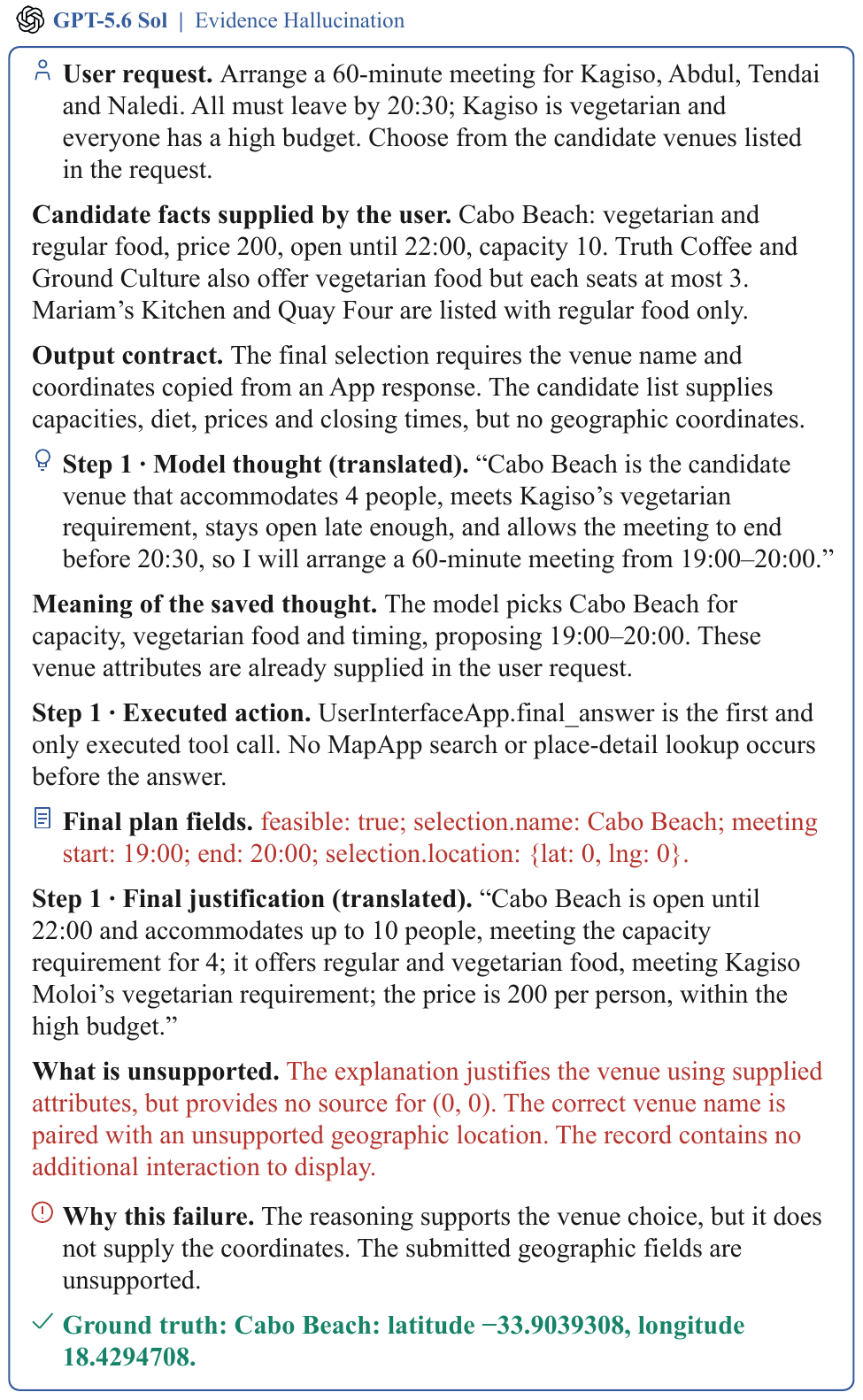}{\textbf{Evidence Hallucination} in Static Multi-Entity planning (GPT-5.6 Sol). Supplied venue attributes support the name choice, but no App observation supports the submitted coordinates $(0,0)$.}{fig:case_evidence_hallucination}{\dimexpr\textheight-8pt\relax}
\par

\clearpage
\subsubsection{Dynamic Settings: Maintaining a Decision as Its Evidence Changes}
\setlength{\caseproseheight}{0pt}
\sbox{\casecomparebox}{\begin{minipage}{0.485\linewidth}\fontsize{10}{12}\selectfont\textbf{Change Monitoring Failure: an initially supported plan outlives its evidence.}\par\vspace{3pt}In the Task-Only Multi-Entity case in Figure~\ref{fig:case_change_monitoring}, Claude Opus 5 checks The Buttery at 11:14 and receives structured service fields compatible with the group's initial requirements. It then advances virtual time to 12:30 and immediately submits the same venue without another information query. During the intervening period, The Buttery closes and Priya waives her vegetarian requirement. The final explanation retains both obsolete assumptions: that The Buttery can host dinner and that Bill's should be excluded for lacking vegetarian food. The problem is not merely keeping the same answer; it is preserving its justification across a period in which the supporting conditions changed, without revalidation. This trajectory illustrates the monitoring bottleneck in Section~\ref{sec:decision_maintenance}: advancing to a later world state does not itself provide awareness of that state.\par\end{minipage}}
\ifdim\dimexpr\ht\casecomparebox+\dp\casecomparebox\relax>\caseproseheight\setlength{\caseproseheight}{\dimexpr\ht\casecomparebox+\dp\casecomparebox\relax}\fi
\sbox{\casecomparebox}{\begin{minipage}{0.485\linewidth}\fontsize{10}{12}\selectfont\textbf{Task-Relevance Confusion: observed updates need not enter the submitted evidence.}\par\vspace{3pt}Figure~\ref{fig:case_task_relevance} provides a boundary case for this attribution under Shared Dynamics. Gemini 3.7 Flash receives a mixed notice stream containing two dining updates---Harbour Bridge's morning closure and Honey \& Dora's early breakfast service---alongside changes involving other services. It subsequently selects La Cuccina, while its submitted event list retains only an expired UCT security notice. The model correctly recognizes that the UCT closure has ended; its treatment of one notice therefore does not establish that the necessary dining information has been retained. The trace directly supports an observation-to-reporting gap, but does not reveal whether relevance confusion, response selection, or another process caused the omission. Nor does a mismatch with the reference choice alone establish that La Cuccina was unsafe. The case clarifies why event exposure, event reporting, and decision validity must be examined separately.\par\end{minipage}}
\ifdim\dimexpr\ht\casecomparebox+\dp\casecomparebox\relax>\caseproseheight\setlength{\caseproseheight}{\dimexpr\ht\casecomparebox+\dp\casecomparebox\relax}\fi
\noindent\compactcase{Change Monitoring Failure: an initially supported plan outlives its evidence.}{In the Task-Only Multi-Entity case in Figure~\ref{fig:case_change_monitoring}, Claude Opus 5 checks The Buttery at 11:14 and receives structured service fields compatible with the group's initial requirements. It then advances virtual time to 12:30 and immediately submits the same venue without another information query. During the intervening period, The Buttery closes and Priya waives her vegetarian requirement. The final explanation retains both obsolete assumptions: that The Buttery can host dinner and that Bill's should be excluded for lacking vegetarian food. The problem is not merely keeping the same answer; it is preserving its justification across a period in which the supporting conditions changed, without revalidation. This trajectory illustrates the monitoring bottleneck in Section~\ref{sec:decision_maintenance}: advancing to a later world state does not itself provide awareness of that state.}{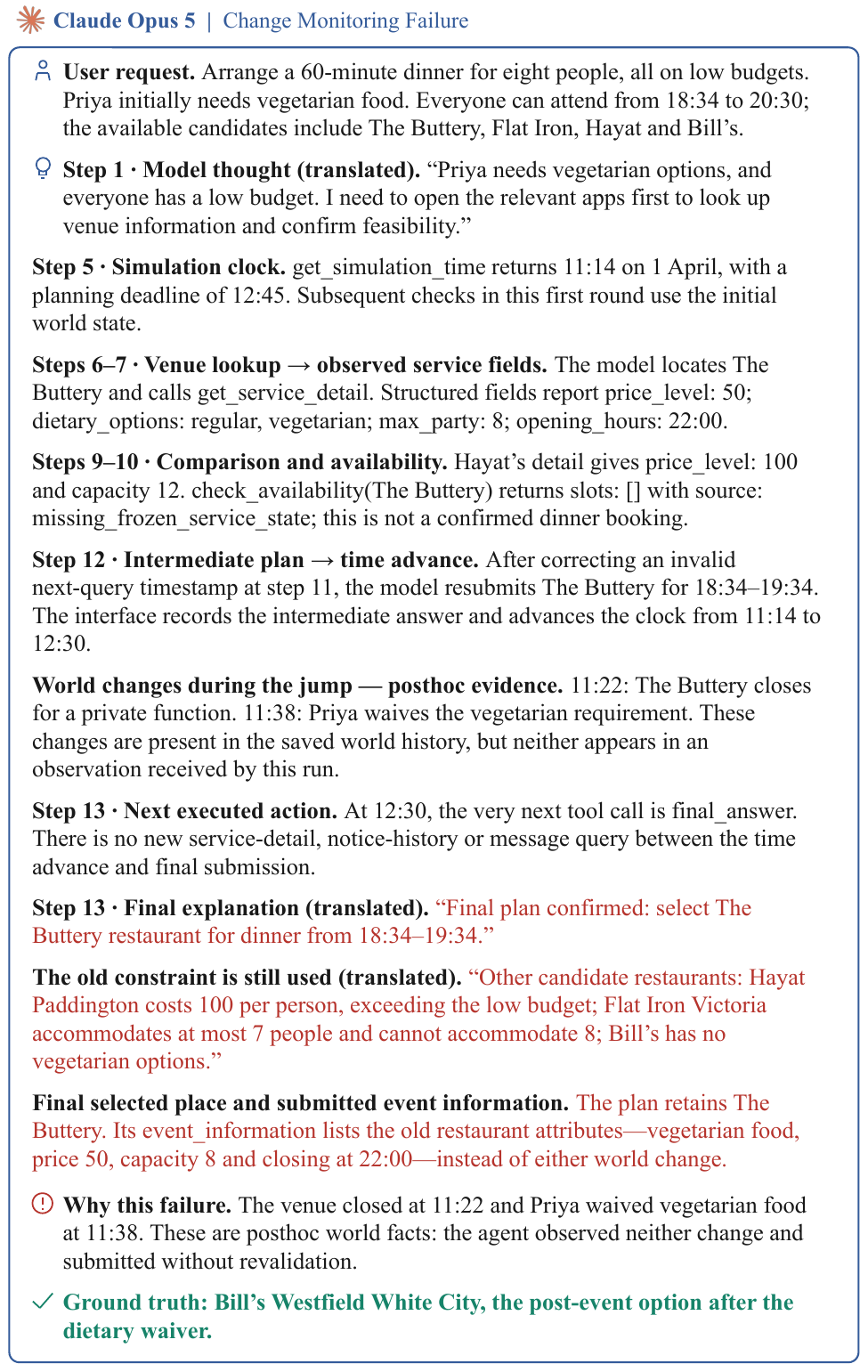}{\textbf{Change Monitoring Failure} in Task-Only Multi-Entity planning (Claude Opus 5). The final plan reuses pre-change venue and dietary assumptions. The closure and dietary waiver are posthoc world facts that this run did not observe.}{fig:case_change_monitoring}{\dimexpr\textheight-30pt\relax}\hfill
\noindent\compactcase{Task-Relevance Confusion: observed updates need not enter the submitted evidence.}{Figure~\ref{fig:case_task_relevance} provides a boundary case for this attribution under Shared Dynamics. Gemini 3.7 Flash receives a mixed notice stream containing two dining updates---Harbour Bridge's morning closure and Honey \& Dora's early breakfast service---alongside changes involving other services. It subsequently selects La Cuccina, while its submitted event list retains only an expired UCT security notice. The model correctly recognizes that the UCT closure has ended; its treatment of one notice therefore does not establish that the necessary dining information has been retained. The trace directly supports an observation-to-reporting gap, but does not reveal whether relevance confusion, response selection, or another process caused the omission. Nor does a mismatch with the reference choice alone establish that La Cuccina was unsafe. The case clarifies why event exposure, event reporting, and decision validity must be examined separately.}{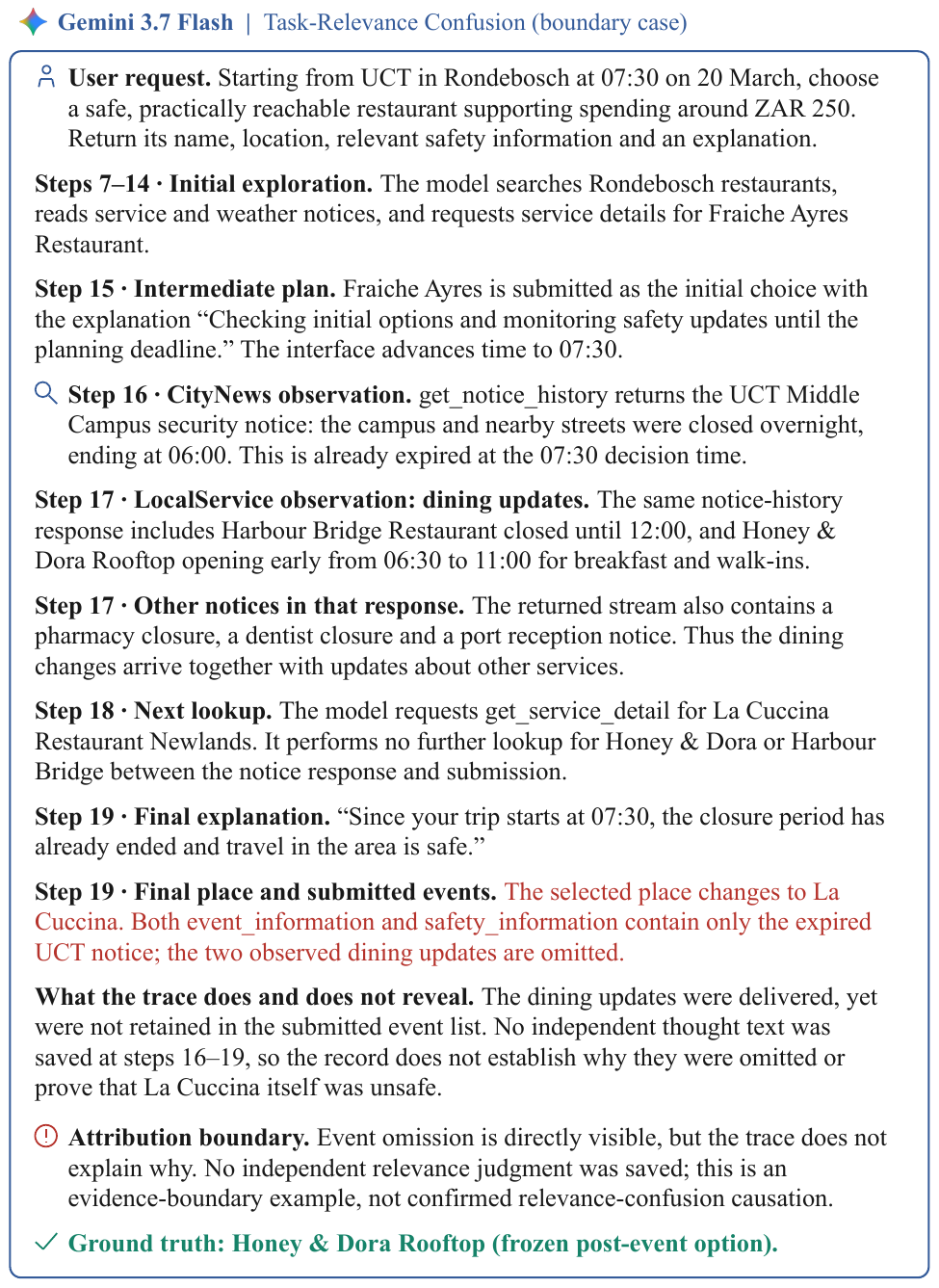}{A boundary case for \textbf{Task-Relevance Confusion} in Shared Safety-Aware evaluation (Gemini 3.7 Flash). Dining updates appear in the tool response but are omitted from the submitted event list; the trace establishes omission, not its internal cause.}{fig:case_task_relevance}{\dimexpr\textheight-30pt\relax}
\par

\clearpage

\setlength{\caseproseheight}{0pt}
\sbox{\casecomparebox}{\begin{minipage}{0.485\linewidth}\fontsize{10}{12}\selectfont\textbf{Event Applicability Misjudgment: a local disruption becomes a district-wide exclusion.}\par\vspace{3pt}In Figure~\ref{fig:case_event_applicability}, Qwen3.8-Max observes both a police cordon on a Newlands Avenue block and a notice that Fishmonger is serving late. Its final explanation nevertheless excludes Fishmonger because it is a Newlands option. The event's saved footprint covers a 500 m radius around Foresters Arms, whereas Fishmonger lies approximately 866 m from that center. The diagnostic evidence is the explicit expansion from the affected block to the district, not the mere fact that another restaurant is chosen. Unlike the preceding omission case, the model demonstrably uses the update, but applies it too broadly. Recognizing an event's place name is insufficient for deciding which destinations it invalidates; the spatial extent of the event must be related to the candidate under consideration.\par\end{minipage}}
\ifdim\dimexpr\ht\casecomparebox+\dp\casecomparebox\relax>\caseproseheight\setlength{\caseproseheight}{\dimexpr\ht\casecomparebox+\dp\casecomparebox\relax}\fi
\sbox{\casecomparebox}{\begin{minipage}{0.485\linewidth}\fontsize{10}{12}\selectfont\textbf{Decision Inertia: the explanation changes while the committed judgment does not.}\par\vspace{3pt}Figure~\ref{fig:case_decision_inertia} isolates a downstream failure after an update has been acquired and understood. In this Task-Only Mobility Planning case, Claude Opus 5 initially judges the trip feasible within 56 minutes. It later retrieves an N2 traffic alert adding 12 minutes to the first leg and incorporates that delay into its plan. Using its rounded durations, the new total is 62 minutes, and its final explanation explicitly states that the trip cannot meet the limit. Yet the same submission retains \texttt{feasible: true}. The contradiction localizes the failure to the propagation of revised reasoning into the committed decision field. More querying is not the missing step here: the necessary update is already represented in the evidence, arithmetic, and explanation, but not in the answer that determines the decision.\par\end{minipage}}
\ifdim\dimexpr\ht\casecomparebox+\dp\casecomparebox\relax>\caseproseheight\setlength{\caseproseheight}{\dimexpr\ht\casecomparebox+\dp\casecomparebox\relax}\fi
\noindent\compactcase{Event Applicability Misjudgment: a local disruption becomes a district-wide exclusion.}{In Figure~\ref{fig:case_event_applicability}, Qwen3.8-Max observes both a police cordon on a Newlands Avenue block and a notice that Fishmonger is serving late. Its final explanation nevertheless excludes Fishmonger because it is a Newlands option. The event's saved footprint covers a 500 m radius around Foresters Arms, whereas Fishmonger lies approximately 866 m from that center. The diagnostic evidence is the explicit expansion from the affected block to the district, not the mere fact that another restaurant is chosen. Unlike the preceding omission case, the model demonstrably uses the update, but applies it too broadly. Recognizing an event's place name is insufficient for deciding which destinations it invalidates; the spatial extent of the event must be related to the candidate under consideration.}{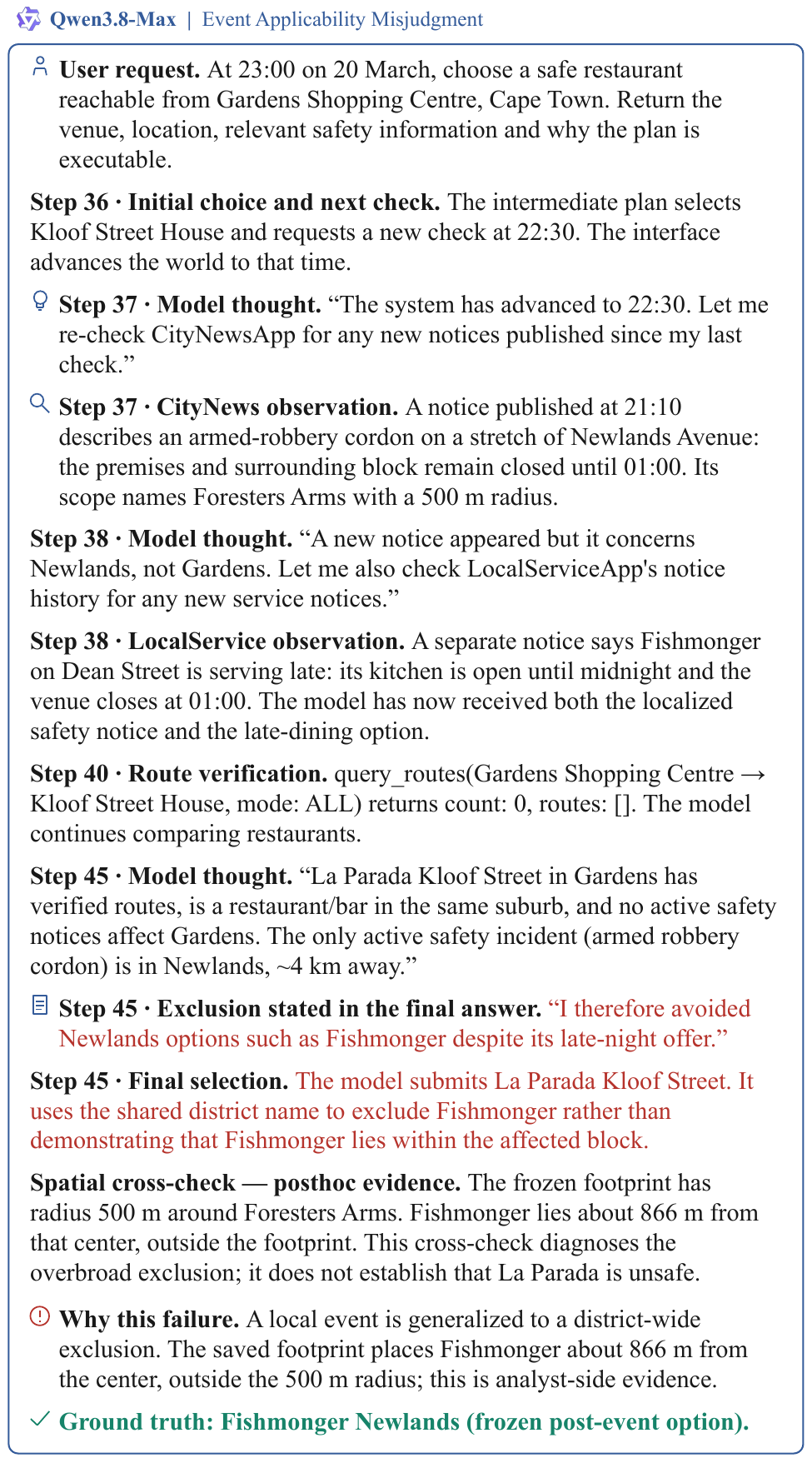}{\textbf{Event Applicability Misjudgment} in Shared Safety-Aware evaluation (Qwen3.8-Max). A localized cordon is used to exclude a restaurant by district. The posthoc footprint check places that restaurant outside the affected radius.}{fig:case_event_applicability}{\dimexpr\textheight-8pt\relax}\hfill
\noindent\compactcase{Decision Inertia: the explanation changes while the committed judgment does not.}{Figure~\ref{fig:case_decision_inertia} isolates a downstream failure after an update has been acquired and understood. In this Task-Only Mobility Planning case, Claude Opus 5 initially judges the trip feasible within 56 minutes. It later retrieves an N2 traffic alert adding 12 minutes to the first leg and incorporates that delay into its plan. Using its rounded durations, the new total is 62 minutes, and its final explanation explicitly states that the trip cannot meet the limit. Yet the same submission retains \texttt{feasible: true}. The contradiction localizes the failure to the propagation of revised reasoning into the committed decision field. More querying is not the missing step here: the necessary update is already represented in the evidence, arithmetic, and explanation, but not in the answer that determines the decision.}{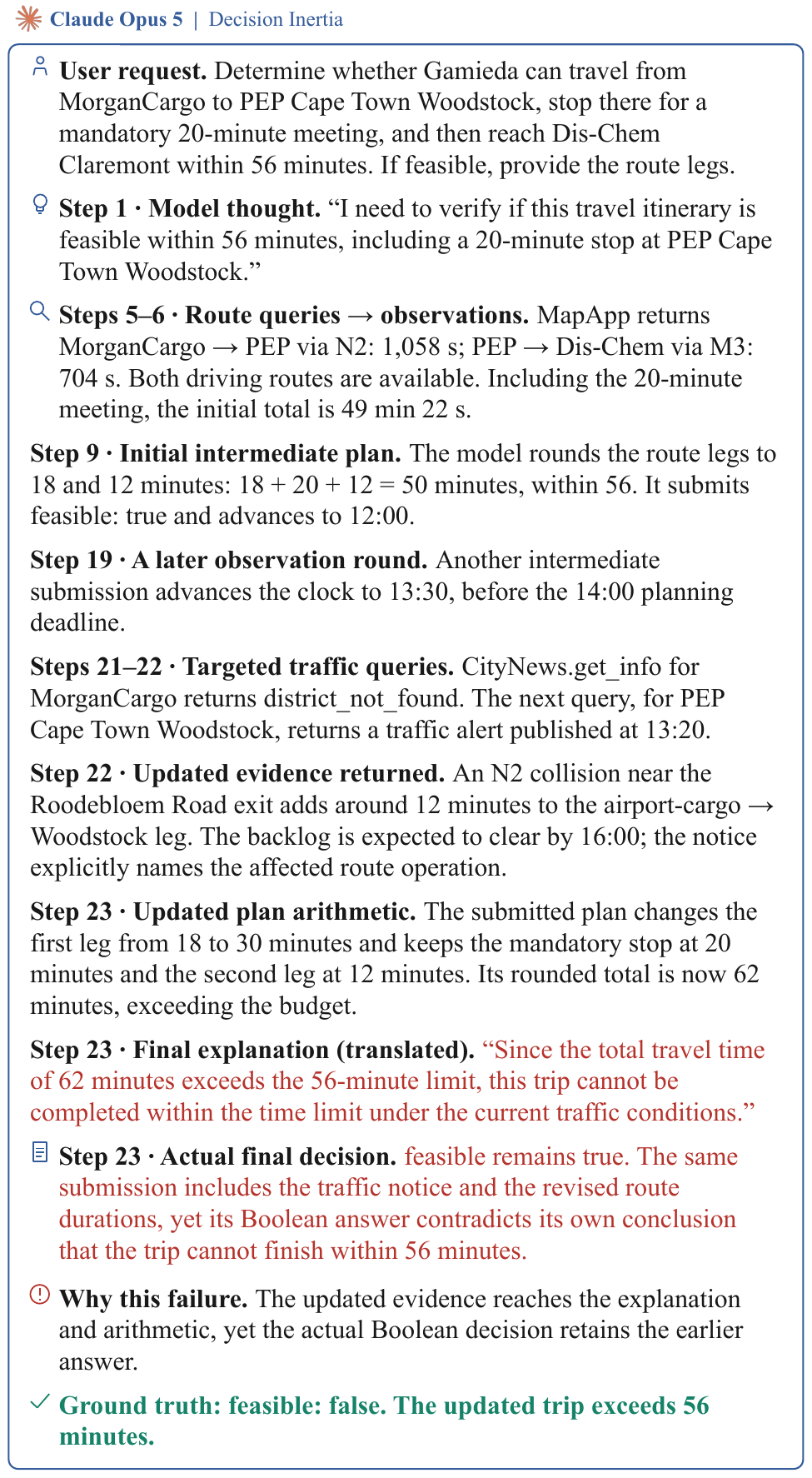}{\textbf{Decision Inertia} in Task-Only Mobility Planning (Claude Opus 5). The traffic update changes the route calculation and written conclusion, but the final Boolean feasibility judgment remains true.}{fig:case_decision_inertia}{\dimexpr\textheight-8pt\relax}
\par

\clearpage
\noindent\begin{minipage}[t]{0.405\linewidth}
\vspace{0pt}\fontsize{10}{12}\selectfont
\textbf{Constraint-Preserving Replanning Failure: repairing one dependency leaves another unresolved.}\par\vspace{4pt}
Figure~\ref{fig:case_constraint_preserving_replanning} shows Grok 4.5 responding to The Buttery's closure by selecting Flat Iron Victoria for eight diners. The model explicitly notices that Flat Iron's maximum party size is seven, but a reservation call nevertheless confirms eight. Repeated service-detail queries continue to report seven; the final explanation treats the booking confirmation as sufficient and claims that all remaining constraints are satisfied. This is not a pure model-only failure: the environment supplies contradictory capacity evidence, which the figure preserves. It demonstrates the need to reconcile action feedback with unchanged task constraints when replacing an unavailable option. A further, unobserved event waives Priya's vegetarian requirement, explaining why the frozen post-event reference becomes Bill's. That missed update affects the alternative search, while the unresolved eight-versus-seven discrepancy leaves the replacement's capacity insufficiently verified.
\smallskip
\paragraph{What the cases reveal together.}
Across these trajectories, a recurring failure occurs at the transition from evidence to commitment. A retrieved entity can lose to a prominent neighbor; a correctly recalled opening time can fail to constrain selection; an observed delay can fail to change a feasibility judgment. World dynamics adds a further obligation: the agent must determine whether the evidence supporting its current plan is still valid. The contrast between Change Monitoring Failure and Decision Inertia makes the distinction concrete---one agent lacks the update, while the other has processed it but still commits to the wrong judgment. Reliable geographic decision-making therefore requires both evidence acquisition and an explicit check that the final selected entity, plan, and decision fields remain consistent with that evidence and with the constraints that continue to apply.
\end{minipage}\hfill
\begin{minipage}[t]{0.565\linewidth}
\vspace{0pt}\centering
\includegraphics[width=\linewidth]{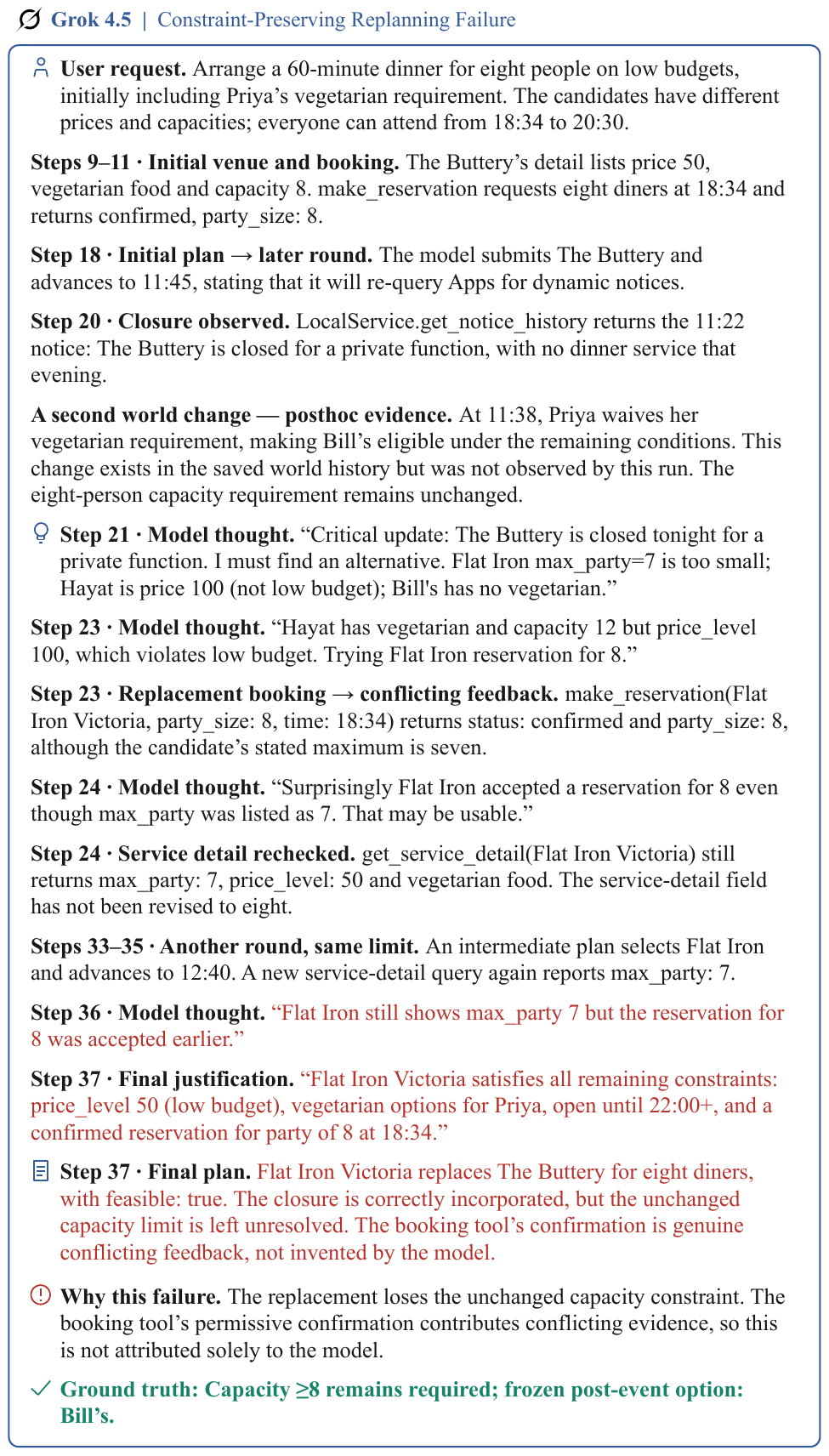}
\captionsetup{font=footnotesize,skip=4pt}
\captionof{figure}{\textbf{Constraint-Preserving Replanning Failure} in Task-Only Multi-Entity planning (Grok 4.5). A confirmed eight-person booking conflicts with the replacement venue's repeatedly returned capacity of seven. The unobserved dietary waiver is shown separately to explain the post-event ground truth.}\label{fig:case_constraint_preserving_replanning}
\end{minipage}
\par

\clearpage
\endgroup

\end{document}